\documentclass{article} 
\usepackage[table,HTML,xcdraw]{xcolor}
\usepackage{iclr2025_conference,times}

\usepackage{hyperref}
\usepackage{url}

\usepackage{multirow}
\usepackage{graphicx} 
\usepackage{array}
\usepackage{booktabs}
\usepackage{algorithm}
\usepackage{algpseudocode}
\usepackage{arydshln}
\usepackage[utf8]{inputenc}
\usepackage[T1]{fontenc}
\usepackage{tikz}
\usepackage{textcomp} 
\usepackage{amssymb,mathrsfs,amsmath}
\usepackage{MnSymbol,bbding,pifont}
\usepackage{graphicx}
\usepackage{subfig}
\usepackage{tcolorbox}
\usepackage{makecell}  
\usepackage{cancel}  
\usepackage{fontawesome}
\usepackage{cleveref}
\crefformat{section}{\S#2#1#3}
\crefformat{subsection}{\S#2#1#3}
\crefformat{subsubsection}{\S#2#1#3}
\crefrangeformat{section}{\S\S#3#1#4 to~#5#2#6}
\crefmultiformat{section}{\S\S#2#1#3}{ and~#2#1#3}{, #2#1#3}{ and~#2#1#3}
\Crefformat{figure}{#2Figure~#1#3}
\Crefmultiformat{figure}{Figs.~#2#1#3}{ and~#2#1#3}{, #2#1#3}{ and~#2#1#3}
\Crefformat{table}{#2Table~#1#3}
\Crefmultiformat{table}{Tabs.~#2#1#3}{ and~#2#1#3}{, #2#1#3}{ and~#2#1#3}
\Crefformat{appendix}{Appx.~\S#2#1#3}
\crefformat{algorithm}{Alg.~#2#1#3}
\Crefformat{equation}{Eq.~(#2#1#3)}
\Crefmultiformat{equation}{Eqs.~(#2#1#3)}{ and~(#2#1#3)}{, (#2#1#3)}{ and~(#2#1#3)}
\usepackage[edges]{forest}
\usepackage[framemethod=tikz]{mdframed}
\usetikzlibrary{arrows.meta,shapes,positioning,shadows,trees}
\usepackage{adjustbox}

\usepackage{marvosym}
\usepackage{CJKutf8}

\title{100 Days After DeepSeek-R1: \\ 
A Survey on Replication Studies and More \\ 
Directions for Reasoning Language Models}

\author{Chong Zhang\thanks{Equal contribution. Co-first authors are listed in random order, as are the co-authors. All authors made substantial and complementary contributions to the development of this work.
}~~$^{1,2}$~~,~~Yue Deng\footnotemark[1]~~$^{1}$~~,~~Xiang Lin\footnotemark[1]~~$^{1}$~~,~~Bin Wang\footnotemark[1]~~$^{1}$~~,~~Dianwen Ng$^{1}$,~~Hai Ye$^{1,3}$, \\
\textbf{Xingxuan Li$^{1}$,~~Yao Xiao$^{1,4}$,~~Zhanfeng Mo$^{1,5}$,~~Qi Zhang$^{2}$,~~Lidong Bing$^{1}$\thanks{\noindent Corresponding author: Lidong Bing <lidong.bing@miromind.ai>. 
}\vspace{2mm}} \\ 
$^{1}$MiroMind \\
$^{2}$Fudan University \\
$^{3}$National University of Singapore \\
$^{4}$Singapore University of Technology and Design \\
$^{5}$Nanyang Technological University  
}

\iclrfinalcopy 
\begin{document}

\pagestyle{firstpage} 

\maketitle

\begin{center}
    \vspace{-3mm}
    \includegraphics[width=0.9\textwidth]{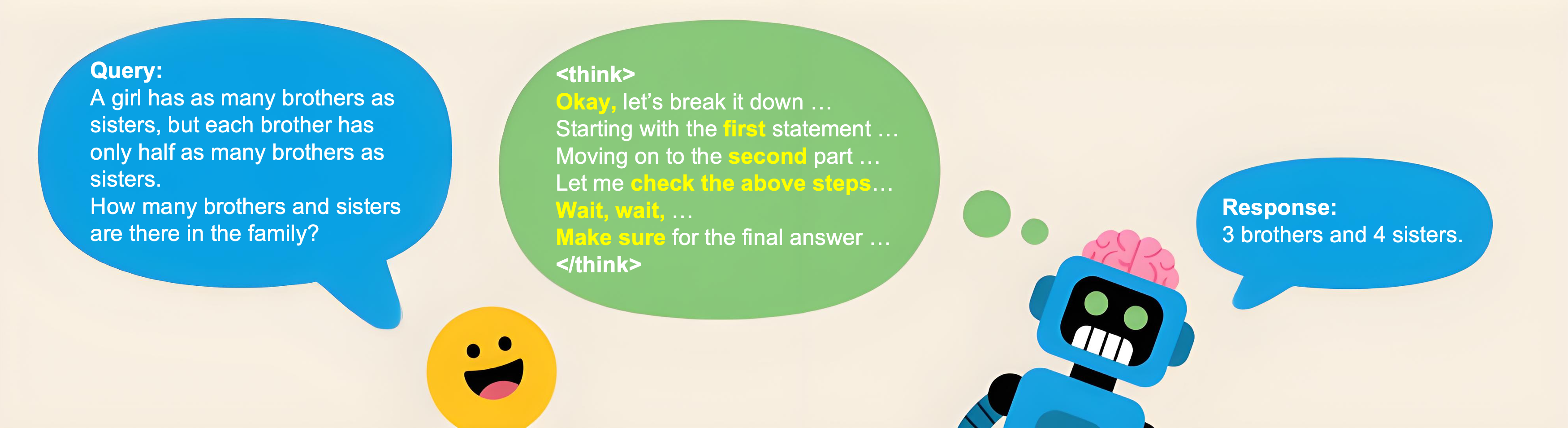} 
    \vspace{4mm}
\end{center}

\begin{abstract}

The recent development of reasoning language models (RLMs) represents a novel evolution in large language models. In particular, the recent release of DeepSeek-R1 has generated widespread social impact and sparked enthusiasm in the research community for exploring the explicit reasoning paradigm of language models. However, the implementation details of the released models have not been fully open-sourced by DeepSeek, including DeepSeek-R1-Zero, DeepSeek-R1, and the distilled small models. As a result, many replication studies have emerged aiming to reproduce the strong performance achieved by DeepSeek-R1, reaching comparable performance through similar training procedures and fully open-source data resources. These works have investigated feasible strategies for supervised fine-tuning (SFT) and reinforcement learning from verifiable rewards (RLVR), focusing on data preparation and method design, yielding various valuable insights. 
In this report, we provide a summary of recent replication studies to inspire future research. We primarily focus on SFT and RLVR as two main directions, introducing the details for data construction, method design and training procedure of current replication studies. 
Moreover, we conclude key findings from the implementation details and experimental results reported by these studies, anticipating to inspire future research. 
We also discuss additional techniques of enhancing RLMs, highlighting the potential of expanding the application scope of these models, and discussing the challenges in development. By this survey, we aim to help researchers and developers of RLMs stay updated with the latest advancements, and seek to inspire new ideas to further enhance RLMs.

\end{abstract}

\newpage

\pagestyle{empty} 

\setcounter{tocdepth}{3} 
\tableofcontents

\newpage
\pagestyle{plain}
\addtolength{\topmargin}{-25.74332pt} 
\setcounter{page}{1}  

\section{Introduction}

Reasoning language models (RLMs) have emerged as a transformative advancement in the evolution of large language models (LLMs), such as OpenAI o-series \citep{openai-o1,openai-o3}, DeepSeek-R1 \citep{Deepseek_R1}, and QwQ series \citep{qwq-32b-preview,qwq-32b}. Unlike conventional LLMs that merely generate unstructured responses, these models incorporate an explicit chain-of-thought process, providing step-by-step reasoning that mimics human cognitive processes--such as invoking self-verification, reflection, and more. This shift quickly attracted attention of the LLM research community, as it meets the growing demand for better explainability in complex tasks like mathematical problem solving, code generation, and logical reasoning, as well as the pursuit of steadily increasing accuracy. The significance of RLMs lies in their potential to enhance the accuracy of language models' response with trustful rationales. Rather than only providing answers, these models also reveal their thought process, which is invaluable for educational purposes, critical decision-making, and debugging AI reasoning errors. 
DeepSeek-R1 \citep{Deepseek_R1} is a prime example of this new generation. It leverages innovative training techniques such as supervised fine-tuning (SFT) and reinforcement learning from verifiable rewards (RLVR), allowing the model to develop powerful reasoning behaviors with an affordable amount of supervised data. Especially, by using methods like RLVR and incorporating cold-start data, it achieves performance comparable to prior models (e.g. OpenAI o1 \citep{openai-o1}) with relatively low training costs.

However, many critical details remain undisclosed which is required to replicate the reasoning performance and the behaviors of self-verification and reflection exhibited by DeepSeek-R1.
Although the DeepSeek-R1 team has publicly released their solution as training the model with Group Relative Policy Optimization (GRPO, \citet{GRPO}) and rule-based reward systems, the optimal design of the reinforcement learning algorithm and reward system remains underexplored.
Moreover, the training data and configurations of the SFT and RLVR stages are not released, leaving the impact on model performance to be further examined.
In response, many replication works have attempted to explore the optimal design for RLMs from various perspectives (see Figure~\ref{fig:paper-taxonomy}), yet a comprehensive list and comparison of these works are still lacking.

\tikzstyle{defaultbox}=[
rectangle,
draw=black,
rounded corners,
text opacity=1,
minimum height=1.5em,
minimum width=5em,
inner sep=2pt,
align=center,
fill opacity=.5,
fill=blue!10, 
]
\tikzstyle{level1}=[
defaultbox, 
minimum height=1.5em,
text=black,
align=center,
inner xsep=5pt,
inner ysep=4pt,
text width=9em,
anchor=center
]
\tikzstyle{level2}=[
level1,
text width=17em,
]
\tikzstyle{leaf}=[
level2, 
font=\normalsize,
align=left,
text width=42em,
]

\definecolor{SFT1}{HTML}{a0c4ff}
\definecolor{SFT2}{HTML}{c0d8ff}
\definecolor{SFT3}{HTML}{e0f0ff}
\definecolor{RL1}{HTML}{90e0ef}
\definecolor{RL2}{HTML}{a8edf5}
\definecolor{RL3}{HTML}{d0fbff}
\definecolor{Other1}{HTML}{bdb2ff}
\definecolor{Other2}{HTML}{d8ccff}
\definecolor{Other3}{HTML}{f0e8ff}
\definecolor{theme}{HTML}{E0D8D3}

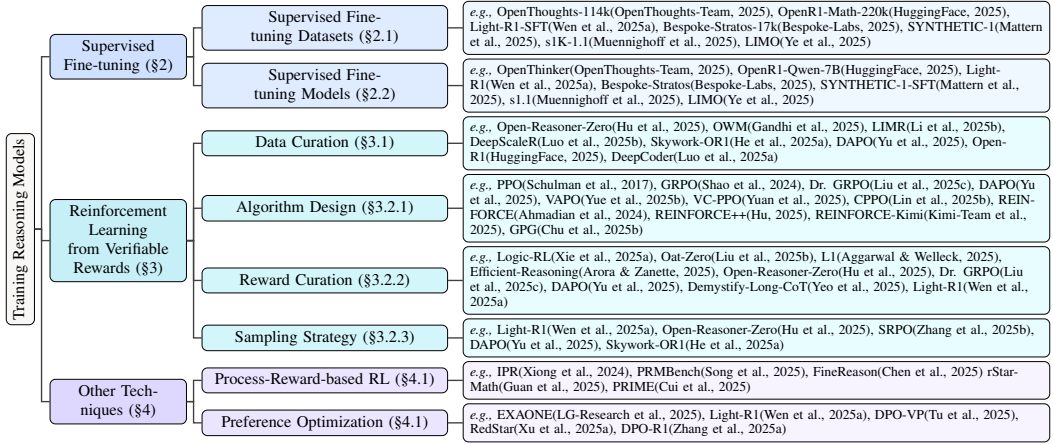
\begin{figure*}[t]
\centering
\resizebox{\textwidth}{!}
{
    \begin{forest}
        forked edges,
        for tree={
            grow=east,
            reversed=true,
            anchor=base west,
            parent anchor=east,
            child anchor=west,
            base=left,
            font=\large,
            rectangle,
            draw=black,
            rounded corners,
            minimum width=4em,
            edge+={darkgray, line width=1pt},
            s sep=3pt,
            ver/.style={rotate=90, child anchor=north, parent anchor=south, anchor=center, align=center, fill=theme!20,} %
        },
        [Training Reasoning Models, ver
            [Supervised Fine-tuning~(\S\ref{sec:sft}), level1, fill=SFT1
                [Supervised Fine-tuning Datasets~(\S\ref{sec:sft_datasets}), level2,fill=SFT2
                    [\textit{e.g.,}
                        OpenThoughts-114k\citep{openthoughts}{,} 
                        OpenR1-Math-220k\citep{openr1}{,}
                        Light-R1-SFT\citep{wen2025light}{,}
                        Bespoke-Stratos-17k\citep{bespoke_stratos}{,}
                        SYNTHETIC-1\citep{2025synthetic1}{,}
                        s1K-1.1\citep{muennighoff2025s1simpletesttimescaling}{,}
                        LIMO\citep{ye2025limoreasoning}
                        , leaf, fill=SFT3]
                ]
                [Supervised Fine-tuning Models~(\S\ref{sec:sft_models}), level2,fill=SFT2
                    [\textit{e.g.,}
                        OpenThinker\citep{openthoughts}{,} 
                        OpenR1-Qwen-7B\citep{openr1}{,}
                        Light-R1\citep{wen2025light}{,}
                        Bespoke-Stratos\citep{bespoke_stratos}{,}
                        SYNTHETIC-1-SFT\citep{2025synthetic1}{,}
                        s1.1\citep{muennighoff2025s1simpletesttimescaling}{,}
                        LIMO\citep{ye2025limoreasoning}
                        , leaf, fill=SFT3]
                ]
            ]   
            [Reinforcement Learning\\from Verifiable Rewards~(\S\ref{sec:rl}), level1,fill=RL1
                [Data Curation~(\S\ref{sec:rl_data}), level2,fill=RL2
                    [\textit{e.g.,}
                        Open-Reasoner-Zero\citep{OpenReasonerZero2025}{,}
                        OWM\citep{gandhi2025cognitive}{,}
                        LIMR\citep{li2025limr}{,}
                        DeepScaleR\citep{deepscaler2025}{,}
                        Skywork-OR1\citep{skywork-or1-2025}{,}
                        DAPO\citep{DAPO}{,}
                        Open-R1\citep{openr1}{,}
                        DeepCoder\citep{deepcoder2025}
                    ,leaf,fill=RL3]
                ]
                [Algorithm Design~(\S\ref{sec:rl_algorithm}), level2,fill=RL2
                    [\textit{e.g.,}
                        PPO\citep{PPO}{,}
                        GRPO\citep{GRPO}{,}
                        Dr. GRPO\citep{DrGRPO}{,}
                        DAPO\citep{DAPO}{,}
                        VAPO\citep{yuyue2025vapoefficientreliablereinforcement-vapo}{,}
                        VC-PPO\citep{VC-PPO}{,}
                        CPPO\citep{lin2025cppo}{,}
                        REINFORCE\citep{reinforce_LLM}{,}
                        REINFORCE++\citep{reinforce++}{,}
                        REINFORCE-Kimi\citep{KIMI_scaling_RL}{,}
                        GPG\citep{chu2025gpg}
                    , leaf,fill=RL3]
                ]
                [Reward Curation~(\S\ref{sec:rl_reward}), level2,fill=RL2
                    [\textit{e.g.,}
                        Logic-RL\citep{xie2025logicrlunleashingllmreasoning}{,}
                        Oat-Zero\citep{liu2025there}{,}
                        L1\citep{aggarwal2025l1}{,}
                        Efficient-Reasoning\citep{arora2025traininglanguagemodelsreason}{,}
                        Open-Reasoner-Zero\citep{OpenReasonerZero2025}{,}
                        Dr. GRPO\citep{DrGRPO}{,}
                        DAPO\citep{DAPO}{,}
                        Demystify-Long-CoT\citep{yeo2025demystifying}{,}
                        Light-R1\citep{wen2025light}
                        , leaf,fill=RL3]
                ]
                [Sampling Strategy~(\S\ref{sec:rl_sampling}), level2,fill=RL2
                    [\textit{e.g.,}
                        Light-R1\citep{wen2025light}{,}
                        Open-Reasoner-Zero\citep{OpenReasonerZero2025}{,}
                        SRPO\citep{zhang2025srpocrossdomainimplementationlargescale}{,}
                        DAPO\citep{DAPO}{,}
                        Skywork-OR1\citep{skywork-or1-2025}
                        , leaf,fill=RL3]
                ]
            ] 
            [Other Techniques~(\S\ref{sec:others}), level1,fill=Other1
                [Process-Reward-based RL~(\S\ref{sec:other_reward}), level2,fill=Other2
                    [\textit{e.g.,}
                        IPR\citep{xiong2024watch}{,}
                        PRMBench\citep{song2025prmbench}{,}
                        FineReason\citep{chen2025finereason}
                        rStar-Math\citep{guan2025rstar}{,} 
                        PRIME\citep{cui2025process}
                    , leaf,fill=Other3]
                ]
                [Preference Optimization~(\S\ref{sec:other_preference_optimization}), level2,fill=Other2
                    [\textit{e.g.,}
                        EXAONE\citep{research2025exaone}{,}
                        Light-R1\citep{wen2025light}{,}
                        DPO-VP\citep{tu2025enhancing}{,}
                        RedStar\citep{xu2025redstar}{,}
                        DPO-R1\citep{zhang2025dpor1}
                    , leaf,fill=Other3]
                ]
            ]
        ]
    \end{forest}

}

\caption{Taxonomy of training methods of reasoning models.}
\label{fig:paper-taxonomy}
\end{figure*}

This survey aims to provide a clear review of the open-source replication works on DeepSeek-R1.  
According to Figure~\ref{fig:paper-taxonomy}, the arrangement of this survey is based on methodology and generally corresponds to the training process of DeepSeek-R1, introducing current replication works on SFT, RLVR, and other technologies enhancing the reasoning capability. 
In introducing the conclusions made by these works, this survey attempts to summarize the common practice of replicating RLMs with comparative analysis from various perspectives, including data resources, sampling strategies, and training configurations.
With the above efforts, we aim to help researchers optimize their own models by effectively referencing the prior works.
The following sections are arranged as follows:
\begin{itemize}
\item 
    \textbf{Supervised Fine-tuning for Reasoning Language Models.} 
    We provide a comprehensive overview of replication works aimed to enhance the reasoning ability of language models through supervised fine-tuning. 
    Recognizing that the starting checkpoints and fine-tuning data resources are the key aspects for the SFT process, we conduct comparative analyses of these aspects to derive meaningful insights.
    We also summarize the common training practices for supervised fine-tuning. 
\item 
    \textbf{Reinforcement Learning from Verifiable Rewards for Reasoning Language Models.} 
    We present recent works that train RLMs using reinforcement learning from verifiable rewards (RLVR) by elaborating on their training data, learning algorithms, and reward system designs.
    Noting that various studies have adopted variants of PPO \citep{PPO} or GRPO \citep{GRPO} with subtle modifications, we attempt to establish a unified theoretical framework to explain these methods, highlighting both the algorithmic changes and the underlying motivations behind each adaptation.
    Moreover, based on the configurations and experiment results of these works, we conclude the possibly common practice for RLVR.
\item 
    \textbf{More Directions for Reasoning Language Models.} 
    We identify that while DeepSeek-R1 advances the training of RLMs, many supervision strategies remain unexplored. We present more directions for RLMs, including reward modeling and preference optimization and examine the strengths and weaknesses of current RLMs, such as their powerful out-of-distribution generalization and occasional overthinking. Finally, we briefly discuss extending RLMs to multimodal and multilingual applications.
    
\end{itemize}

\section{Supervised Fine-tuning for Reasoning Language Models}
\label{sec:sft}

DeepSeek-R1 distilled models \citep{Deepseek_R1}, e.g., the DeepSeek-R1-Distill-Qwen series, exhibit strong reasoning capabilities despite their smaller sizes. Since then, several works \citep{openr1, openthoughts, wen2025light, zhao202514millionopensourcedistilled} have attempted to reproduce this reasoning ability in smaller models by applying Supervised Fine-Tuning (SFT) on their own curated datasets. These datasets typically consist of math or coding problems and, more importantly, include one or more validated Chain-of-thoughts (CoTs) 
from DeepSeek-R1. This section aims to provide a comprehensive overview of how these studies approach the reproduction of distilled reasoning models.

\subsection{SFT Datasets}
\label{sec:sft_datasets}

In this subsection, we provide a comprehensive overview of datasets used for SFT, starting with their curation processes, followed by detailed descriptions of individual datasets. We also examine key properties such as token length distributions, data contamination risks, and cross-dataset dependencies, with the goal of highlighting best practices and common patterns in constructing high-quality reasoning datasets.

\subsubsection{Data Collection and Curation Pipeline}
A growing number of reasoning-focused datasets have been curated based on a shared set of principles aimed at improving reasoning capability through SFT. Most efforts begin by collecting questions across diverse domains, such as math, science, coding, and puzzles, either from existing benchmarks or web crawling. 

After raw data collection, multiple rounds of filtering are typically employed to enhance quality. These include deduplication (e.g., via embedding similarity or n-gram), rejection sampling, and ground-truth verification. To ensure high coverage and data richness, many datasets explicitly emphasize difficulty and diversity during the selection process, often using heuristics or model pass rates to prioritize harder problems. For example, Light-R1 \citep{wen2025light} applies thresholding based on model correctness to form a challenging subset from a broader base. Further, most datasets rely on verified CoTs or solutions to ensure correctness and quality. Verification methods vary by domain. For instance, math problems are often validated by Math Verify \citep{math_verify}, coding questions through execution or unit tests, and general tasks by LLM judges. This combination of domain-aware validation and selective retention allows curators to distill high-quality reasoning traces that better support supervised fine-tuning.

While these datasets span multiple domains, the majority are mainly focused on math and coding tasks {as observed in Table~\ref{tab1:sft_datasets_details}}. Broader coverage across diverse reasoning tasks, such as science, logic puzzles, and open-ended questions, remains relatively limited. Notable exceptions include DeepSeek-R1 and AM \citep{zhao202514millionopensourcedistilled}, which incorporate a wider range of domains during both data collection and distillation, aiming to foster more generalizable reasoning capabilities.

\subsubsection{Existing Dataset Details}

\begin{table}[ht]
\centering
\resizebox{\textwidth}{!}{%
\begin{tabular}{lccccc}
\toprule
\textbf{Project} & \textbf{Size of SFT Data} & \textbf{Math} & \textbf{Code} & \textbf{Other-Reasoning} & \textbf{Non-Reasoning}  \\
\midrule
DeepSeek-R1~\citep{Deepseek_R1} 
& 800k
& \faCheck  & \faCheck & \faCheck & \faCheck \\

AM~\citep{zhao202514millionopensourcedistilled}
& 1.4M
& \faCheck  & \faCheck & \faCheck & \faCheck \\

Synthetic-1~\citep{2025synthetic1} 
& 894k
& \faCheck  & \faCheck & \faCheck & \faTimes \\

OpenThoughts~\citep{openthoughts} 
& 114k 
& \faCheck  & \faCheck & \faCheck & \faTimes \\

Bespoke-Stratos~\citep{bespoke_stratos} 
& 16.7k
& \faCheck  & \faCheck & \faCheck & \faTimes \\

Light-R1~\citep{wen2025light} 
& 76k / 3.6k 
& \faCheck  & \faTimes & \faTimes & \faTimes \\

Open-R1~\citep{openr1} 
& 220k
& \faCheck  & \faTimes & \faTimes & \faTimes \\

S1k-1.1~\citep{muennighoff2025s1simpletesttimescaling} 
& 1k
& \faCheck  & \faTimes & \faCheck & \faTimes \\

LIMO~\citep{ye2025limoreasoning} 
& 817 
& \faCheck  & \faTimes & \faTimes & \faTimes \\
\bottomrule
\end{tabular}%
}
\caption{Summary of recent projects including SFT data and their corresponding categories. Other-Reasoning includes science, puzzles, etc.
}
\label{tab1:sft_datasets_details}
\end{table}

\paragraph{DeepSeek-R1.} \citet{Deepseek_R1} curates a distillation dataset of 800k training samples, comprising 600k reasoning examples and 200k non-reasoning examples, such as writing, role-playing, and other general-purpose tasks. According to the available report, parts of the non-reasoning data appear to be reused from the SFT dataset of DeepSeek-V3 \citep{deepseekai2024deepseekv3}. To create the distillation dataset, DeepSeek-R1 itself is used to generate the distillation traces. However, this interpretation is based on limited details provided, as the exact methodology has not been fully disclosed. Notably, the dataset is not publicly available. 

\paragraph{AM.}  \citet{zhao202514millionopensourcedistilled}
curates a large-scale reasoning dataset\footnote{\url{https://huggingface.co/datasets/a-m-team/AM-DeepSeek-R1-Distilled-1.4M}} comprising 1.4M samples across various domains. The curation mainly involves three stages: (i) Raw data collection, where they collect raw data from multiple data sources. (ii) Comprehensive data filtering, \textit{i.e.,} deduplication via embeddings-based similarity, upsampling of difficult problems using an LLM. (iii) CoT distillation. For samples lacking reasoning traces or whose ground truths fail verification, new CoTs are generated using DeepSeek-R1. To ensure correctness of the answer,
a sequence of verifiers is applied, including Math Verify and Qwen2.5-7B-Instruct \citep{qwen2.5}. Code-related problems with test cases are additionally verified through execution.

\paragraph{Synthetic-1.} \citet{2025synthetic1} constructs an 894k-sample reasoning dataset\footnote{\url{https://huggingface.co/datasets/PrimeIntellect/SYNTHETIC-1-SFT-Data}} distilled from DeepSeek-R1, covering domains such as math, coding, and STEM. Verification is domain-specific: Math Verify is used for math questions, execution-based validation is applied for coding problems, and LLM judges assess the remaining type of problems.

\paragraph{OpenThoughts.} OpenThoughts\footnote{\url{https://huggingface.co/datasets/open-thoughts/OpenThoughts-114k}} \citep{openthoughts} curates a synthetic reasoning dataset with 114k examples from several sources. It covers multiple domains, including math, science, coding, and puzzles. The CoTs are generated by DeepSeek-R1 and verified.
In particular, they use an LLM as a judge to verify the ground-truth answers for math and puzzle problems, and rely on code execution and unit tests to validate coding problems.

\paragraph{Bespoke Stratos.} \citet{bespoke_stratos} curates a reasoning dataset\footnote{\url{https://huggingface.co/datasets/bespokelabs/Bespoke-Stratos-17k}} of 17k examples distilled from DeepSeek-R1, covering domains such as coding, math, science, and puzzles. They apply rejection sampling to eliminate reasoning traces with incorrect solutions. 
In particular, they use GPT-4o-mini \citep{openai2024gpt4o} as a judge to filter out the traces with incorrect answers of math questions, increasing the proportion of retained examples from 25\% to 73\%, compared to a rule-based approach.

\paragraph{Light-R1.} Light-R1 \citep{wen2025light} constructs a high-quality SFT dataset\footnote{\url{https://huggingface.co/datasets/qihoo360/Light-R1-SFTData}} comprising 79k samples distilled from DeepSeek-R1. They begin by collecting 1 million math problems from various sources and use DeepScaleR-1.5B-Preview \citep{deepscaler2025} to generate initial responses. Only questions with low pass rates (below $\alpha$) are selected for further processing with DeepSeek-R1, yielding around 76k examples. From this set, only correct long-form CoT responses are retained, with one chosen per question to create an SFT dataset of over 70k examples, filtered for both difficulty and diversity.
While training solely on this dataset was effective in reproducing the distilled model, a second stage was introduced to further enhance quality
by leveraging  DeepSeek-R1-Distill-Qwen-32B.
This phase retains only those questions with pass rates of DeepSeek-R1-Distill-Qwen-32B below $\alpha$  and where DeepSeek-R1's responses were not consistently correct or incorrect. The result is a refined Stage 2 SFT dataset containing approximately 3k examples.

\paragraph{Open-R1.} OpenR1-Math-220k\footnote{\url{https://huggingface.co/datasets/open-r1/OpenR1-Math-220k}} \citep{openr1} is a large-scale dataset for math reasoning tasks. \citet{openr1} collects 220k math problems from NuminaMath 1.5 \citep{numina_math_datasets} and generate 2 to 4 CoTs for each problem using DeepSeek-R1. To ensure that each problem includes at least one correct answer, most of the CoTs are verified by Math Verify \citep{math_verify}, with Llama-3.3-70B-Instruct \citep{dubey2024llama3herdmodels} serving as a judge for 12\% of the samples.
Among the 220k problems, 94k are considered higher quality. According to \citet{openr1}, the 94k subset achieves better performance in SFT, as the 131k extended problems may contain easier questions.

\paragraph{S1k-1.1.} \citet{muennighoff2025s1simpletesttimescaling} curates a large-scale reasoning dataset\footnote{\url{https://huggingface.co/datasets/simplescaling/s1K-1.1}} by collecting 59k questions from 16 diverse sources. Each question is paired with a reasoning trace and a solution generated by DeepSeek-R1, forming question-trace-solution triplets. After decontamination and deduplication, a three-stage filtering process produces a high-quality, diverse, and challenging subset of 1,000 samples, designed for minimal-resource training.

\paragraph{LIMO.} While not directly focused on reproducing distilled DeepSeek-R1 models, LIMO\footnote{\url{https://huggingface.co/datasets/GAIR/LIMO}} \citep{ye2025limoreasoning} still offers valuable insights into reasoning model development.  LIMO first constructs tens of millions of problems from various established datasets, such as NuminaMath. They then apply a baseline difficulty filter using Qwen2.5-Math-7B-Instruct \citep{qwen2.5}, removing problems that can be solved within a few attempts.
Next, they collect reasoning traces from human experts and state-of-the-art models, including DeepSeek-R1, DeepSeek-R1-Distill-Qwen-32B, and Qwen2.5-32B-Instruct, and conduct a thorough analysis to identify key characteristics of high-quality reasoning chains, namely, Optimal Structural Organization, Effective Cognitive Scaffolding, and Rigorous Verification.
Finally, they use a hybrid approach combining rule-based filtering with LLM-assisted curation to select high-quality solutions for each question. The resulting dataset of 817 questions demonstrates strong performance when used to fine-tune base models.

\begin{figure}
    \centering
    \includegraphics[width=1\linewidth]{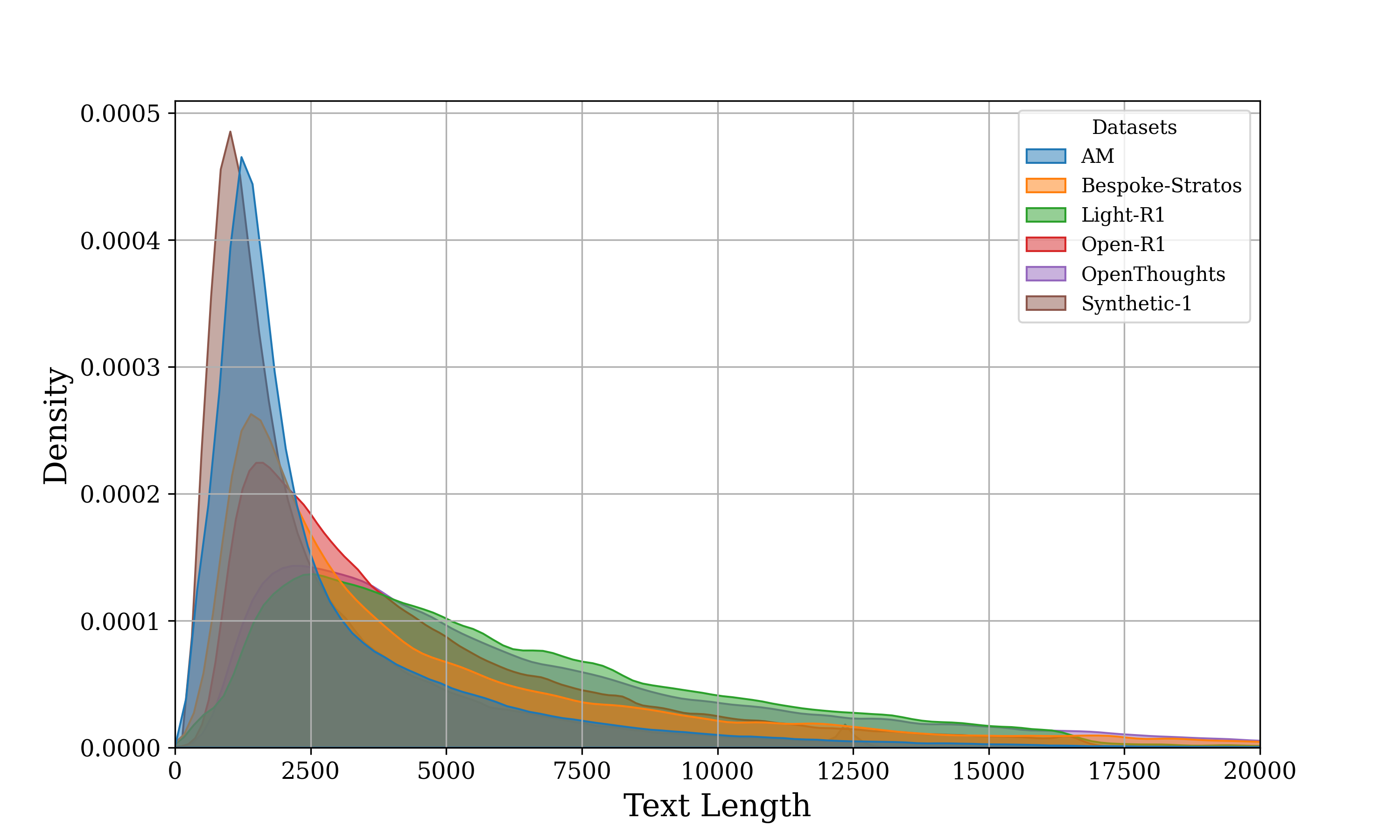}
    \caption{Token length distributions for the aforementioned SFT datasets. The x-axis is truncated at 20,000 tokens, as examples exceeding this length are rare.}
    \label{fig:len_dist}
\end{figure}

\subsubsection{Analysis and Discussions}

\paragraph{Length Distributions.} Figure~\ref{fig:len_dist} illustrates the token length distributions of the datasets discussed above. Although all the long CoTs of these datasets originate from the same teacher model, \textit{i.e.,} DeepSeek-R1, their distributions exhibit observable
differences. 
For instance, datasets such as AM and Synthetic-1 are skewed toward shorter sequences, whereas Light-R1 and Open-R1 display broader distributions with longer tails, suggesting a higher proportion of complex problems, which typically elicit longer CoTs.

\paragraph{Data Decontamination.} Among all the works discussed above, only the technical reports of Light-R1 and LIMO explicitly mention conducting proper data decontamination against popular reasoning benchmarks, \textit{e.g.,} AIME24/25, MATH500, and GPQA Diamond \citep{rein2024gpqa}, during dataset curation. Notably, \citet{wen2025light} point out that MATH500 is partially compromised across several open-source datasets, including OpenThoughts, Open-R1, Bespoke Stratos, and others.

\paragraph{Cross-Referencing Dataset Sources.}
Recently, a growing number of math reasoning datasets have been released to support the SFT of LLMs. However, many of these datasets are not created in isolation, \textit{i.e.,} they frequently collect or derive data from preexisting datasets, often with overlapping or reused examples. In some cases, multiple datasets originate from the same root sources, making it challenging to understand their relationships and evaluate model dataset quality fairly. For example, many datasets have obtained problems from NuminaMath.

\begin{figure}
    \centering
    \includegraphics[width=1.15\linewidth]{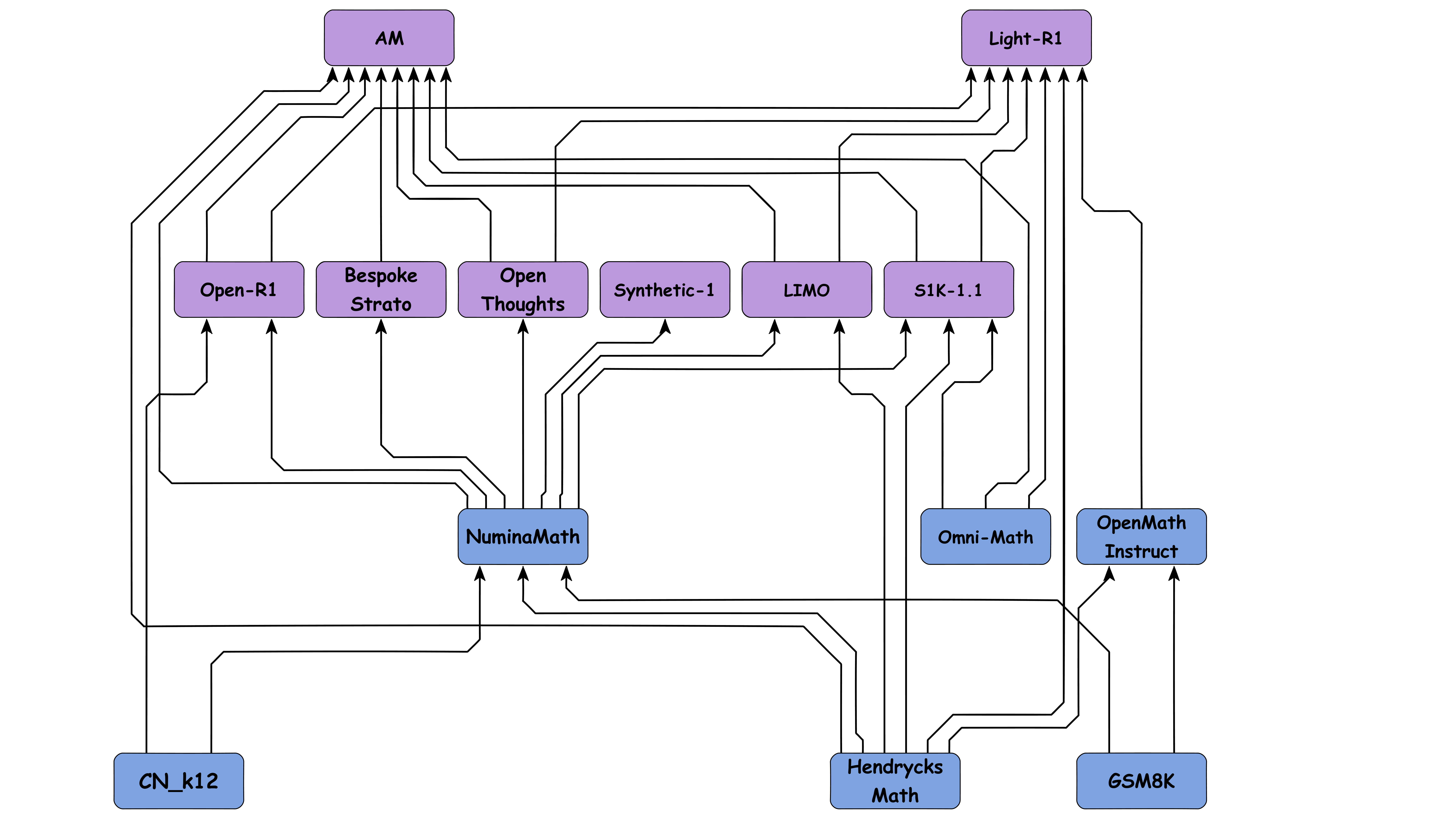}
   \caption{An illustration of cross-referenced dataset sources for popular math reasoning datasets. Arrows point from source datasets to target datasets that incorporate some of their data. The figure does not reflect dataset sizes, nor does it imply that a target dataset includes all data from its source, or only data from the source(s) indicated by the arrows. Datasets highlighted in lilac contain Chain-of-Thought traces extracted from DeepSeek-R1. 
   }
   \label{fig:dataset-cross-ref}
\end{figure}

To clarify these connections and reduce confusion, we present Figure~\ref{fig:dataset-cross-ref}, which illustrates the cross-referencing structure among popular math reasoning datasets. It highlights the web of dependencies and shared data across benchmarks, helping researchers better interpret results and avoid redundant training or evaluation setups. 

\subsection{Training \& Performance Comparison}
\label{sec:sft_models}

In this subsection, we first formalize the supervised fine-tuning (SFT) of reasoning language models,
and then provide a detailed overview of the configuration used in the current replication studies for SFT.

\paragraph{Supervised Fine-tuning.}
Given a dataset $\mathcal{D}_{\text{SFT}} \triangleq \left\{ (q_i, c_i) \right\}_{i=1}^{|\mathcal{D}|}$, where each sample $(q_i, c_i)$ consists of a question $q_i$ and a long CoT $c_i$. The long CoT can be further decomposed into a complex intermediate rationale followed by a final answer. SFT updates the parameters of the policy model $\pi_{\theta}$ by minimizing the negative log-likelihood loss: 
\begin{align}
\mathcal{L}_{\text{SFT}}(\theta) \triangleq - \mathbb{E}_{(q, c) \sim \mathcal{D}_{\text{SFT}}} \left[ \log \pi_{\theta}(c \mid q) \right],
\end{align}
where $\pi_{\theta}(c \mid q)$ denotes the probability assigned by the policy to the CoT response $c$ conditioned on the question $q$. 
This objective encourages the model to imitate the supervised demonstrations by maximizing the likelihood of the reference completions.

\begin{table}[ht]
\centering
\resizebox{\textwidth}{!}{%
\begin{tabular}{llccc}
\toprule
\textbf{Project} & \textbf{Initial Checkpoint} & \textbf{AIME24} & \textbf{AIME25} & \textbf{MATH500} \\
\midrule
DeepSeek-R1~\citep{Deepseek_R1} 
& Qwen2.5-Math-7B / Qwen2.5-32B-Base
& 55.5 / 72.6 & -- & 92.8 / 94.3 \\

AM~\citep{zhao202514millionopensourcedistilled}
& Qwen2.5-32 / 72B-Base
& 72.7 / 76.5 & --  & 96.2 / 97.0 \\

Light-R1~\citep{wen2025light} 
& Qwen2.5-32B-Instruct 
&  73.0 
&  64.3 
& -- \\

LIMO~\citep{ye2025limoreasoning} 
& Qwen2.5-32B-Instruct  
& 57.1 & 44.5 & 94.8 \\

S1k-1.1~\citep{muennighoff2025s1simpletesttimescaling} 
& Qwen2.5-32B-Instruct  
& 56.7 & 50.0 & 94.4 \\

OpenThoughts~\citep{openthoughts} 
& Qwen2.5-7 / 32B-Instruct
& 31.3 / 68.0 
& 23.3 / 49.3 
& 83.2 / 90.6 \\

Bespoke-Stratos~\citep{bespoke_stratos} 
& Qwen2.5-7 / 32B-Instruct 
& 20.0 / 63.3 
& --
& 82.0 / 93.0 \\

Open-R1~\citep{openr1} 
& Qwen2.5-Math-7B-Instruct 
& 36.7 & 40.0 & 90.6 \\

Synthetic-1~\citep{2025synthetic1} 
& Qwen-2.5-7B-Instruct
& 30.0 & 26.6 & 85.6 \\

\bottomrule
\end{tabular}%
}
\caption{Summary of recent projects including initial checkpoints and their corresponding benchmark results. Results are taken from corresponding papers. Dashes (--) indicate unavailable results. Note that the reported Open-R1 performance is from the ``default'' split of the dataset.}
\label{tab2:sft_datasets_base_models}
\end{table}

\paragraph{Comparison.}
In practice, SFT stage plays a crucial role in allowing the base model to learn high-quality reasoning traces from stronger models. 
Table~\ref{tab2:sft_datasets_base_models} presents a comparative overview of SFT results on common math reasoning benchmarks, AIME24/25 and MATH500 \citep{hendrycks2021measuringmathematicalproblemsolving}, highlighting the impact of different dataset choices and initial checkpoints.

While many approaches emphasize scaling up the number of training samples to boost performance, LIMO and S1k-1.1 demonstrate that strong results can be achieved with significantly smaller, carefully curated datasets. 
In particular, only the DeepSeek-R1-distilled series and AM fine-tune  from Qwen2.5-base models, while other methods rely on the stronger Qwen2.5-Instruct models.
A related study by \citet{li2025small} also indicates that instruct models exhibit higher learning efficiency than their base counterparts. 

Furthermore, DeepSeek uniquely incorporates non-reasoning data for SFT. 
By contrast, other works focus primarily on math and coding reasoning problems, leaving the interplay between reasoning and non-reasoning data underexplored.

\paragraph{Training Details.}

Although the technical report of DeepSeek-R1 does not mention the training hyperparameters of the distillation models, we aggregate this information from replication studies to better understand common training setups. 
For long-context tasks such as complex reasoning, the RoPE scaling factor (\(\theta\)) and maximum context length in the model configuration are often adjusted to enable extended context capabilities~\citep{chen2023extending}. 
For example, Open-R1~\citep{openr1} sets \(\theta = 300{,}000\) and the context length to 32,768 tokens. 
Commonly used learning rates include \(1.0 \times 10^{-5}\) and \(5.0 \times 10^{-5}\), with typical batch sizes of 96 or 128.
Additionally, packing is usually employed to improve training efficiency \citep{wang2024packinganalysispackingappropriate}.
\section{Reinforcement Learning from Verifiable Rewards for Reasoning Language Models}
\label{sec:rl}

This section focuses on reinforcement learning from verifiable rewards (RLVR) for reasoning language models.  
First, we provide a detailed examination of techniques for training reasoning language models with RLVR, including training data preparation, reinforcement learning (RL) algorithms, and the designs of reward systems and data sampling strategies during training.  
Specifically, for RL algorithms and their variants, we provide an in-depth discussion of the motivation and rationale behind each of them. 
According to the implementation details and experiment results from the replication studies, we summarize the key insights from several aspects.  
We also introduce ongoing efforts to extend reasoning language models beyond closed-book examinations on scientific subjects, adapting them to a broader range of tasks. 

\subsection{RL Datasets}
\label{sec:rl_data}

DeepSeek-R1-Zero achieved strong performance on reasoning and knowledge tasks through a standalone RLVR process. 
The curated high-quality datasets employed during its RLVR process are instrumental to the success.
Therefore, several replication studies have explored strategies for efficiently creating training datasets by leveraging open-source data and powerful models.
In this subsection, we introduce the datasets used in RLVR. These datasets cover various tasks that are verifiable during RL training, in which we mainly focus on datasets for math and coding problem solving.
We introduce the curation of each dataset, including the selection of data resources, the construction of verified questions and answers, and the detailed pre-processing procedures. 
Table~\ref{tab:rl_datasets_stats} displays an overview for the statistics of these datasets. 

\begin{table}[ht]
\centering
\resizebox{\textwidth}{!}{%
\begin{tabular}{lccc}
\toprule
\textbf{Dataset} & \textbf{Organization} & \textbf{Size} & \textbf{Categories}  \\
\midrule

DeepScaleR~\citep{deepscaler2025}
& Agentica Project
& 40k
& Math \\

Skywork-OR1~\citep{skywork-or1-2025}
& Skywork
& 129k
& Math, Code \\

Open-Reasoner-Zero~\citep{OpenReasonerZero2025}
& StepFun
& 129k
& Math, Reasoning \\

Big Math~\citep{albalak2025bigmathlargescalehighqualitymath}
& SynthLabs
& 251k
& Math \\

DeepMath-103k~\citep{he2025deepmath103klargescalechallengingdecontaminated}
& Tencent
& 103k
& Math \\

Curated Thoughts~\citep{curatedthoughts}
& Bethge Lab, University of Tuebingen
& 222k
& Math \\

DAPO-Math-17k~\citep{DAPO}
& ByteDance Seed
& 17k
& Math \\

LIMR~\citep{li2025limr}
& GAIR, Shanghai Jiao Tong University
& 1k
& Math \\

Math-RLVR~\citep{su2025crossingrewardbridgeexpanding}
& Tencent AI Lab
& 773k
& Math \\

SYNTHETIC-1~\citep{2025synthetic1}
& Prime Intellect
& 144k
& Code \\

DeepCoder~\citep{deepcoder2025}
& Agentica Project
& 24k
& Code \\

Open-R1-CodeForces~\citep{penedo2025codeforces}
& Hugging Face
& 10k
& Code \\

KodCode-V1~\citep{xu2025kodcodediversechallengingverifiable}
& Microsoft GenAI
& 484k
& Code \\

Code-r1-12k~\citep{code-r1}
& University of Illinois Urbana-Champaign
& 12k
& Code \\

Multi-Subject-RLVR~\citep{su2025crossingrewardbridgeexpanding}
& Tencent AI Lab
& 638k
& General \\

II-Thought-RL-v0~\citep{2025iithought}
& Intelligent-Internet
& 342k
& General \\

\bottomrule
\end{tabular}%
}
\caption{Verified open-source off-the-shelf datasets curated for RL training and their corresponding categories. The statistics for SYNTHETIC-1 denotes the size of its subset of algorithmic coding problems. }
\label{tab:rl_datasets_stats}

\end{table}

\paragraph{DeepScaleR-Preview.}

DeepScaleR \citep{deepscaler2025} collected around 40k unique contest-level math problems from AIME (1984-2023), AMC (prior to 2023), Omni-MATH \citep{gao2024omnimathuniversalolympiadlevel} and Still \footnote{\url{https://github.com/RUCAIBox/Slow_Thinking_with_LLMs}} datasets. 
They extract the answer for each problem using gemini-1.5-pro-002, removing duplicate questions, and filtering out unverifiable samples.

\paragraph{Skywork-OR1.}

Skywork-OR1 \citep{skywork-or1-2025} is trained on math and code tasks during the RL phase.
The data resource of math is generally similar to DeepScaleR \citep{deepscaler2025}, with including extra challenging problems from NuminaMath-1.5 \citep{numina_math_datasets}. 
The code training data is from LeetCode \citep{xia2025leetcodedatasettemporaldatasetrobust} and TACO \citep{li2023taco}. 
In preprocessing, Skywork-OR1 performed an elaborated verification of data samples. 
Skywork-OR1 removed all instances with external URLs or potential figures to verify the validity of problems. 
Math samples are verified using Math-Verify \citep{math_verify}, while code samples are required to include a complete set of unit test cases, with its solution passing all corresponding tests.
After preprocessing, deduplication is performed, resulting in a total of 105k math samples and 13.7k code samples.  
Skywork-OR1 also marked the difficulty level of each sample based on its pass rate when evaluated by DeepSeek-R1-Distilled models.

\paragraph{Open-Reasoner-Zero.}

Open-Reasoner-Zero \citep{OpenReasonerZero2025} identifies three key aspects for data curation: quantity, diversity, and quality. 
Open-Reasoner-Zero collects a total of 129k training data, of which 72k are mainly cleaned from OpenR1-Math-220k \citep{openr1}, and the rest are collected from various sources, including AIME (up to 2023), MATH, Numina-Math collection and Tulu3 MATH.
It also leverages additional synthetic data using programmatic approaches to cover other reasoning domains. 
In controlling data quality, Open-Reasoner-Zero filters out samples with unverifiable formats, such as multiple-choice and proof-oriented problems. 
Also, it filters out non-English samples for better training stability and final model performance. 
Additionally, Open-Reasoner-Zero selects a challenging subset of 13k samples from the complete dataset using a intermediate model checkpoint during training, which is used to support the curriculum learning on difficulty of this work, aiming to address the shortcomings of the model and enhancing its performance on challenging scenarios.

\paragraph{Big-Math.}

Big-Math \citep{albalak2025bigmathlargescalehighqualitymath} includes a massive amount of high-quality samples with open-ended problems and uniquely verifiable closed-form solutions. 
Each sample is categorized by its mathematics domain (eg. sequences and series).
Additionally, this dataset provides a new source of 47,000 problems deriving from multiple-choice problems, namely Big-Math-Reformulated. 
Big-Math performs a very strict filtering and cleaning process to ensure the quality of data samples. 
First, a strict deduplication based on exact matching and semantic similarity is performed, together with a test set decontamination using MATH-500 and Omni-MATH test sets.
Second, possibly invalid samples are removed, including problems with hyperlinks and problems that are unsolvable in 8 rollouts from Llama-3.1-405B or 64 rollouts from Llama-3.1-8B. 
Third, possibly unverifiable samples are removed, including problems with hyperlinks, multiple choice problems, yes/no and true/false problems, multi-part questions, questions asking for a proof, and non-English problems. 
Last, miscellaneous unnecessary information (eg. problem scoring) are cleaned from data samples. 

\paragraph{DeepMath-103K.}

DeepMath-103K \citep{he2025deepmath103klargescalechallengingdecontaminated} is collected to serve as a credible and challenging training data resource while ensuring no contamination with existing benchmarks. 
From a difficulty distribution estimation of current source datasets, MMIQC and WebInstructSub are selected as the resources of DeepMath-103K as these datasets are sourced more broadly from web content and contain more challenging questions. 
Then, a strict decontamination against common benchmarks is performed to ensure the integrity of this datasets.
Moreover, each samples is rated with its difficulty by prompting GPT-4o based on the annotation guidelines provided by the Art of Problem Solving \citep{aops}, and is verified through a rigorous two-stage process. 
In question formatting and filtering, problem types inherently unsuitable for verification (e.g., proofs) were discarded, and questions phrased conversationally were automatically rewritten into a standardized format that seeks a single, specific numerical or symbolic answer.
In answer verification via consistency check, three distinct solution paths are generated for each sample and only samples where all paths extract identical final answers are retained in the final dataset.
The procedure ensures that every problem included in DeepMath-103K possesses a final answer that is robustly verifiable using automated rules.

\paragraph{Other Datasets for Math Problem Solving.}

CuratedThoughts \citep{curatedthoughts} is curated from prevailing SFT datasets, filtering out improper samples to support stable RL training. 
It is collected from OpenR1-Math-220k \citep{openr1}, OpenThoughts-114k and OpenThoughts-114k-Math \citep{openthoughts}, removing multi-part questions, questions asking for a proof, questions referring to figures or charts, and questions without a valid answer. 
DAPO-Math-17k \citep{DAPO} modifies the questions from the AoPS website \citep{aops} so that the expected answer would always be an integer. The simple answers making them easy to parse to minimize errors from math verifiers, providing accurate reward signals during RL training.
LIMR \citep{li2025limr} proposes the learning impact measurement to filter a small amount of samples whose learning patterns complement the model’s overall performance trajectory, demonstrating that these samples tend to be more valuable for optimization. 

\paragraph{Coding Problem Solving Datasets. }

A verified sample for coding problem solving should contain a number of unit tests covering the typical cases, boundary conditions, exceptional or invalid inputs, and performance extremes, as well as ensuring full coverage of all branches, conditions, and their combinations.
The algorithmic coding problems subset of SYNTHETIC-1 \citep{2025synthetic1} is curated from Apps, Codecontests, Codeforces and TACO datasets.
LLM-based post-processing is applied to additionally translate Python problems into Javascript, Rust and C++ problems, resulting in a total of 144k samples.
DeepCoder-Preview \citep{deepcoder2025} examined popular coding data resources and filtered out easy samples and unverifiable samples with noisy questions or responses, or flawed or missing test cases.
They have chosen verified samples from TACO \footnote{\url{https://huggingface.co/datasets/likaixin/TACO-verified}}, SYNTHETIC-1 and LiveCodeBench (v5, May 1, 2023 - July 31, 2024) as their train set, and LiveCodeBench (v5, August 1, 2024 - February 1, 2025) as their test set. 
It is ensured that all problems within this dataset are fully verifiable and have no less than 5 test cases.
Open-R1-CodeForces \citep{penedo2025codeforces} collects more than 10k unique samples with the solutions and unit tests validated from the very first contests of CodeForces all the way to 2025. 
KodCode-V1 \citep{xu2025kodcodediversechallengingverifiable} is a fully-synthetic dataset by collecting and rewriting questions from 12 distinct resources and generating the solutions, test cases, and difficulty levels by DeepSeek-R1. 
Code-r1-12k \citep{code-r1} consists of 2K LeetCode samples with generally reliable test cases and 10K verified samples from TACO.

\paragraph{General Domain Datasets. }

It is an exciting idea to expand the RL training paradigm to more domains than formatted tasks such as math and coding problem solving. 
To this aim, \citet{su2025crossingrewardbridgeexpanding} proposes Math-RLVR and Multi-Subject-RLVR which are annotated with free-form reference answers. 
The prediction accuracy on these datasets during training are verified by LLMs. 
Math-RLVR consists of 773k samples covering three educational levels: elementary, middle, and high school. 
Multi-Subject-RLVR consists of 638k college-level samples written by domain experts for examination purposes, extracted from ExamQA \citep{yu2021self} which covers at least 48 first-level subjects.
Similarly, II-Thought \citep{2025iithought} proposes a comprehensive dataset consists of 342k samples covering math, science, code and riddle domains.

\subsection{RL Components}
\label{sec:rl_obj}

With the release of DeepSeek-R1-Zero and DeepSeek-R1, DeepSeek showcases its success in fine-tuning LLMs for complex reasoning tasks through RL.
Building on carefully curated training data, replication studies have focused on configuring key aspects of the RL framework to achieve competitive performance: the adoption of effective RL algorithms (e.g., GRPO), and the design of reward systems.
Some studies have also explored advanced data sampling strategies to further boost performance.
This subsection reviews representative efforts in fine-tuning reasoning language models with RL from the above aspects, highlighting their key contributions from a conclusive perspective.
Table \ref{tab:rl_datasets_base_models} provides a comparative view of the methodology for the mentioned studies.

\begin{table}[h]
\centering
\resizebox{\textwidth}{!}{%
\begin{tabular}{llcccccc}
\toprule
\textbf{Model} & \textbf{Initial Checkpoint} & \textbf{Data Size} & \textbf{RL} & \textbf{Reward} \\
\midrule

DeepSeek-R1~\citep{Deepseek_R1}
& DeepSeek-V3-Base       
& -- 
& GRPO
& Accuracy, Format \\

DeepSeek-R1-Zero~\citep{Deepseek_R1}
& DeepSeek-V3-Base       
& --
& GRPO
& Accuracy, Format \\

\hdashline

VAPO~\citep{yuyue2025vapoefficientreliablereinforcement-vapo}
& Qwen2.5-32B-Base   
& --
& VAPO
& Accuracy \\

VC-PPO~\citep{VC-PPO}
& Qwen2.5-32B-Base   
& --
& VC-PPO
& Accuracy \\

Open-Reasoner-Zero-32B~\citep{OpenReasonerZero2025}
& Qwen2.5-32B-Base   
& 129k
& PPO
& Accuracy \\

SRPO~\citep{zhang2025srpocrossdomainimplementationlargescale}
& Qwen2.5-32B-Base   
& --
& SRPO
& Accuracy, Format \\

DAPO~\citep{Deepseek_R1}
& Qwen2.5-32B-Base         
& 17k
& DAPO
& Accuracy, Length \\

Skywork-OR1-32B-Preview~\citep{skywork-or1-2025}
& DeepSeek-R1-Distill-Qwen-32B
& 105k
& GRPO
& Accuracy, Format\\

Light-R1-14B-DS~\citep{wen2025light}
& Light-R1-14B-DS-SFT 
& -- 
& GRPO
& Accuracy, Length \\

Logic-RL~\citep{xie2025logicrlunleashingllmreasoning}
& Qwen2.5-7B-Instruct-1M
& 5k
& REINFORCE++
& Accuracy, Format  \\

Qwen2.5-7B-SimpleRL-Zero~\citep{zeng2025simplerl} 
& Qwen2.5-Math-7B   
& 8k
& PPO
& Accuracy, Format \\

Oat-Zero-7B~\citep{DrGRPO} 
& Qwen2.5-Math-7B   
& --
& Dr. GRPO
& Accuracy \\

MiMo-7B-RL-Zero~\citep{xiaomi2025mimo} 
& MiMo-7B-Base  
& 130k
& GRPO
& Accuracy \\

Mini-R1~\citep{mini-r1}
& Qwen2.5-3B-Instruct 
& 50k
& GRPO
& Accuracy, Format \\

TinyZero~\citep{tinyzero}
& Qwen2.5-3B-Base 
& --
& PPO
& Accuracy, Format \\

DeepScaleR-1.5B-Preview~\citep{deepscaler2025}
& Deepseek-R1-Distilled-Qwen-1.5B
& 40k
& GRPO
& Accuracy, Format \\

GPG-1.5B~\citep{chu2025gpg}
& Deepseek-R1-Distilled-Qwen-1.5B
& --
& GPG
& Accuracy, Format \\

\bottomrule
\end{tabular}%
}
\caption{An overview of the algorithm selction and reward design of competitive open-source DeepSeek-R1 replication studies on RLVR. Models from DeepSeek-R1 series \citep{Deepseek_R1} are separately listed for comparison. Dashes (--) indicate unavailable numbers.} 
\label{tab:rl_datasets_base_models}
\end{table}

\subsubsection{Algorithms}
\label{sec:rl_algorithm}

As the most prevalent outcome-reward-based RL methods, PPO \citep{PPO} and GRPO \citep{GRPO} are widely used for fine-tuning LLMs.  
Interestingly, recent replication studies have introduced various modifications to these methods, tailoring them for specific purposes to enhance training effectiveness. We review several representative RL-based LLM fine-tuning algorithms, including REINFORCE \citep{reinforce_LLM}, PPO \citep{PPO}, GRPO \citep{GRPO}, and their variants \citep{DrGRPO,DAPO,VC-PPO,reinforce++,KIMI_scaling_RL, lin2025cppo}. 
Moreover, we outline the modifications in these methods along with their underlying motivations, aiming to provide a clear overview of the methodological advancements in outcome-reward-based RL training methods.

\paragraph{LLM Policy Optimization.} Recent studies have introduced a groundbreaking post-training paradigm that enhances LLMs' reasoning capabilities through RL-based training. In this framework, the LLM's answer generation process for each query is formulated as an answer sampling policy, and our objective is to optimize this LLM policy to maximize the expected reward of the generated responses. According to \citet{Deepseek_R1,OpenReasonerZero2025,KIMI_scaling_RL}, large-scale RL-based LLM policy optimization enables the base LLM to achieve a steady improvement in reasoning accuracy while also exhibiting the emergence of long-chain reasoning in its chain-of-thought.

Suppose each reasoning data pair $(q,a)$ is i.i.d sampled from an underlying distribution $\mathcal{D}$, where each $q$ is a query and $a$ is the corresponding ground-truth answer. Let $\pi_{\theta}(\cdot | \cdot)$ be the target LLM policy parameterized by $\theta$. The expected reward of the LLM on a sample $(q,a)$ is $\mathbb{E}_{o\sim \pi_{\theta}(\cdot | q)} [r(o, a)]$, where $o$ is an LLM-generated response to $q$, and $r(\cdot,\cdot)$ is a predefined reward function that quantifies whether the response $o$ yields $a$. The objective of RL-based fine-tuning is to maximize the expected reward over the data distribution, i.e., 
\begin{align}
\max_{\theta}
    J(\pi_{\theta})
    \triangleq
    \mathbb{E}_{(q,a)\sim \mathcal{D}}
    \mathbb{E}_{o\sim \pi_{\theta}(\cdot|q)}
    [r(o, a)].
\end{align}
A straightforward approach to maximize $J(\pi_{\theta})$ is to gradually improve the LLM's parameter $\theta$ towards the policy gradient direction $\nabla_{\theta} J(\pi_{\theta})$. However, since $ \nabla_{\theta} \mathbb{E}_{o \sim \pi_{\theta}(\cdot | q)} r(o, a) $ is the gradient of an integral dependent on $ \pi_\theta $, $ \nabla_{\theta} J(\pi_{\theta}) $ is intractable to compute via standard Monte Carlo sampling. Fortunately, the RL community has developed two powerful policy gradient estimators: REINFORCE \citep{reinforce} and Importance Sampling \citep{Sutton_RL}:
\begin{align}
    \nabla_{\theta}
    \mathbb{E}_{o\sim \pi_{\theta}(\cdot|q)} r(o, a)
    =
    \begin{cases}      
        \mathbb{E}_{o \sim \pi_{\theta}(\cdot|q)}
        \left[
        \nabla_{\theta} \log \pi_{\theta}(o|q) \cdot r(o, a) 
        \right]\ &\text{(REINFORCE)},
        \\
        \mathbb{E}_{o \sim \pi_{\theta'}(\cdot|q)}
        \left[
        \nabla_{\theta}
        \left(
        \frac{\pi_{\theta}(o|q)}{\pi_{\theta'}(o|q)}
        \right)
        \cdot
        r(o, a)
        \right]\ &\text{(Importance Sampling)},
    \end{cases}
    \label{eq:rl_policy_grad}
\end{align}
where $\pi_{\theta'}$ is any parameter-frozen LLM policy. Hence, the policy gradient $\nabla_{\theta} J(\pi_\theta)$ can be effectively approximated using standard Monte Carlo sampling: for each data pair $(q,a)$, we independently generate $G$ responses to $q$, denoted by $\{o_i\}_{i=1}^G$, using the current LLM $\pi_\theta$ or the frozen LLM $\pi_{\theta'}$, and then approximate the policy gradient estimators by
\begin{small}
\begin{align}   
\nabla_{\theta} J(\pi_\theta)
=
\begin{cases}
\mathbb{E}_{(q,a)\sim \mathcal{D},\{o_i\}_{i=1}^G\sim \pi_{\theta}(\cdot|q)}
\left[
\frac{1}{G}
\sum_{i=1}^G
\nabla_{\theta} \log \pi_{\theta}(o_i|q) \cdot r(o_i, a) 
\right]
&\text{(REINFORCE)},
\\
\mathbb{E}_{(q,a)\sim \mathcal{D},\{o_i\}_{i=1}^G\sim \pi_{\theta'}(\cdot|q)}
\left[
\frac{1}{G}
\sum_{i=1}^G
\nabla_{\theta}
\left(
\frac{\pi_{\theta}(o_i|q)}{\pi_{\theta'}(o_i|q)}
\right)
\cdot
r(o_i, a)
\right]
& \text{(Importance Sampling)},
\end{cases}
\label{eq:rl_policy_grad_estimator}
\end{align}
\end{small}

For each query $q$, the procedure of generating $G$ independent responses $\{o_{i}\}_{j=1}^G$ from $\pi_{\theta}(\cdot|q)$ is called the `rollout phase'. During this phase, the LLM policy explores enormous response samples of varying quality. Then  $\theta$ is updated to increase the likelihood $ \pi_{\theta}(o_i|q) $ where $ r(o_i, a)$ is large, thereby improving the likelihood of generating responses with high rewards. Specifically, REINFORCE is an on-policy method that requires generating new rollouts using the latest LLM policy $ \pi_{\theta} $. In contrast, the importance sampling estimator can be implemented in an off-policy manner with improved sampling efficiency, as it can reuse past rollouts generated from $\pi_{\theta'}$ by storing the corresponding probability terms $ \pi_{\theta'}(o_i | q)$. A common choice is to implement $\pi_{\theta'}$ as $\pi_{\theta_{\mathrm{old}}}$, a past snapshot of the target LLM $\pi_\theta$, which is updated periodically.

In practice, the reward signals $\{r(o_i, a)\}_{i=1}^G$ are highly sparse, leading to high variance in rollout phases and policy gradient estimation. To mitigate these issues, various techniques have been developed to stabilize LLM policy gradient estimation in (\ref{eq:rl_policy_grad_estimator}). These techniques generally fall into three categories: 1) reducing sampling variance by reward normalization or using actor-critic advantage estimation, 2) stabilizing parameter updates by clipping the importance sampling weight $ \pi_\theta(o_i|q) / \pi_{\theta_{\mathrm{old}}}(o_i|q)$, and 3) constraining policy shifts by penalizing the KL-divergence $\mathrm{KL}(\pi_{\theta} | \pi_{\mathrm{ref}})$ between the current LLM policy $\pi_{\theta}$ and a fixed reference LLM policy $\pi_{\mathrm{ref}}$. 

\paragraph{PPO.} Since its introduction in \citet{PPO}, Proximal Policy Optimization (PPO) has become one of the most popular actor-critic RL algorithms for LLM policy optimization \citep{RLHF, OpenReasonerZero2025}. In addition to the target LLM policy $\pi_\theta$, which serves as the actor model, PPO introduces a critic model $V_{\phi} $—another LLM designed to learn the value for the responses generated by the actor LLM $\pi_\theta$. Specifically, the PPO objective is
\begin{equation}
\begin{aligned}
J_{\mathrm{PPO}}(\pi_\theta) \triangleq 
&\ \mathbb{E}_{(q, a) \sim \mathcal{D},\left\{o_i\right\}_{i=1}^G \sim \pi_{\theta_{\text{old}}}(\cdot \mid q)} 
\\
\Bigg[
&
\frac{1}{G} 
\sum_{i=1}^G
\frac{1}{\left| o_i \right|} 
\sum_{t=1}^{\left|o_i\right|} 
\left(
\min \Big( r_{i, t}(\theta) \textcolor{red}{\hat{A}_{i, t}(\phi)}, 
 \textcolor{red}{\operatorname{clip}}\left(r_{i, t}(\theta), 1 - \varepsilon, 1 + \varepsilon\right) \textcolor{red}{\hat{A}_{i, t}(\phi)} \Big) 
 \right)
 \Bigg],
\end{aligned}
\end{equation}
where $r_{i,t}(\theta)\triangleq \pi_{\theta}(o_{i,t}|q,o_{i,<t})/ \pi_{\theta_{\mathrm{old}}}(o_{i,t}|q,o_{i,<t})$ denotes the likelihood ratio between the current LLM policy $\pi_\theta$ and the past LLM policy $\pi_{\theta'}$ calculated on the $t$-th token prediction step; $\hat{A}_{i, t}(\phi)$ denotes the Generalized Advantage Estimator (GAE) \citep{GAE} computed using the estimated value $V_{\phi}(o_{i,t}|q,o_{i,<t})$, which estimates the quality of each response generation state. $ V_{\phi} $ is trained along with $\pi_{\theta}$ to predict the value of the response generated by $\pi_{\theta}$. In practice \citep{OpenReasonerZero2025}, GAE is observed to be a more robust response quality estimator than the raw reward $r(q_i,a^*_i)$, leading to more stable LLM policy optimization.

\paragraph{GRPO.} Group Relative Policy Optimization (GRPO) is first proposed \citep{Deepseek_R1} as an effective and efficient variant of PPO. Specifically, GRPO discards the critic model and GAE calculation in PPO to improve efficiency and memory consumption. To reduce the reward sampling variance, GRPO normalizes the rewards within a group of $G$ rollout outs. In addition to clipping the likelihood ratio terms, GRPO further introduces KL-divergence penalty to ensure that $\pi_\theta$ would not be driven far away from the initial SFT LLM. Specifically, the GRPO objective is 
\begin{small}
\begin{equation}
\begin{aligned}
J_{\mathrm{GRPO}}(\pi_\theta) \triangleq 
&\ \mathbb{E}_{(q, a) \sim \mathcal{D},\left\{o_i\right\}_{i=1}^G \sim \pi_{\theta_{\text{old}}}(\cdot \mid q)} 
\\
\Bigg[
&
\frac{1}{G} 
\sum_{i=1}^G 
\frac{1}{\left| o_i \right|} 
\sum_{t=1}^{\left|o_i\right|} 
\left(
\min \Big( r_{i, t}(\theta) \textcolor{red}{\hat{A}_{i, t}}, 
 \operatorname{clip}\left(r_{i, t}(\theta), 1 - \varepsilon, 1 + \varepsilon\right) \textcolor{red}{\hat{A}_{i, t}} \Big) 
 -
 \textcolor{red}{\beta \mathrm{KL}(\pi_{\theta}|\pi_{\mathrm{ref}})}_{i,t}
 \right)
 \Bigg],
\end{aligned}
\end{equation}
\end{small}

where $\hat{A}_{i,t} \triangleq (r(o_i,a) - \mathrm{mean}(\{r(o_i,a)\}_{i=1}^G))/\mathrm{std}(\{r(o_i,a)\}_{i=1}^G)$ denotes the group relative reward, and $\mathbf{r} \triangleq \{r(o_i,a)\}_{i=1}^G$ denotes the rewards of the response group corresponding to each sample $(q,a)$. GRPO also incorporates the K3 KL-divergence estimator \citep{KL_K3}:
\begin{align}
    \mathrm{KL}(\pi_{\theta}|\pi_{\mathrm{ref}})_{i,t}
    \triangleq 
    \frac{\pi_{\mathrm{ref}}(o_{i,t}|q,o_{i,<t})}{\pi_{\theta}(o_{i,t}|q,o_{i,<t}))} - \log \frac{\pi_{\mathrm{ref}}(o_{i,t}|q,o_{i,<t})}{\pi_{\theta}(o_{i,t}|q,o_{i,<t}))} - 1.
\end{align} 
DeepSeek-R1 \citep{Deepseek_R1} shows that GRPO achieves stable large-scale LLM policy optimization that incentivizes the long CoT pattern in large-scale LLMs.

\paragraph{REINFORCE++.} REINFORCE++ \citep{reinforce++} stabilizes the policy gradient update by incorporating token-wise KL-divergence penalty into the reward function. Like GRPO, REINFORCE++ normalized the penalized rewards within each rollout group. Formally, the training objective is
\begin{equation}
\begin{aligned}
J_{\mathrm{REINFORCE++}}(\pi_\theta) \triangleq 
&\ \mathbb{E}_{(q, a) \sim \mathcal{D},\left\{o_i\right\}_{i=1}^G \sim \pi_{\theta_{\text{old}}}(\cdot \mid q)} 
\\
\Bigg[
&
\frac{1}{G} 
\sum_{i=1}^G 
\frac{1}{\left| o_i \right|} 
\sum_{t=1}^{\left|o_i\right|} 
\left(
\min \Big( r_{i, t}(\theta) \textcolor{red}{\hat{A}_{i, t}}, 
 \operatorname{clip}\left(r_{i, t}(\theta), 1 - \varepsilon, 1 + \varepsilon\right) \textcolor{red}{\hat{A}_{i, t}} \Big)
 \right)
 \Bigg],
\end{aligned}
\end{equation}
where $\hat{A}_{i,t} \triangleq (\hat{r}_i - \mathrm{mean}(\{\hat{R}_i\}_{i=1}^G))/\mathrm{std}(\{\hat{R}_i\}_{i=1}^G)$, and $\hat{R}_i$ is the penalized reward defined as 
\begin{align}
    \hat{R}_{i,t}
    \triangleq
    r(o_i,a) - 
    \beta 
    \sum_{j=t}^{\left| o_i\right|}
    \log
    \left(
    \frac{\pi_{\theta_{\mathrm{old}}}(o_{i,j}|q,o_{i,<j})}{\pi_{\theta_{\mathrm{ref}}}(o_{i,j}|q,o_{i,<j})}
    \right).
\end{align}

\paragraph{REINFORCE on LLM.} \citet{KIMI_scaling_RL} shows that REINFORCE-like policy gradient can achieve stable training on 72B LLMs. Compared to the basic REINFORCE in (\ref{eq:rl_policy_grad_estimator}), \citet{KIMI_scaling_RL} employs centralized rewards and adds K2 KL-divergence penalty \citep{KL_K3}. This yields the following modified REINFORCE policy gradient
\begin{equation}
\begin{aligned}
\nabla_{\theta}J_{\mathrm{KIMI}}(\pi_\theta)\triangleq 
&\ \mathbb{E}_{(q, a) \sim \mathcal{D},\left\{o_i\right\}_{i=1}^G \sim \pi_{\theta_{\text{old}}}(\cdot \mid q)} 
\\
\Bigg[
&
\frac{1}{G} 
\sum_{i=1}^G 
\frac{1}{\left| o_i \right|} 
\sum_{t=1}^{\left|o_i\right|} 
\left(
\nabla_{\theta} 
\log \pi_{\theta}(o_i|q) (r(o_i,a) - \textcolor{red}{\bar{r}})
-
\textcolor{red}{
\frac{\beta}{2}
\nabla_{\theta}
\left(
\log
\frac{\pi_{\theta}(o_i|q)}{\pi_{\theta_{\mathrm{old}}}(o_i|q)}
\right)^2
}
 \right)
 \Bigg],
\end{aligned}
\end{equation}
where $\bar{r}\triangleq \mathrm{mean}(\{r(o_i,a)\}_{i=1}^G)$ denotes the average reward among each rollout group.

\paragraph{DAPO.} \citet{DAPO} identifies several critical shortcomings in the original GRPO algorithm, including entropy collapse, training instability, and biased loss.
Entropy collapse refers to the rapid decline in policy entropy during training, where the sampled responses for certain prompts become nearly identical, reducing diversity.
Additionally, the algorithm suffers from a gradient decreasing issue: when some prompts consistently achieve perfect accuracy, the resulting zero advantage leads to inefficient and unstable training.
Moreover, the original sample level loss computation, where token level losses are first averaged within each sample and then aggregated across samples, introduces a length bias, as tokens in longer responses contribute less to the overall loss.
To address these issues, the authors propose the Decoupled Clip and Dynamic sAmpling Policy Optimization (DAPO) algorithm:
\begin{equation}
\begin{aligned}
J_{\mathrm{DAPO}}(\pi_\theta) \triangleq 
&\ \mathbb{E}_{(q, a) \sim \mathcal{D},\left\{o_i\right\}_{i=1}^G \sim \pi_{\theta_{\text{old}}}(\cdot \mid q)} \\\Bigg[ 
&\quad \frac{1}{\textcolor{red}{\sum_{i=1}^G \left| o_i \right|}} \textcolor{red}{\sum_{i=1}^G \sum_{t=1}^{\left|o_i\right|}} \min \Big( r_{i, t}(\theta) \hat{A}_{i, t}, 
 \operatorname{clip}\left(r_{i, t}(\theta), 1 - \textcolor{red}{\varepsilon_{\text{low}}}, 1 + \textcolor{red}{\varepsilon_{\text{high}}}\right) \hat{A}_{i, t} \Big) \Bigg] \\
\text{subject to} \quad 
&\ \textcolor{red}{0 < \left| \left\{ o_i \,\middle|\, \text{is\_equivalent}(a, o_i) \right\} \right| < G},
\end{aligned}
\end{equation}
where $\hat{A}_{i,t} \triangleq (r(o_i,a) - \mathrm{mean}(\{r(o_i,a)\}_{i=1}^G))/ \mathrm{std}(\{r(o_i,a)\}_{i=1}^G)$.
Specifically, $\varepsilon$ is decoupled into $\varepsilon_{\text{low}}$ and $\varepsilon_{\text{high}}$, with $\varepsilon_{\text{high}}$ set to a higher value to allow more room for increasing low-probability tokens, helping to address entropy collapse issues.
The dynamic sampling constraint $0 < \left| \left\{ o_i \,\middle|\, \text{is\_equivalent}(a, o_i) \right\} \right| < G$ ensures that rollouts from a given prompt contain both correct and incorrect outputs, resulting in nonzero advantages that improve training efficiency and stability.
Finally, the average is computed over all tokens instead of the original sample level loss. This encourages longer sequences to make greater contributions to the overall gradient update, which is critical in long CoT RL scenarios.

\paragraph{Dr. GRPO.} Similarly, \citet{DrGRPO} reveals two biases in the original GRPO algorithm: a response level length bias and a question level difficulty bias. The response level length bias arises from dividing by $|o_i|$, which encourages the policy to favor shorter correct responses while preferring longer incorrect ones. The question level difficulty bias comes from normalizing the centered outcome reward by the standard deviation $\text{std}(\{R_i\}_{i=1}^G$, which causes questions with lower variance in rewards, typically those that are either too easy or too hard, to receive greater weight during policy updates. To address these biases, they propose GRPO Done Right (Dr. GRPO):
\begin{small}
\begin{equation}
\begin{aligned}
    J_{\mathrm{Dr. GRPO}}(\pi_{\theta})
    &\ \triangleq \mathbb{E}_{(q, a) \sim \mathcal{D},\left\{o_i\right\}_{i=1}^G \sim \pi_{\theta_{\text{old}}}(\cdot \mid q)}\\
    &\
    \Bigg[ 
    \frac{1}{G}
    \sum_{i=1}^G
    \frac{1}{\textcolor{red}{\cancel{|o_i|}}}
    \sum_{t=1}^{|o_i|}
    \Big(
    \min
    \left(
    r_{i,t}(\theta)\hat{A}_{i,t},
    \mathrm{clip}
    (r_{i,t}(\theta), 1-\varepsilon, 1+ \varepsilon) \hat{A}_{i,t}
    \right)
    -
    \beta 
    \mathrm{KL}(\pi_{\theta}|\pi_{\mathrm{ref}})
    \Big)
    \Bigg],\\
\text{where} \quad &\ \hat{A}_{i,t} \triangleq (r(o_i,a) - \mathrm{mean}(\{r(o_i,a)\}_{i=1}^G))/ \textcolor{red}{\cancel{\mathrm{std}(\{r(o_i,a)\}_{i=1}^G)}}.
\end{aligned}
\end{equation} 
\end{small}
By eliminating both the normalization terms $1/|o_i|$ and $1/\mathrm{std}(\{r(o_i,a)\}_{i=1}^G)$, Dr. GRPO is able to enhance token efficiency without compromising reasoning performance.

\paragraph{CPPO.}  \citet{lin2025cppo} highlights a key limitation of GRPO: Although effective, it demands substantial computational resources due to the need for sampling multiple completions per query. To address this issue, they propose Completions Pruning Policy Optimization (CPPO): 
\begin{small}
\begin{equation}
\begin{aligned}
    J_{\mathrm{CPPO}}(\pi_{\theta})
    &\ \triangleq \mathbb{E}_{(q, a) \sim \mathcal{D},\left\{o_i\right\}_{i=1}^G \sim \pi_{\theta_{\text{old}}}(\cdot \mid q)}\\
    &\
    \left[ 
    \textcolor{red}{
    \frac{1}{|\mathcal{I}|}
    \sum_{i\in\mathcal{I}}
    }
    \frac{1}{|o_i|}
    \sum_{t=1}^{|o_i|}
    \Big(
    \min
    \left(
    r_{i,t}(\theta)\hat{A}_{i,t},
    \mathrm{clip}
    (r_{i,t}(\theta), 1-\varepsilon, 1+ \varepsilon) \hat{A}_{i,t}
    \right)
    -
    \beta 
    \mathrm{KL}(\pi_{\theta}|\pi_{\mathrm{ref}})
    \Big)
    \right],
\end{aligned}
\end{equation}
\end{small}
where $\mathcal{I}\triangleq \{i\in \{1,\dots, G\}\ |\ |\hat{A}_{i,t}| \geqslant \gamma)\}$, $\gamma$ is a predefined threshold that filters out low-impact completions, and $\hat{A}_{i,t} \triangleq (r(o_i,a) - \mathrm{mean}(\{r(o_i,a)\}_{i=1}^G))/ \mathrm{std}(\{r(o_i,a)\}_{i=1}^G)$. This ensures that only completions (i.e. rollout samples) with sufficiently high absolute advantage contribute to the policy update.
As a result, CCPO accelerates the training process by skipping the forward pass and gradient backpropagation on rollout samples with low advantages.

\paragraph{GPG.}
\citet{chu2025gpg} proposes Group Policy Gradient (GPG), a REINFORCE-based method that removes complex components and directly optimizes the true objective, bypassing the use of surrogate losses:
\begin{equation}
\begin{aligned}
    J_{\mathrm{GPG}}(\pi_{\theta})
    \triangleq \mathbb{E}_{(q, a) \sim \mathcal{D},\left\{o_i\right\}_{i=1}^G \sim \pi_{\theta}(\cdot \mid q)}
    \Bigg[ 
    \frac{1}{\sum_{i=1}^{G}|o_i|}
    \sum_{i=1}^G
    \sum_{t=1}^{|o_i|}
    (
    -\log\pi_\theta(o_{i,t}|q, o_{i,<t})\hat{A}_{i,t}
    )
    \Bigg],
\end{aligned}
\end{equation}
where $\hat{A}_{i,t} \triangleq (r(o_i,a) - \mathrm{mean}(\{r(o_i,a)\}_{i=1}^G))/ \mathrm{std}(\{r(o_i,a)\}_{i=1}^G)$. This eliminates the need for the implementation of the critic model and the reference model, offering significant advantages for scalability in distributional training.

\paragraph{VC-PPO.} \citet{VC-PPO} identifies two critical failure modes of PPO in long CoT reasoning: value initialization bias and reward signal decay. The former issue stems from initializing the value model with a reward model trained only on \texttt{<EOS>} tokens, resulting in a position-dependent advantage bias that favors shorter completions. The latter issue is caused by the trace decay rate $\lambda < 1$ used in reward propagation during GAE computation \citep{GAE}, which severely weakens the reward signal for earlier tokens in long sequences. To address these issues, they propose Value-Calibrated PPO (VC-PPO), which introduces two modifications:
1) value pretraining, where the value model is pretrained under a fixed SFT policy using Monte Carlo returns (i.e., setting $\lambda=1$) to eliminate the initial value bias;
2) decoupled-GAE, which uses different GAE trace decay $\lambda$ values for policy and value updates, allowing the critic model to achieve unbiased advantage estimation by setting $\lambda=1$, and enabling the actor model to achieve variance reduction in advantage estimation by setting $\lambda=0.95$. This two-pronged calibration significantly improves PPO’s stability and performance on long-form reasoning benchmarks like AIME24.

\paragraph{VAPO.} \citet{yuyue2025vapoefficientreliablereinforcement-vapo} introduces Value-based Augmented Proximal Policy Optimization (VAPO), a value-based RL framework outperforming value-free methods in long Chain-of-Thought reasoning. VAPO adopts Value-Pretraining and Decoupled-GAE from VC-PPO \citep{VC-PPO} to mitigate value model bias. To handle heterogeneous sequence lengths, VAPO uses token-level loss from DAPO \citep{DAPO} and employs the Length-Adaptive GAE trace decay rate $\lambda_{\text{policy}} \triangleq 1 - 1/(\alpha l)$, where $\alpha$ is a scaling hyperparameter and $l$ denotes the length of each rollout sample. To address reward sparsity, the VAPO objective employs the Clip-Higher trick proposed in DAPO and integrates an additional negative log-likelihood (NLL) penalty:
\begin{small}
\begin{align}
J_{\mathrm{VAPO}}(\pi_{\theta})
\triangleq &
J_{\mathrm{PPO-CH}}(\pi_{\theta}) + \textcolor{red}{
\mu 
J_{\mathrm{NLL}}(\pi_{\theta})},
\\
J_{\mathrm{PPO-CH}}(\pi_{\theta})
\triangleq &
\
\mathbb{E}_{(q, a) \sim \mathcal{D},\left\{o_i\right\}_{i=1}^G \sim \pi_{\theta_{\text{old}}}(\cdot \mid q)} 
\notag
\\
\Bigg[ 
&\quad \frac{1}{\sum_{i=1}^G \left| o_i \right|} \sum_{i=1}^G \sum_{t=1}^{\left|o_i\right|} \min \Big( r_{i, t}(\theta) \hat{A}_{i, t}(\phi) , 
 \operatorname{clip}\left(r_{i, t}(\theta), 1 - \textcolor{red}{\varepsilon_{\text{low}}}, 1 + \textcolor{red}{\varepsilon_{\text{high}}}\right) \hat{A}_{i, t}(\phi) \Big) \Bigg],
 \\
J_{\mathrm{NLL}}(\pi_{\theta})
\triangleq &
\mathbb{E}_{(q, a) \sim \mathcal{D},\left\{o_i\right\}_{i=1}^G \sim \pi_{\theta_{\text{old}}}(\cdot \mid q)}
\left[
-
\frac{1}{\sum_{i\in \mathcal{T}}|o_i| }
\sum_{i\in \mathcal{T}}
\sum_{t=1}^{|o_i|}
\log \pi_{\theta}(o_{i,t}|q,o_{i,<t})
\right],
\end{align}
\end{small}
with $\hat{A}_{i,t}(\phi)$ denoting the GAE \citep{GAE} estimated by the critic model $V(\phi)$, $\mathcal{T}$ denoting the set of rollout indices that achieve correct answers and $\mu>0$ denoting the penalty rate. In essence, the introduced NLL penalty is interpreted to perform SFT over the correct rollout samples. Furthermore, VAPO adopts the Group Sampling technique proposed in GRPO \citep{GRPO} to effectively generate discriminative positive and negative samples within the same prompt context. Collectively, these integrated approaches make VAPO a robust benchmark for effective and stable long-form reasoning within value-based RL frameworks.

\subsubsection{Rewards}
\label{sec:rl_reward}

Rewards are the cornerstone of RL training, as they define the optimization objective and guide the model’s behavior. A well-designed reward provides clear and consistent signals that help the agent learn effective policies. However, reward models are often prone to reward hacking \citep{amodei2016concrete, everitt2017reinforcement, weng2024reward}, prompting recent research to favor rule-based outcome reward systems. These typically fall into three categories:

\paragraph{Accuracy Rewards.} Accuracy rewards evaluate whether a response is correct, typically assigning a score of 1 for correct answers and 0 or -1 for incorrect ones. They are widely regarded as the most fundamental type of reward and are sometimes used exclusively, reflecting a minimalist approach to reward design \citep{OpenReasonerZero2025, DrGRPO}.

\paragraph{Format Rewards.} Format rewards encourage responses to follow predefined structures or reasoning formats, typically rewarding correct formatting with 1 and penalizing deviations with 0 or -1. 
While they are intended to promote clarity and consistency, \citet{OpenReasonerZero2025} found that models trained solely with accuracy rewards, when guided by well-designed prompts, can still quickly learn and reinforce the desired formatting. This suggests that explicit format rewards may be unnecessary in some cases. Moreover, such rewards may inadvertently incentivize reward hacking behaviors \citep{OpenReasonerZero2025, xie2025logicrlunleashingllmreasoning}.

\paragraph{Length Rewards.} These rewards influence the verbosity of the model’s output. Some approaches reward generating responses of a desired length \citep{aggarwal2025l1}, while others incentivize brevity without sacrificing accuracy \citep{arora2025traininglanguagemodelsreason}. \citet{DAPO} implement a linear length penalty when responses exceed a predefined maximum length, whereas \citet{yeo2025demystifying} propose a cosine-based reward that encourages longer reasoning processes for incorrect answers and more concise ones for correct responses, which is also adopted by \citet{wen2025light}.

\subsubsection{Sampling Strategies}
\label{sec:rl_sampling}

Intuitively, properly selecting samples during training is crucial for effective RL training.
On the one hand, curriculum learning methods that gradually increase task difficulty during training improve the utility of difficult samples. Several studies have adopted these strategies to boost the training process. 
Open-Reasoner-Zero \citep{OpenReasonerZero2025} leveraged a two-step curriculum learning process to efficiently use difficult samples.
SRPO \citep{zhang2025srpocrossdomainimplementationlargescale} adopts a two-stage training paradigm to bridge the gap between math and coding tasks: the first stage emphasizes mathematical data to develop step-by-step reasoning, followed by coding data to build on this foundation, enabling progressive improvement in both domains.
On the other hand, the proper use of rejection sampling techniques could improve the general sample efficiency and stabilize training. 
Light-R1 \citep{wen2025light} implements a broader two-sided weight clipping mechanism for importance sampling. 
This mechanism limit the influence of extreme values to stabilize the training process. 
DAPO \citep{DAPO} and Skywork-OR1 \citep{skywork-or1-2025} adopt dynamic sampling which filters out zero advantage sample groups to increase sample efficiency and training stability, according to their claims that these groups do not contribute to the policy loss, but may contribute to the KL loss or entropy loss, leading to a more unstable training process. 
Besides, Skywork-OR1 \citep{skywork-or1-2025} and SRPO \citep{zhang2025srpocrossdomainimplementationlargescale} introduce epoch-level history resampling strategies that drops samples with all correctly predicted samples in last epoch to focus training on harder cases, enhancing learning efficiency.
MiMo \citep{xiaomi2025mimo} further argues that such a strategy introduces instability in policy updates, and develops an easy data resampling strategy, managing to improve sampling efficiency without risking policy collapse.
This strategy maintains an easy data pool during training where problems with perfect pass rates are stored, and samples data from this pool with a 10\% probability when performing rollouts. 

\subsection{Analysis and Discussions}

\begin{table}[h]
\centering
\resizebox{\textwidth}{!}{%
\begin{tabular}{llccc}
\toprule
\textbf{Model} & \textbf{Initial Checkpoint} & \textbf{AIME24} & \textbf{AIME25} & \textbf{MATH500} \\
\midrule

DeepSeek-R1~\citep{Deepseek_R1}
& DeepSeek-V3-Base
& 79.8* & -- & 97.3* \\

DeepSeek-R1-Zero~\citep{Deepseek_R1}
& DeepSeek-V3-Base
& 71.0* & -- & 95.9* \\

OpenAI o4 mini~\citep{openai-o3} 
& --
& 93.4* & 92.7* & -- \\

Seed-Thinking-v1.5~\citep{seed-thinking-1.5} 
& --
& 86.7* & 74.0* & -- \\

Qwen3-235B~\citep{qwen3} 
& --
& 85.7* & 81.5* & -- \\

\hdashline

VAPO~\citep{yuyue2025vapoefficientreliablereinforcement-vapo}
& Qwen2.5-32B-Base
& 60.4 & -- & -- \\

SRPO~\citep{zhang2025srpocrossdomainimplementationlargescale}
& Qwen2.5-32B-Base
& 50.0 & -- & -- \\

DAPO~\citep{Deepseek_R1}
& Qwen2.5-32B-Base
& 50.0 & -- & -- \\

VC-PPO~\citep{VC-PPO}
& Qwen2.5-32B-Base
& 48.8 & -- & -- \\

Open-Reasoner-Zero-32B~\citep{OpenReasonerZero2025}
& Qwen2.5-32B-Base
& 48.1 & 36.0 & 92.2 \\

DeepSeek-R1-Zero-Qwen-32B~\citep{Deepseek_R1}
& Qwen2.5-32B-Base
& 47.0* & -- & 91.6* \\

\hdashline

Skywork-OR1-32B-Preview~\citep{skywork-or1-2025} 
& DeepSeek-R1-Distill-Qwen-32B
& 79.7 & 69.0 & -- \\

DeepSeek-R1-Distill-Qwen-32B~\citep{Deepseek_R1} 
& Qwen2.5-32B-Base
& 72.6$^\dagger$ & -- & 94.3$^\dagger$ \\

\hdashline

Light-R1-14B-DS~\citep{wen2025light} 
& Light-R1-14B-DS-SFT 
& 74.0 & 60.2 & -- \\

DeepSeek-R1-Distill-Qwen-14B~\citep{Deepseek_R1} 
& Qwen2.5-Math-14B
& 69.7$^\dagger$ & -- & 93.9$^\dagger$ \\

\hdashline

Skywork-OR1-Math-7B~\citep{skywork-or1-2025} 
& DeepSeek-R1-Distill-Qwen-7B
& 69.8 & 52.3 & -- \\

Skywork-OR1-7B-Preview~\citep{skywork-or1-2025} 
& DeepSeek-R1-Distill-Qwen-7B
& 63.6 & 45.8 & -- \\

Light-R1-7B-DS~\citep{wen2025light} 
& DeepSeek-R1-Distill-Qwen-7B
& 59.1 & 44.3 & -- \\

DeepSeek-R1-Distill-Qwen-7B~\citep{Deepseek_R1} 
& Qwen2.5-Math-7B
& 55.5$^\dagger$ & -- & 92.8$^\dagger$ \\

MiMo-7B-RL-Zero~\citep{xiaomi2025mimo}
& MiMo-7B-Base
& 56.4 & 46.3 & 93.6 \\

Oat-Zero-7B~\citep{DrGRPO}
& Qwen2.5-Math-7B
& 43.3 & -- & 80.0 \\

Qwen2.5-7B-SimpleRL-Zero~\citep{zeng2025simplerl}
& Qwen2.5-Math-7B
& 36.7 & -- & 77.4 \\

\hdashline

DeepScaleR-1.5B-Preview~\citep{deepscaler2025}
& Deepseek-R1-Distilled-Qwen-1.5B
& 43.1 & -- & 87.8 \\

GPG-1.5B~\citep{chu2025gpg}
& Deepseek-R1-Distilled-Qwen-1.5B
& 33.3 & -- & 87.6 \\

DeepSeek-R1-Distill-Qwen-1.5B~\citep{Deepseek_R1} 
& Qwen2.5-Math-1.5B
& 28.9$^\dagger$ & -- & 83.9$^\dagger$ \\
\bottomrule
\end{tabular}%
}
\caption{Performance on math reasoning tasks of competitive open-source DeepSeek-R1 replication studies on RLVR, where models trained from base models and other models are separately listed for better comparison.
Performance of popular proprietary RLVR models (marked with *) and R1-distilled models (marked with $^\dagger$) are also listed for better comparison. 
Dashes (--) indicate unavailable results. 
}
\label{tab:rl_math}
\end{table}

As displayed in Table \ref{tab:rl_math}, although DeepSeek-R1 \citep{Deepseek_R1} reported that smaller models (e.g., not larger than 32B) may fail to match the performance of distillation models through RL training, the community has actively sought solutions to this limitation. 
In this subsection, we present insights from several projects that attempt to replicate RLVR performance on LLMs ranging from 1.5B to 32B.

\subsubsection{Recipes of Training Data}

\paragraph{Quantity and Diversity. }
Quantity and diversity are emphasized as key aspects in training reasoning language models suitable for multiple tasks. 
Skywork-OR1 \citep{skywork-or1-2025}, Seed-Thinking-v1.5 \citep{seed-thinking-1.5}, MiMo \citep{xiaomi2025mimo} and Qwen3 series \citep{qwen3} have all proposed that they leverage massive RL data from various domains during training. 
Open-Reasoner-Zero \citep{OpenReasonerZero2025} leverages data synthesis and self-distillation to expand the training dataset.

\paragraph{Difficulty.}
Several works introduce their data preparation pipeline to construct challenging datasets for RL training, providing inspirations on difficulty rating.
Light-R1 \citep{wen2025light} and Skywork-OR1 \citep{skywork-or1-2025} conduct offline data selection that leverages a trained checkpoint to sample and verify responses for the query of each training sample, keeping only the samples with a moderate pass rate, and filtering out the samples with overly high or low pass rates, indicating that the corresponding queries are either too easy or too hard.
DeepScaleR \citep{deepscaler2025} revealed that samples with an overly high pass rate are too easy for model training, while samples with a zero pass rate are often unverifiable or contain errors, therefore, both should be filtered out.
Open-Reasoner-Zero \citep{OpenReasonerZero2025} further adopts this filtering strategy to construct training data from synthetic and distilled data that is more noisy.
KodCode \citep{xu2025kodcodediversechallengingverifiable} performs difficulty rating with an off-the-shelf LLM rather than a trained checkpoint of their own model. 
LIMR \citep{li2025limr} proposes the learning impact measurement to filter samples whose learning patterns complement the model’s overall performance trajectory, proving that these samples tend to be more valuable for training. 

\paragraph{Data Cleaning. }
As a fundamental step during data preparation, data cleaning is crucial to construct less noisy datasets for effective training. 
Especially, for RL training on reasoning tasks, several works emphasize the necessity to filter out unsolvable questions and unverifiable answers. 
DAPO \citep{DAPO} modifies the questions from the original samples, ensuring that the expected answers are always integers. The advantages of such simple answers is that they are easy to parse, minimizing errors generated by formula parsers, and providing accurate reward signals during RLVR. 
BigMath \citep{albalak2025bigmathlargescalehighqualitymath} conducts a very strict cleaning process to remove questions with hyperlinks, referring to figures or charts, containing multiple sub-questions, and non-English questions. 
It also removes questions being unsuitable for verification, including multiple-choice and mathematical proving questions. 
These strategies are also adopted by various works including Open-Reasoner-Zero \citep{OpenReasonerZero2025} and SRPO \citep{zhang2025srpocrossdomainimplementationlargescale}, in which MiMo \citep{xiaomi2025mimo} performs data cleaning via an off-the-shelf LLM. 

\paragraph{De-duplication and Decontamination.}
De-duplication and decontamination are also important in constructing RL training datasets. 
When collecting data from multiple sources, Light-R1 \citep{wen2025light} emphasized the necessity of decontamination to ensure fair evaluation, and DeepScaleR \citep{deepscaler2025} and Skywork-OR1 \citep{skywork-or1-2025} performed elaborated de-duplication for efficient training.

\paragraph{Curriculum Learning Based on Data Difficulty.}
The training process of Open-Reasoner-Zero \citep{OpenReasonerZero2025} adopts curriculum learning: the 32B model was initially trained for 1100 steps with data sampled from the full dataset, followed by the selection of a challenging 13k subset based on the model's success rate, which was then used to fine-tune the model for improved performance on the most difficult reasoning problems.

\paragraph{Overall.}
We observe a consistent trend: datasets used for RL training are carefully designed to include data where models are likely to make mistakes (i.e., models neither consistently succeed nor completely fail). Such uncertainty creates opportunities for learning.
This calibrated challenge level encourages models to engage in deeper reasoning and reflection, often resulting in longer and more informative responses.

\subsubsection{RL Algorithm Design}
\paragraph{REINFORCE, PPO, and GRPO.}
As discussed in Section~\ref{sec:rl_obj}, existing efforts primarily utilize algorithms such as REINFORCE \citep{reinforce}, PPO \citep{PPO}, GRPO \citep{GRPO}, and their variants.
For instance, Open-Reasoner-Zero \citep{OpenReasonerZero2025} adopts vanilla PPO instead of GRPO. Their empirical studies reveal that vanilla PPO offers a notably stable and robust training process when using GAE parameters of $\lambda=1.0$ and $\gamma=1.0$. 
Similarly, Logic-RL \citep{xie2025logicrlunleashingllmreasoning} reports that PPO ($\lambda=1.0$ and $\gamma=1.0$) significantly outperforms GRPO and REINFORCE++ \citep{reinforce++} in both accuracy and reward. Moreover, REINFORCE++ consistently surpasses GRPO across nearly all evaluation metrics, with GRPO demonstrating the weakest overall performance among the three RL algorithms. However, Logic-RL only trains on its K\&K text logic dataset, which may not generalize to math and coding domains.
Nonetheless, GRPO and its variants have also yielded promising results. Light-R1 \citep{wen2025light} reports substantial performance improvements using GRPO on DeepSeek-R1-Distill-Qwen-14B.

\paragraph{Variants.}
DAPO~\citep{DAPO} utilizes GRPO with clip-higher, dynamic sampling, and token-level policy gradient loss, and achieves similarly strong results on AIME 2024 using Qwen2.5-32B-Base.
In addition, MiMo-7B-RL-Zero \citep{xiaomi2025mimo} adopts GRPO with clip-higher and dynamic sampling to achieve effective training starting from a 7B base model. 
Dr. GRPO \citep{DrGRPO} demonstrates effectiveness on a 7B base model by introducing simple modifications to the original GRPO--namely removing the normalization terms for response length and advantage.

\paragraph{Overall.}
The community has made engineering efforts to adapt the algorithms for more stable training. However, these algorithms are not significantly different from the theoretical perspective.

\subsubsection{Model Size and Type}
\paragraph{Model sizes.}
Existing studies have demonstrated the effectiveness of RL across a wide range of model sizes, from 1.5B to 32B parameters.
Specifically, DeepScaleR \citep{deepscaler2025} surpasses o1-preview on AIME24 using DeepSeek-R1-Distill-Qwen-1.5B by scaling RL training.
Open-Reasoner-Zero \citep{OpenReasonerZero2025}, Logic-RL \cite{xie2025logicrlunleashingllmreasoning}, and Dr. GRPO \citep{DrGRPO} have all shown that RL is effective on 7B base models.
Light-R1~\citep{wen2025light} further demonstrates that GRPO yields strong performance on DeepSeek-R1-Distill-Qwen-14B.
In addition, both Open-Reasoner-Zero~\citep{OpenReasonerZero2025} and DAPO~\citep{DAPO} provide evidence that GRPO remains effective when applied to 32B base models.

\paragraph{Model types.}
Studies have shown the effectiveness of RL in different model types, including both base and long-CoT models (i.e., R1-distilled models and their fine-tuned variants).
Light-R1 \citep{wen2025light} first enhances R1-distilled models through SFT, then demonstrates further performance improvements via RL optimization.
Skywork-OR1 \citep{skywork-or1-2025} achieves significant gains by applying RL directly to R1-distilled models.

\subsubsection{Context Length}
\paragraph{Maximum Response Length.}
Improvements in reasoning capabilities are often associated with longer responses, as reflection and rethinking may occur during the reasoning process.
As such, maximum response length is another important factor in RL training. If the allowed response length is too short, longer rollouts may be cut off, resulting in zero reward even when the reasoning trajectories are valid. 
To address this, Light-R1~\citep{wen2025light} sets the maximum response length to 24k, with training response length converging to approximately 9k. 
DAPO~\citep{DAPO} uses a 16k maximum response length, where training response length converges to around 5k.

\paragraph{Curriculum Learning Based on Maximum Response Length.}
DeepScaleR \citep{deepscaler2025} adopts a progressive approach, gradually increasing the maximum response length from 8k to 16k and then to 24k, with performance improving consistently at each step. 
Similarly, Skywork-OR1 \citep{skywork-or1-2025} employs such multi-stage training with progressively extended maximum response lengths, reaching up to 32k. 

\paragraph{Truncated Rollouts.}
Skywork-OR1 \citep{skywork-or1-2025} conducts an ablation study to examine whether truncated rollouts should be masked (i.e., excluded from advantage calculation to avoid penalizing these rollouts). However, experimental results show that applying this masking strategy does not yield improved scaling behavior in later training stages, typically when the context length reaches 32k. Consequently, Skywork-OR1 opts not to apply masking for truncated rollouts during training.

\subsubsection{Reward Modeling}
\paragraph{Accuracy Reward.}
One of the key factors contributing to the success of RL in LLMs is the use of straightforward accuracy rewards. The minimal reward function design reduces the risk of reward hacking by leaving little room for unintended optimization.
However, rule-based reward system could fail in corner cases. To address this limitation, Seed-Thinking-v1.5 \citep{seed-thinking-1.5} proposes a more generalizable approach that leverages LLMs to evaluate a wide range of scenarios. Their framework includes a Seed-Verifier for straightforward answer verification and a Seed-Thinking-Verifier designed for cases requiring in-depth analytical reasoning. 
MiMo~\citep{xiaomi2025mimo} proposes Test Difficulty Driven Reward as a sample-difficulty-aware mechanism for better accuracy rewarding. 

\paragraph{Other Rewards.}
Apart from accuracy reward, several works have explored the incorporation of additional rewards or penalties.
For example, Open-Reasoner-Zero \citep{OpenReasonerZero2025}, Logic-RL \citep{xie2025logicrlunleashingllmreasoning}, and DeepScaleR \citep{deepscaler2025} integrate formatting considerations into their reward modeling. Notably, Open-Reasoner-Zero reports that the format reward rapidly saturates, typically reaching its maximum within approximately 60 steps.
However, there are no rigorous ablation studies to prove the effectiveness of the format reward.
DAPO \citep{DAPO} assigns a punitive reward to truncated samples to reduce the reward noise of the training process, which successfully stabilizes the generation entropy and results in better performance.

\subsubsection{KL Loss}
The KL loss is commonly utilized to constrain the divergence between the online policy and the frozen reference policy. 
However, ablation studies in Open-Reasoner-Zero \citep{OpenReasonerZero2025} suggest that KL regularization may not be essential for large-scale RL. In fact, it can significantly restrict the increase in response length.
Similarly, DAPO \citep{DAPO}, Dr. GRPO \citep{DrGRPO}, SRPO \citep{zhang2025srpocrossdomainimplementationlargescale} and MiMo \citep{xiaomi2025mimo} omit the KL loss during training and still achieve strong performance on various model sizes.
On the other hand, Light-R1 \citep{wen2025light} and Logic-RL \citep{xie2025logicrlunleashingllmreasoning} retain the KL loss and also report substantial improvements.
Skywork-OR1 \citep{skywork-or1-2025} conducts an ablation study on DeepSeek-R1-Distill-Qwen-7B and observes that incorporating KL loss causes the actor to stay too closely aligned with the reference model. The KL divergence quickly drops toward zero, limiting policy exploration. As a result, performance on AIME24 plateaus, with limited improvement over training. Based on these findings, Skywork-OR1 chooses not to apply KL loss during training.

\subsection{RLVR on Other Tasks}
\label{sec:rl_env}

The complex reasoning ability of DeepSeek-R1 has been significantly enhanced through RLVR, enabling its success in various reasoning-intensive tasks, such as complex context understanding and problem solving.
Fundamentally, RLVR allows an LLM agent to learn and perform any task with feasible answer verification, stimulating its complex reasoning ability without requiring human-guided process supervision.
Building on this inspiration, several works have explored the effectiveness of the complex reasoning paradigm with RLVR on various tasks.

\paragraph{Logical Reasoning.}
TinyZero \citep{tinyzero} and Mini-R1 \citep{mini-r1} attempts to reproduce the ``aha moment'' of Deepseek R1 on the countdown game with simple rule-based outcome reward. 
The Countdown game is a numerical puzzle in which players use a set of randomly drawn numbers and basic arithmetic operations (+, -, $\times$, $\div$) to reach or approximate a given target number as closely as possible.
In their rule-based reward system, there contains (1) format reward that ensures the model's response follows think-answer format, (2) validity reward that guarantees that the answer uses each number exactly once, and (3) accuracy reward that requires the final answer to correctly compute the target number.
Similarly, \citet{dalal2025sudoku, dalal2025teaching} examine the reward design on solving Sudoku puzzles. 
Apart from the necessary format compliance reward, these works carefully crafted the validity and accuracy rewards of the puzzle. 
For validity, solutions must be presented in a readable grid format, adhering to criteria such as the number of rows and columns, and the proper use of box-drawing characters. 
Additionally, solutions are required to fully preserve the original clues given by the inputs, and comply with the game's rules which prohibit repeated digits in any row, column, or 3$\times$3 box.  
For accuracy, solutions are evaluated based on the ratio of empty cells correctly filled by the model, with completely correct solutions receiving an extra large reward.
Logic-RL\citep{xie2025logicrlunleashingllmreasoning} and ZebraLogic \citep{lin2025zebralogicscalinglimitsllms} explore the capability of reasoning language models on deductive reasoning puzzles, where the methods can only be supervised by the correctness of outcome, and unsolvable samples may exist. 
These works have found that the design of format rewards should be elaborated to avoid hacking and encourage reflection, and extra reward on completely correct answer is necessary, which should also apply to those tasks with simple answer format, such as math/code problem solving and QA tasks. 

\paragraph{Application-oriented Tasks.} 
Reasoning language models are expected to learn to tackle real-world application-oriented tasks through thinking, planning and reflecting. 
To this end, SWE-RL \citep{wei2025swe} introduces a RL-based approach for GitHub issue fixing. 
Given the incorrect code context and the corresponding issue information, the LLM is required to reasoning about the issue and fix the issue by generating a corrected program. 
SWE-RL designs a rule-based reward function that computes the sequence similarity between the generated solution and the ground truth repaired program (extracted from the oracle patch merged by the pull request) as the reward. Additionally, solutions with wrong format should receive a large negative penalty.
Similar to SWE-RL, MT-R1-Zero \citep{feng2025mtr1zeroadvancingllmbasedmachine} proposes a a rule-metric mixed reward mechanism for training on machine translation. 
RAG-RL \citep{huang2025rag} equips LLM with retrieval-augmented generation (RAG) capabilities for multi-hop QA using a rule-based reward system comprising of three components: answer rewards, citation rewards, and formatting rewards.
Answer rewards evaluate the exact match between model predictions and ground truth results to incentivize correct final answers. 
Citation rewards count the recall of relevant citations cited in the final answer to encourage effective citations. 
Formatting rewards utilize a binary function to enforce the desired output format, ensuring to present proper XML tags and required headings while preventing excessive text and raw Unicode. 
RLSF \citep{jha2024rlsf} attempts to address chemistry tasks with rule-based evaluation generated by \citet{RDKit} as the rewards, which is a token-level vector based on the presence or absence of any syntactical errors. 
These chemical tasks includes (1) Forward synthesis, which involves predicting the product of a chemical reaction based on given reactants and reagents; (2) Retrosynthesis, which involves determining the reactants required to create a specific product; and (3) Molecular generation, which involves generating a molecule that meets specific requested chemical and biological properties in natural language.

\paragraph{Exploration Beyond Supervision.}

Through the reinforcement learning process, exciting observations reveal that LLMs have demonstrated remarkably promising and unexpected capabilities.
Several works have discovered the emergence of new abilities in LLMs through RL on complex reasoning tasks, without the guidance of any supervision.
RL-Poet \citep{rl-poet} tunes Pleias-350M with 200K verses, transforming the small language model into a poet using only format rewards rather than outcome-based rewards.
The rule-based reward system of RL-Poet consists of three components: (1) Non-repetition reward that penalizes repetition; (2) Verse reward that enforces a structured verse format, requiring the mean line length to be within 30\% of the length of prompt, and at least 80\% of lines to start with uppercase; and (3) Quatrain reward that ensures the output to be formatted with four-line verse blocks, adhering to the standard quatrain structure.
The trained model could generate literary poems across diverse topics and moods, demonstrating its creative writing capabilities.
More excitingly, \citet{dalal2025sorting} explore the potential of RL for knowledge discovery, by extending its utility to discover a more efficient sorting algorithm. 
During the RL process, the LLM is required to improve the efficiency of a baseline sorting algorithm on a series of competitive test cases.
These test cases are meticulously designed to emphasize pattern diversity, size scaling, data type variation, and difficulty distribution.
The model is optimized using rule-based rewards. In addition to the necessary format and validity rewards, the critical performance reward evaluates the logarithmic execution time on given test cases. 
Experiments show that although the baseline sorting algorithm (i.e., Timsort) is already good enough, the model have discovered several outstanding algorithms, in which the Hybrid Partitioning with QuickSelect achieved a 47.92x speedup over the baseline Timsort implementation on a dataset of 42,385 elements with a Gaussian distribution.
These results highlight the potential of complex reasoning language models to surpass the limitations of supervised data resources, and even humans, by adopting RL training strategies.

\section{More Directions}
\label{sec:others}

    While recent efforts have made significant progress in replicating and extending the capabilities of DeepSeek-R1, several open questions and challenges remain in the development of robust reasoning language models. In this section, we highlight emerging directions that appear in recent literature or hold potential to shape the next generation of reasoning models.
    Each subsection explores a complementary aspect, from alternative training methods and alignment strategies to broader concerns around generalizability, robustness, safety, and inclusivity.  

    \subsection{Alternative Approaches for Reasoning Enhancement}
    \label{sec:others_methods}

    While reinforcement learning from verifiable rewards (RLVR) has driven notable progress in reasoning language models, its current form remains limited, particularly in capturing intermediate reasoning steps and aligning with human expectations. To address these gaps, recent research has explored alternative approaches that complement or extend traditional RLVR techniques. In this section, we discuss two emerging directions: (i) more expressive and step-aware reward modeling methods, and (ii) preference optimization strategies that reduce computational overhead while improving training stability. 

    \label{sec:other_reward}

        \noindent\paragraph{Reward Modeling Techniques.} 
        The effectiveness of reinforcement learning in training reasoning language models is largely dependent on the quality and alignment of the reward model, particularly its ability to reflect human preferences or factual correctness rooted in scientific principles. Traditionally, reward model quality is assessed by its accuracy in assigning feedback. However, as argued by \citet{razin2025makes}, accuracy alone is insufficient. An accurate reward model does not necessarily make for an effective teacher. Empirical results from reinforcement learning with human feedback (RLHF) setups demonstrate that low reward variance can significantly hinder learning, regardless of a reward model's accuracy. Specifically, it leads to a flattened optimization landscape, resulting in slow convergence and worse performance than less accurate models with higher reward variance. Additionally, reward models can suffer from compatibility issues in behavior across different language models. A reward model that provides more informative gradients and high reward variance for one model may produce low reward variance for another, significantly impeding effective learning.

        To improve the optimization landscape characterized by low reward variance, enhancing the reward model can be achieved through several strategies. Process-level Reward Modeling (PRM) \citep{xiong2024watch, crplanner, song2025prmbench, ma2025s, lyu2025exploringlimitoutcomereward} transcends simplistic outcome-based score annotations by providing feedback at each intermediate step within a reasoning process, rather than solely assessing the final outcome. This granular supervision enables models to navigate complex, multi-step tasks more effectively, ensuring that each action aligns with the desired reasoning trajectory. By focusing on step-level evaluations, PRM introduces a more stochastic reward pattern, enhancing the model’s adaptability and robustness in dynamic environments.
        Notably, rStar-Math \citep{guan2025rstar} enhances PRM by incorporating a Process Preference Model (PPM). The PPM is trained to assess intermediate reasoning steps by providing step-level preference data, rather than relying solely on final outcomes. This approach allows the model to distinguish between more and less promising reasoning paths during Monte Carlo Tree Search. Additionally, they implement a self-evolution strategy within rStar-Math, where both the policy small language model and the PPM are iteratively refined from scratch. Through multiple rounds of evolution, these models progressively enhance their reasoning capabilities, leading to significant performance improvements.
        Besides, PRIME \citep{cui2025process} introduces a scalable reinforcement learning framework to enhance reasoning capabilities in language models. PRIME employs an implicit PRM that is trained solely on outcome labels, thereby eliminating the need for expensive, manually annotated step-level labels. This approach enables online updates using policy rollouts and outcome labels, ensuring scalability and adaptability during reinforcement learning training. By integrating implicit process rewards with traditional outcome rewards, PRIME effectively computes advantages during policy updates, enhancing training efficiency and addressing challenges related to credit assignment in complex reasoning tasks. Particularly, this method reduces development overhead and mitigates issues like reward hacking and overoptimization by avoiding explicit process-level annotations and facilitating online updates of the PRM.

    \noindent\paragraph{Preference Optimization.} \label{sec:other_preference_optimization}
    Although the training method of DeepSeek-R1 significantly enhances the model’s reasoning ability, it requires substantial computational resources for online RL training. In contrast to online approaches like PPO \citep{PPO} and GRPO \citep{GRPO}, Direct Preference Optimization (DPO) demands much less computational resources~\citep{Rafailov2023DirectPO}. By simply constructing chosen and rejected pairs, models can be trained directly, making DPO a more efficient alternative to PPO and GRPO. 
    
    Several works adopt DPO to improve the reasoning performance of language models.
    EXAONE Deep \citep{research2025exaone} presents a series of reasoning language models which utilize 1.6M data samples for supervised fine-tuning (SFT), 20K instances for preference optimization with the method from \citep{xiao2025simper}, and 10K for online RL training. As a result, EXAONE Deep 2.4B and 7.8B outperform models in comparable sizes, and EXAONE Deep 32B shows competitive performance against leading models. However, the training data is not publicly open-sourced, and there is no detail on how performance can be improved after preference optimization.
    Light-R1~\citep{wen2025light} employs a method of curriculum post-training with three training stages--two SFT stages and a DPO stage at the last. The first SFT stage uses 76k training samples, and the second one uses 3k highly difficult samples. To construct the preference pairs for the last DPO stage, it uses the rollouts from DeepSeek-R1 with verified correct answers as chosen samples, and the rollouts from the checkpoint after the second stage with verified incorrect answers as the corresponding reject samples. 
    Iterative DPO \citep{tu2025enhancing} finds that DPO can rapidly improve the model's reasoning capabilities by using various methods of constructing chosen and rejected pairs. Through multiple rounds of DPO training, they show that this method can rival the performance of online RL approaches, such as Simple-RL-Zero~\citep{zeng2025simplerlzooinvestigatingtamingzero}. 
    RedStar \citep{xu2025redstar} also studies DPO training to enhance the model’s reasoning capabilities. It constructs positive and negative samples using a rule-based reward: a reward of 1 is assigned if verification succeeds, and 0 otherwise. The study compares DPO with PPO and REINFORCE++~\citep{reinforce++} and shows that DPO training is more effective than PPO in improving the model’s reasoning abilities.
    DPO-R1 \citep{zhang2025dpor1} explores the feasibility of DPO and RAFT~\citep{dong2023raft}, indicating that DPO substantially enhances model performance while maintaining high training efficiency, and incorporating a SFT warm-up phase before DPO further boosts performance. Nonetheless, DPO still lags slightly behind PPO in overall effectiveness in their experiments. 
    
    \subsection{Generalizability}
    
    Achieving robust generalization is often considered as a critical challenge in the deep learning era. 
    However, current studies of reasoning language models in pre-training, supervised fine-tuning, and reinforcement learning have demonstrated that these models are well generalized to handle out-of-distribution tasks when learning to improve their reasoning ability. 

    \paragraph{Continual Pre-training.}
    
    Continual pre-training on mathematical reasoning tasks has been shown to substantially enhance both specialized and general reasoning abilities in language models. GRPO \citep{GRPO} studied the continual pre-training in mathematical reasoning and demonstrated that models like DeepSeekMath-Base 7B exhibit enhanced performance not only in mathematical problem-solving but also on benchmarks such as the Massive Multitask Language Understanding (MMLU, \citet{xuan2025mmlu}) and Big-Bench Hard (BBH, \citet{bbh-suzgun2022challengingbigbenchtaskschainofthought}). For instance, DeepSeekMath-Base 7B achieved a 54.9\% score on MMLU and 59.5\% on BBH, surpassing its precursor, DeepSeek-Coder-Base-v1.5, which scored 49.1\% and 55.2\% respectively. This suggests that mathematical training can positively influence a model’s general reasoning capabilities.

    \paragraph{Supervised Fine-tuning.}

    REFT \citep{trung2024reft} and Light-R1 \citep{wen2025light} validate that supervised fine-tuning (SFT) plays a critical role in enhancing the generalization of language models by providing structured, high-quality reasoning examples that serve as strong inductive priors. It bootstraps initial reasoning capabilities, enabling models to internalize latent abstractions and problem-solving strategies that transfer across tasks. By exposing the model to diverse solution paths, SFT establishes a stable base policy for further reinforcement learning that significantly improves the efficiency and effectiveness of subsequent reward-based optimization, reducing reward hacking and guiding exploration toward more reliable, outcome-driven reasoning behaviors.
    The power of SFT on generalization is further emphasized by LIMO~\citep{ye2025limoreasoning}, which demonstrates that carefully curated, high-quality training examples play a pivotal role in enabling broader generalization. The study highlights the importance of strategic data selection in fostering versatile reasoning capabilities, as well as robustness to out-of-distribution (OOD) variations, such as Chinese math problems that were not present in the training set. 

    \paragraph{Reinforcement Learning.}

    Current outcome-reward-based reinforcement learning (RL) for reasoning language models has demonstrated strong potential for out-of-domain generalization.
    Through an RL process, reasoning language models demonstrate strong generalization capability across tasks, languages, and modalities, being far beyond what is possible with imitative learning alone.
    Llama3-SWE-RL \citep{huang2025rag} demonstrates improved results on five out-of-domain tasks, including function coding, library use, code reasoning, mathematics, and general language understanding, despite being trained solely on the code repair task.
    In contrast, a supervised fine-tuning baseline led to an average performance degradation.
    RL-Poet \citep{rl-poet} demonstrates the ability to generate literary poems in multiple languages with correct poetic rules, despite being trained almost exclusively on English data.
    Compared to the prevailing imitative learning paradigm, these results highlight the potential of achieving artificial general intelligence through general reinforcement learning.
    In other respects, \citet{tang2025rl} introduces Any-Generation Reward Optimization (AGRO) that enhances the generalizability of reasoning language models by integrating learning from both on-policy and off-policy experiences. Specifically, AGRO leverages both current (on-policy) data, collected from the model’s existing policy, and historical (off-policy) data, experiences gathered from previous policies, to enable models to learn from a broader spectrum of scenarios. This comprehensive exposure mitigates the risk of overfitting and enhances the model’s adaptability to novel situations. Consequently, AGRO-trained models are better equipped to generalize across diverse tasks and environments, a critical attribute for deploying reasoning language models in real-world applications where variability is inherent.
    
    In comparing the roles the supervised fine-tuning (SFT) and outcome-reward-based reinforcement learning (RL) play in the context of generalization, \citet{chu2025sft} demonstrates that RL significantly enhances a model’s ability to generalize across both textual and visual domains. In contrast, SFT often encourages memorization of the training data, which can impair performance on out-of-distribution tasks. Interestingly, while RL drives generalization, SFT remains crucial for stabilizing the model’s output format—an essential property that facilitates effective downstream RL optimization, highlighting the complementary nature of SFT and RL in shaping models that can acquire and transfer knowledge across diverse, multimodal tasks.
    However, recent studies have raised concerns regarding the limitations of RL when applied to reasoning language models. \citet{yue2025doesreinforcementlearningreally} points out that RL training in reasoning language models may narrow the scope of reasoning capabilities while enhancing sampling efficiency, and that RL-trained models generally underperform compared to base models in pass@k metrics at larger k values. Similarly, \citet{hochlehnert2025sober} observes that the generalization ability of RL methods on smaller language models is significantly limited, possibly due to the restricted prior knowledge available for RL training to exploit. 
    
    In summary, these findings underscore both the promise and challenges of applying RL to reasoning and generalization in reasoning language models. While general RL approaches demonstrate encouraging out-of-domain performance gains and broader adaptability, careful attention must be given to potential trade-offs.

    \subsection{Safety}

        Ensuring the safety and robustness of large language models (LLMs) against vulnerabilities and attacks is a critical research area that has been widely explored in previous literature~\citep{kaddour2023challenges,zhao2023survey,das2025security}. However, reasoning language models introduce new safety challenges arising from their training algorithms, adversarial attacks during inference, and vulnerabilities related to their deployment environments~\citep{zhou2025hidden,jiang2025safechain}. In this section, we review recent advancements addressing these emerging concerns and highlight promising approaches for enhancing detection and defense mechanisms.

        \paragraph{Self-Evolution and Reward Hacking.} Reasoning language models have demonstrated significant potential, paving the way toward superintelligent models capable of continuous self-improvement~\citep{leike2023superalignment,li2024large,tao2024survey}. However, their self-evolution process introduces safety concerns and risks producing uncontrollable outcomes misaligned with human values and preferences~\citep{taubenfeld2024systematic}. 
        A promising direction for improving these models involves using reinforcement learning algorithms with reward signals~\citep{DAPO,Deepseek_R1,openai-o1}. However, this approach inevitably introduces the issue of reward hacking, a longstanding challenge within the reinforcement learning research community~\citep{amodei2016concrete,everitt2017reinforcement,everitt2021reward,weng2024reward}. Reward hacking occurs when a model exploits flaws or ambiguities in the reward function, primarily because the reinforcement learning environment is imperfect and struggles to provide complete and accurate reward signals. 

        \paragraph{Jailbreaking on Reasoning Language Models.} Jailbreak attacks and defenses play crucial roles in maintaining the robustness and security of large language models (LLMs)~\citep{zhou2024easyjailbreak,yi2024jailbreak}. The same philosophy applies to reasoning language models~\citep{kuo2025h}. Recent work by \cite{sabbaghi2025adversarial} introduces an adversarial reasoning method for constructing effective jailbreak trajectories, achieving an attack success rate of 56\% on OpenAI-o1~\citep{openai-o1} and 100\% on Deepseek-R1~\citep{Deepseek_R1}. \cite{yao2025mousetrap} discusses inherent flaws in reasoning language models and proposes a novel jailbreaking attack targeting reasoning language models. \cite{arrieta2025o3} also states that Deepseek-R1 produces more unsafe responses than OpenAI models.        
        These results highlight the importance of employing safety-focused supervised fine-tuning and reinforcement learning to safeguard reasoning language models against adversarial attacks. However, previous studies indicate that incorporating safety alignment can inadvertently compromise the reasoning capabilities of these models~\citep{huang2025safety}. Moreover, \cite{zhao2025trade} and \citep{jiang2025safechain} observes substantial decreases in both helpfulness and harmlessness in reasoning language models compared to baseline models.

        \paragraph{Overthinking.} Reasoning language models allow for extended reasoning chains during inference, but this capability can sometimes cause issues like overthinking~\citep{sui2025stopoverthinkingsurveyefficient,chen2024not}. Commercial model services typically charge more for output tokens, including reasoning tokens, than input tokens. Thus, attacks like OverThink~\citep{kumar2025overthink} can trigger excessive reasoning, raising operational and environmental costs. Additionally, overthinking suggests that the model is heavily reliant on internal simulations of potential actions and outcomes. Consequently, studies from \cite{cuadron2025danger} and \cite{feng2025reasoning} have emphasized that reasoning language models may exhibit reduced performance in agentic scenarios when environmental feedback is neglected.

        Effective safety measures for reasoning language models typically combine prevention, detection, and mitigation strategies. First, the methodologies to mitigate reward hacking include better algorithm design~\citep{amodei2016concrete,uesato2020avoiding,pan2022effects} and training strategies~\cite{denison2024sycophancy,li2025output}. \cite{guan2024deliberative} introduces reasoning alignment over safety policies to enhance a model's robustness against jailbreak attacks. \cite{jiang2025safechain} introduces decoding strategies aimed at improving the safety of reasoning language models, with some performance trade-offs, and provides post-training datasets for better alignment. Additionally, several studies have equipped safeguard models with reasoning capabilities to more effectively detect potential threats~\citep{liu2025guardreasoner, wen2025thinkguard}.

    \subsection{Multimodal and Multilingual}

        Multimodal reasoning language models are primarily developed via two predominant approaches: post-alignment~\citep{zhang2023video,chu2024qwen2,chen2024internvl,grattafiori2024llama} and mixed-modality pretraining~\citep{team2023gemini,team2024gemini,nguyen2025spirit}. However, both approaches generally yield weaker reasoning capabilities compared to single-modality models~\citep{wu2023multimodal,liang2024survey,wang2024exploring}. Recent studies have sought to improve test-time scaling for multimodal reasoning language models across various modalities, including visual~\citep{liu2024diving,wang2025multimodal,du2025virgo,sun2025mm}, audio~\citep{du2024cot,ma2025audio,xie2025audio,li2025reinforcement}, and others, such as 3D data, tabular information, and sensor inputs~\citep{wang2024chain,dai2025multimodal}. Furthermore, research by \cite{du2025virgo} demonstrates that reasoning capabilities developed in single-modality reasoning language models can effectively transfer to multimodal contexts. However, applying advanced RL and PRM to multimodal large reasoning language models remains a challenging yet promising research direction~\citep{wu2025boosting}.

        The challenges associated with multilingual reasoning language models differ primarily due to the limited availability of resources in certain languages, resulting in weaker performance from the base model~\citep{nguyen2023culturax,qin2024multilingual}. Research in this area remains limited, with two central issues emerging: (1) evaluating the extent to which reasoning abilities trained predominantly in English can generalize effectively to other languages, and (2) determining whether multilingual contexts necessitate specialized model capabilities to effectively facilitate insight or trigger "aha" moments. \cite{xuan2025mmlu} observes that reasoning-enhanced models do not uniformly improve multilingual capabilities, emphasizing the importance of targeted multilingual reasoning enhancements in reasoning language models. Researchers have also proposed multilingual SFT and RL algorithms to enhance consistency across different languages~\citep{lai2024mcot,chai2024xcot,wang2025demystifying}. We anticipate that future research will focus on more efficient training strategies for multilingual reasoning language models, with particular emphasis on improving performance in low-resource languages.

\section{Conclusions}
In this survey, we present a comprehensive overview of the replication efforts inspired by DeepSeek-R1, with a particular emphasis on the methodologies and insights underpinning supervised fine-tuning and reinforcement learning approaches. We explore how open-source projects have curated instruction-tuning datasets, implemented outcome-reward-based reinforcement learning strategies, and designed reward systems aimed at enhancing models' reasoning capabilities. Beyond synthesizing trends from current initiatives, we also offer our perspective on promising future directions for the field. These include the expansion of reasoning skills beyond mathematical and coding tasks, the advancement of model safety and interpretability, and the refinement of reward mechanisms to foster more sophisticated reasoning behaviors. We hope this survey not only captures the recent progress but also provides a solid foundation for ongoing research and marks a step forward toward achieving artificial general intelligence.

\bibliography{iclr2025_conference}

\begin{thebibliography}{164}
\providecommand{\natexlab}[1]{#1}
\providecommand{\url}[1]{\texttt{#1}}
\expandafter\ifx\csname urlstyle\endcsname\relax
  \providecommand{\doi}[1]{doi: #1}\else
  \providecommand{\doi}{doi: \begingroup \urlstyle{rm}\Url}\fi

\bibitem[Aggarwal \& Welleck(2025)Aggarwal and Welleck]{aggarwal2025l1}
Pranjal Aggarwal and Sean Welleck.
\newblock L1: Controlling how long a reasoning model thinks with reinforcement learning.
\newblock \emph{arXiv preprint arXiv:2503.04697}, 2025.

\bibitem[Ahmadian et~al.(2024)Ahmadian, Cremer, Gall{\'e}, Fadaee, Kreutzer, Pietquin, {\"U}st{\"u}n, and Hooker]{reinforce_LLM}
Arash Ahmadian, Chris Cremer, Matthias Gall{\'e}, Marzieh Fadaee, Julia Kreutzer, Olivier Pietquin, Ahmet {\"U}st{\"u}n, and Sara Hooker.
\newblock Back to basics: Revisiting {REINFORCE}-style optimization for learning from human feedback in {LLM}s.
\newblock In Lun-Wei Ku, Andre Martins, and Vivek Srikumar (eds.), \emph{Proceedings of the 62nd Annual Meeting of the Association for Computational Linguistics (Volume 1: Long Papers)}, pp.\  12248--12267, Bangkok, Thailand, August 2024. Association for Computational Linguistics.
\newblock \doi{10.18653/v1/2024.acl-long.662}.
\newblock URL \url{https://aclanthology.org/2024.acl-long.662/}.

\bibitem[Albalak et~al.(2025)Albalak, Phung, Lile, Rafailov, Gandhi, Castricato, Singh, Blagden, Xiang, Mahan, and Haber]{albalak2025bigmathlargescalehighqualitymath}
Alon Albalak, Duy Phung, Nathan Lile, Rafael Rafailov, Kanishk Gandhi, Louis Castricato, Anikait Singh, Chase Blagden, Violet Xiang, Dakota Mahan, and Nick Haber.
\newblock Big-math: A large-scale, high-quality math dataset for reinforcement learning in language models, 2025.
\newblock URL \url{https://arxiv.org/abs/2502.17387}.

\bibitem[Amodei et~al.(2016)Amodei, Olah, Steinhardt, Christiano, Schulman, and Man{\'e}]{amodei2016concrete}
Dario Amodei, Chris Olah, Jacob Steinhardt, Paul Christiano, John Schulman, and Dan Man{\'e}.
\newblock Concrete problems in ai safety.
\newblock \emph{arXiv preprint arXiv:1606.06565}, 2016.

\bibitem[AoPS(2025)]{aops}
AoPS.
\newblock Aops wiki:competition ratings.
\newblock \url{https://artofproblemsolving.com/wiki/index.php/AoPS_Wiki:Competition_ratings}, 2025.
\newblock Accessed: May 1, 2025.

\bibitem[Arora \& Zanette(2025)Arora and Zanette]{arora2025traininglanguagemodelsreason}
Daman Arora and Andrea Zanette.
\newblock Training language models to reason efficiently, 2025.
\newblock URL \url{https://arxiv.org/abs/2502.04463}.

\bibitem[Arrieta et~al.(2025)Arrieta, Ugarte, Valle, Parejo, and Segura]{arrieta2025o3}
Aitor Arrieta, Miriam Ugarte, Pablo Valle, Jos{\'e}~Antonio Parejo, and Sergio Segura.
\newblock o3-mini vs deepseek-r1: Which one is safer?
\newblock \emph{arXiv preprint arXiv:2501.18438}, 2025.

\bibitem[Bespoke-Labs(2025)]{bespoke_stratos}
Bespoke-Labs.
\newblock Bespoke-stratos: The unreasonable effectiveness of reasoning distillation.
\newblock https://www.bespokelabs.ai/blog/bespoke-stratos-the-unreasonable-effectiveness-of-reasoning-distillation, 2025.
\newblock Accessed: 2025-01-22.

\bibitem[ByteDance-Seed(2025)]{seed-thinking-1.5}
ByteDance-Seed.
\newblock Seed-thinking-v1.5: Advancing superb reasoning models with reinforcement learning.
\newblock \url{https://github.com/ByteDance-Seed/Seed-Thinking-v1.5}, 2025.
\newblock Accessed: 2025-04-10.

\bibitem[Chai et~al.(2024)Chai, Yang, Sun, Guo, Liu, Wang, Liang, Bai, Li, Peng, et~al.]{chai2024xcot}
Linzheng Chai, Jian Yang, Tao Sun, Hongcheng Guo, Jiaheng Liu, Bing Wang, Xiannian Liang, Jiaqi Bai, Tongliang Li, Qiyao Peng, et~al.
\newblock xcot: Cross-lingual instruction tuning for cross-lingual chain-of-thought reasoning.
\newblock \emph{arXiv preprint arXiv:2401.07037}, 2024.

\bibitem[Chen et~al.(2025)Chen, Xu, Zhang, Chan, Liu, Bing, Zhao, Luu, and Rong]{chen2025finereason}
Guizhen Chen, Weiwen Xu, Hao Zhang, Hou~Pong Chan, Chaoqun Liu, Lidong Bing, Deli Zhao, Anh~Tuan Luu, and Yu~Rong.
\newblock Finereason: Evaluating and improving llms' deliberate reasoning through reflective puzzle solving.
\newblock \emph{arXiv preprint arXiv:2502.20238}, 2025.

\bibitem[Chen et~al.(2023)Chen, Wong, Chen, and Tian]{chen2023extending}
Shouyuan Chen, Sherman Wong, Liangjian Chen, and Yuandong Tian.
\newblock Extending context window of large language models via positional interpolation.
\newblock \emph{arXiv preprint arXiv:2306.15595}, 2023.
\newblock URL \url{https://arxiv.org/abs/2306.15595}.

\bibitem[Chen et~al.(2024{\natexlab{a}})Chen, Xu, Liang, He, Pang, Yu, Song, Liu, Zhou, Zhang, et~al.]{chen2024not}
Xingyu Chen, Jiahao Xu, Tian Liang, Zhiwei He, Jianhui Pang, Dian Yu, Linfeng Song, Qiuzhi Liu, Mengfei Zhou, Zhuosheng Zhang, et~al.
\newblock Do not think that much for 2+ 3=? on the overthinking of o1-like llms.
\newblock \emph{arXiv preprint arXiv:2412.21187}, 2024{\natexlab{a}}.

\bibitem[Chen et~al.(2024{\natexlab{b}})Chen, Wu, Wang, Su, Chen, Xing, Zhong, Zhang, Zhu, Lu, et~al.]{chen2024internvl}
Zhe Chen, Jiannan Wu, Wenhai Wang, Weijie Su, Guo Chen, Sen Xing, Muyan Zhong, Qinglong Zhang, Xizhou Zhu, Lewei Lu, et~al.
\newblock Internvl: Scaling up vision foundation models and aligning for generic visual-linguistic tasks.
\newblock In \emph{Proceedings of the IEEE/CVF conference on computer vision and pattern recognition}, pp.\  24185--24198, 2024{\natexlab{b}}.

\bibitem[Chu et~al.(2025{\natexlab{a}})Chu, Zhai, Yang, Tong, Xie, Schuurmans, Le, Levine, and Ma]{chu2025sft}
Tianzhe Chu, Yuexiang Zhai, Jihan Yang, Shengbang Tong, Saining Xie, Dale Schuurmans, Quoc~V Le, Sergey Levine, and Yi~Ma.
\newblock Sft memorizes, rl generalizes: A comparative study of foundation model post-training.
\newblock \emph{arXiv preprint arXiv:2501.17161}, 2025{\natexlab{a}}.

\bibitem[Chu et~al.(2025{\natexlab{b}})Chu, Huang, Zhang, Wei, and Wang]{chu2025gpg}
Xiangxiang Chu, Hailang Huang, Xiao Zhang, Fei Wei, and Yong Wang.
\newblock Gpg: A simple and strong reinforcement learning baseline for model reasoning.
\newblock \emph{arXiv preprint arXiv:2504.02546}, 2025{\natexlab{b}}.

\bibitem[Chu et~al.(2024)Chu, Xu, Yang, Wei, Wei, Guo, Leng, Lv, He, Lin, et~al.]{chu2024qwen2}
Yunfei Chu, Jin Xu, Qian Yang, Haojie Wei, Xipin Wei, Zhifang Guo, Yichong Leng, Yuanjun Lv, Jinzheng He, Junyang Lin, et~al.
\newblock Qwen2-audio technical report.
\newblock \emph{arXiv preprint arXiv:2407.10759}, 2024.

\bibitem[Cuadron et~al.(2025)Cuadron, Li, Ma, Wang, Wang, Zhuang, Liu, Schroeder, Xia, Mao, et~al.]{cuadron2025danger}
Alejandro Cuadron, Dacheng Li, Wenjie Ma, Xingyao Wang, Yichuan Wang, Siyuan Zhuang, Shu Liu, Luis~Gaspar Schroeder, Tian Xia, Huanzhi Mao, et~al.
\newblock The danger of overthinking: Examining the reasoning-action dilemma in agentic tasks.
\newblock \emph{arXiv preprint arXiv:2502.08235}, 2025.

\bibitem[Cui et~al.(2025)Cui, Yuan, Wang, Wang, Li, He, Fan, Yu, Xu, Chen, et~al.]{cui2025process}
Ganqu Cui, Lifan Yuan, Zefan Wang, Hanbin Wang, Wendi Li, Bingxiang He, Yuchen Fan, Tianyu Yu, Qixin Xu, Weize Chen, et~al.
\newblock Process reinforcement through implicit rewards.
\newblock \emph{arXiv preprint arXiv:2502.01456}, 2025.

\bibitem[Dai et~al.(2025)Dai, Han, and Liu]{dai2025multimodal}
Yue Dai, Soyeon~Caren Han, and Wei Liu.
\newblock Multimodal graph constrastive learning and prompt for chartqa.
\newblock \emph{arXiv preprint arXiv:2501.04303}, 2025.

\bibitem[Dalal(2025{\natexlab{a}})]{dalal2025sorting}
Hrishbh Dalal.
\newblock Ai as algorithm designer: Teaching llms to improve sorting through trial and error in grpo.
\newblock \emph{Personal Website}, March 2025{\natexlab{a}}.
\newblock URL \url{https://hrishbh.com/ai-as-algorithm-designer-teaching-llms-to-improve-sorting-through-trial-and-error-in-grpo/}.

\bibitem[Dalal(2025{\natexlab{b}})]{dalal2025sudoku}
Hrishbh Dalal.
\newblock Teaching language models to invent or optimize efficient sudoku algorithms through reinforcement learning, 3 2025{\natexlab{b}}.
\newblock URL \url{https://hrishbh.com/teaching-language-models-to-invent-or-optimize-efficient-sudoku-algorithms-through-reinforcement-learning/}.

\bibitem[Dalal(2025{\natexlab{c}})]{dalal2025teaching}
Hrishbh Dalal.
\newblock Teaching language models to solve sudoku through reinforcement learning, 3 2025{\natexlab{c}}.
\newblock URL \url{https://hrishbhdalal.com/projects/teaching-language-models-sudoku}.
\newblock Accessed on March 19, 2025.

\bibitem[Das et~al.(2025)Das, Amini, and Wu]{das2025security}
Badhan~Chandra Das, M~Hadi Amini, and Yanzhao Wu.
\newblock Security and privacy challenges of large language models: A survey.
\newblock \emph{ACM Computing Surveys}, 57\penalty0 (6):\penalty0 1--39, 2025.

\bibitem[DeepSeek-AI(2024)]{deepseekai2024deepseekv3}
DeepSeek-AI.
\newblock Deepseek-v3 technical report, 2024.
\newblock URL \url{https://arxiv.org/abs/2412.19437}.

\bibitem[Denison et~al.(2024)Denison, MacDiarmid, Barez, Duvenaud, Kravec, Marks, Schiefer, Soklaski, Tamkin, Kaplan, et~al.]{denison2024sycophancy}
Carson Denison, Monte MacDiarmid, Fazl Barez, David Duvenaud, Shauna Kravec, Samuel Marks, Nicholas Schiefer, Ryan Soklaski, Alex Tamkin, Jared Kaplan, et~al.
\newblock Sycophancy to subterfuge: Investigating reward-tampering in large language models.
\newblock \emph{arXiv preprint arXiv:2406.10162}, 2024.

\bibitem[Dong et~al.(2023)Dong, Xiong, Goyal, Zhang, Chow, Pan, Diao, Zhang, Shum, and Zhang]{dong2023raft}
Hanze Dong, Wei Xiong, Deepanshu Goyal, Yihan Zhang, Winnie Chow, Rui Pan, Shizhe Diao, Jipeng Zhang, Kashun Shum, and Tong Zhang.
\newblock Raft: Reward ranked finetuning for generative foundation model alignment.
\newblock \emph{arXiv preprint arXiv:2304.06767}, 2023.

\bibitem[Doria(2025)]{rl-poet}
Alexander Doria.
\newblock Rl, reasoning \& writing - grpo on base model.
\newblock \url{https://colab.research.google.com/drive/1Ty0ovsrpw8i-zJvDhlSAtBIVw3EZfHK5}, 2025.

\bibitem[Du et~al.(2024)Du, Ma, Yang, Deng, Chen, Yang, Xiang, Liu, and Qin]{du2024cot}
Yexing Du, Ziyang Ma, Yifan Yang, Keqi Deng, Xie Chen, Bo~Yang, Yang Xiang, Ming Liu, and Bing Qin.
\newblock Cot-st: Enhancing llm-based speech translation with multimodal chain-of-thought.
\newblock \emph{arXiv preprint arXiv:2409.19510}, 2024.

\bibitem[Du et~al.(2025)Du, Liu, Li, Zhao, Huo, Wang, Chen, Liu, Wang, and Wen]{du2025virgo}
Yifan Du, Zikang Liu, Yifan Li, Wayne~Xin Zhao, Yuqi Huo, Bingning Wang, Weipeng Chen, Zheng Liu, Zhongyuan Wang, and Ji-Rong Wen.
\newblock Virgo: A preliminary exploration on reproducing o1-like mllm.
\newblock \emph{arXiv preprint arXiv:2501.01904}, 2025.

\bibitem[Dubey et~al.(2024)Dubey, Jauhri, Pandey, Kadian, Al-Dahle, Letman, Mathur, Schelten, Yang, Fan, et~al.]{dubey2024llama3herdmodels}
Abhimanyu Dubey, Abhinav Jauhri, Abhinav Pandey, Abhishek Kadian, Ahmad Al-Dahle, Aiesha Letman, Akhil Mathur, Alan Schelten, Amy Yang, Angela Fan, et~al.
\newblock The llama 3 herd of models.
\newblock \emph{arXiv preprint arXiv:2407.21783}, 2024.

\bibitem[Everitt et~al.(2017)Everitt, Krakovna, Orseau, Hutter, and Legg]{everitt2017reinforcement}
Tom Everitt, Victoria Krakovna, Laurent Orseau, Marcus Hutter, and Shane Legg.
\newblock Reinforcement learning with a corrupted reward channel.
\newblock \emph{arXiv preprint arXiv:1705.08417}, 2017.

\bibitem[Everitt et~al.(2021)Everitt, Hutter, Kumar, and Krakovna]{everitt2021reward}
Tom Everitt, Marcus Hutter, Ramana Kumar, and Victoria Krakovna.
\newblock Reward tampering problems and solutions in reinforcement learning: A causal influence diagram perspective.
\newblock \emph{Synthese}, 198\penalty0 (Suppl 27):\penalty0 6435--6467, 2021.

\bibitem[Feng et~al.(2025{\natexlab{a}})Feng, Dou, and Kong]{feng2025reasoning}
Xiachong Feng, Longxu Dou, and Lingpeng Kong.
\newblock Reasoning does not necessarily improve role-playing ability.
\newblock \emph{arXiv preprint arXiv:2502.16940}, 2025{\natexlab{a}}.

\bibitem[Feng et~al.(2025{\natexlab{b}})Feng, Cao, Ren, Su, Chen, Zhang, Xu, Hu, Wu, and Liu]{feng2025mtr1zeroadvancingllmbasedmachine}
Zhaopeng Feng, Shaosheng Cao, Jiahan Ren, Jiayuan Su, Ruizhe Chen, Yan Zhang, Zhe Xu, Yao Hu, Jian Wu, and Zuozhu Liu.
\newblock Mt-r1-zero: Advancing llm-based machine translation via r1-zero-like reinforcement learning, 2025{\natexlab{b}}.
\newblock URL \url{https://arxiv.org/abs/2504.10160}.

\bibitem[Gandhi et~al.(2025)Gandhi, Chakravarthy, Singh, Lile, and Goodman]{gandhi2025cognitive}
Kanishk Gandhi, Ayush Chakravarthy, Anikait Singh, Nathan Lile, and Noah~D Goodman.
\newblock Cognitive behaviors that enable self-improving reasoners, or, four habits of highly effective stars.
\newblock \emph{arXiv preprint arXiv:2503.01307}, 2025.

\bibitem[Gao et~al.(2024)Gao, Song, Yang, Cai, Miao, Dong, Li, Ma, Chen, Xu, Tang, Wang, Zan, Quan, Zhang, Sha, Zhang, Ren, Liu, and Chang]{gao2024omnimathuniversalolympiadlevel}
Bofei Gao, Feifan Song, Zhe Yang, Zefan Cai, Yibo Miao, Qingxiu Dong, Lei Li, Chenghao Ma, Liang Chen, Runxin Xu, Zhengyang Tang, Benyou Wang, Daoguang Zan, Shanghaoran Quan, Ge~Zhang, Lei Sha, Yichang Zhang, Xuancheng Ren, Tianyu Liu, and Baobao Chang.
\newblock Omni-math: A universal olympiad level mathematic benchmark for large language models, 2024.
\newblock URL \url{https://arxiv.org/abs/2410.07985}.

\bibitem[Grattafiori et~al.(2024)Grattafiori, Dubey, Jauhri, Pandey, Kadian, Al-Dahle, Letman, Mathur, Schelten, Vaughan, et~al.]{grattafiori2024llama}
Aaron Grattafiori, Abhimanyu Dubey, Abhinav Jauhri, Abhinav Pandey, Abhishek Kadian, Ahmad Al-Dahle, Aiesha Letman, Akhil Mathur, Alan Schelten, Alex Vaughan, et~al.
\newblock The llama 3 herd of models.
\newblock \emph{arXiv preprint arXiv:2407.21783}, 2024.

\bibitem[Guan et~al.(2024)Guan, Joglekar, Wallace, Jain, Barak, Helyar, Dias, Vallone, Ren, Wei, et~al.]{guan2024deliberative}
Melody~Y Guan, Manas Joglekar, Eric Wallace, Saachi Jain, Boaz Barak, Alec Helyar, Rachel Dias, Andrea Vallone, Hongyu Ren, Jason Wei, et~al.
\newblock Deliberative alignment: Reasoning enables safer language models.
\newblock \emph{arXiv preprint arXiv:2412.16339}, 2024.

\bibitem[Guan et~al.(2025)Guan, Zhang, Liu, Shang, Sun, Zhu, Yang, and Yang]{guan2025rstar}
Xinyu Guan, Li~Lyna Zhang, Yifei Liu, Ning Shang, Youran Sun, Yi~Zhu, Fan Yang, and Mao Yang.
\newblock rstar-math: Small llms can master math reasoning with self-evolved deep thinking.
\newblock \emph{arXiv preprint arXiv:2501.04519}, 2025.

\bibitem[Guo et~al.(2025)Guo, Yang, Zhang, Song, Zhang, Xu, Zhu, Ma, Wang, Bi, et~al.]{Deepseek_R1}
Daya Guo, Dejian Yang, Haowei Zhang, Junxiao Song, Ruoyu Zhang, Runxin Xu, Qihao Zhu, Shirong Ma, Peiyi Wang, Xiao Bi, et~al.
\newblock Deepseek-r1: Incentivizing reasoning capability in llms via reinforcement learning.
\newblock \emph{arXiv preprint arXiv:2501.12948}, 2025.

\bibitem[He et~al.(2025{\natexlab{a}})He, Liu, Liu, Yan, Wang, Cheng, Zhang, Zhang, Xu, Shen, Li, Zeng, Wei, Cheng, An, Liu, and Zhou]{skywork-or1-2025}
Jujie He, Jiacai Liu, Chris~Yuhao Liu, Rui Yan, Chaojie Wang, Peng Cheng, Xiaoyu Zhang, Fuxiang Zhang, Jiacheng Xu, Wei Shen, Siyuan Li, Liang Zeng, Tianwen Wei, Cheng Cheng, Bo~An, Yang Liu, and Yahui Zhou.
\newblock Skywork open reaonser series.
\newblock \url{https://capricious-hydrogen-41c.notion.site/Skywork-Open-Reaonser-Series-1d0bc9ae823a80459b46c149e4f51680}, 2025{\natexlab{a}}.
\newblock Notion Blog.

\bibitem[He et~al.(2025{\natexlab{b}})He, Liang, Xu, Liu, Chen, Wang, Song, Yu, Liang, Wang, Zhang, Wang, Tu, Mi, and Yu]{he2025deepmath103klargescalechallengingdecontaminated}
Zhiwei He, Tian Liang, Jiahao Xu, Qiuzhi Liu, Xingyu Chen, Yue Wang, Linfeng Song, Dian Yu, Zhenwen Liang, Wenxuan Wang, Zhuosheng Zhang, Rui Wang, Zhaopeng Tu, Haitao Mi, and Dong Yu.
\newblock Deepmath-103k: A large-scale, challenging, decontaminated, and verifiable mathematical dataset for advancing reasoning, 2025{\natexlab{b}}.
\newblock URL \url{https://arxiv.org/abs/2504.11456}.

\bibitem[Hendrycks et~al.(2021)Hendrycks, Burns, Kadavath, Arora, Basart, Tang, Song, and Steinhardt]{hendrycks2021measuringmathematicalproblemsolving}
Dan Hendrycks, Collin Burns, Saurav Kadavath, Akul Arora, Steven Basart, Eric Tang, Dawn Song, and Jacob Steinhardt.
\newblock Measuring mathematical problem solving with the math dataset, 2021.
\newblock URL \url{https://arxiv.org/abs/2103.03874}.

\bibitem[Hochlehnert et~al.(2025{\natexlab{a}})Hochlehnert, Bhatnagar, Udandarao, Albanie, Prabhu, and Bethge]{hochlehnert2025sober}
Andreas Hochlehnert, Hardik Bhatnagar, Vishaal Udandarao, Samuel Albanie, Ameya Prabhu, and Matthias Bethge.
\newblock A sober look at progress in language model reasoning: Pitfalls and paths to reproducibility.
\newblock \emph{arXiv preprint arXiv:2504.07086}, 2025{\natexlab{a}}.

\bibitem[Hochlehnert et~al.(2025{\natexlab{b}})Hochlehnert, Bhatnagar, Udandarao, Prabhu, and Bethge]{curatedthoughts}
Andreas Hochlehnert, Hardik Bhatnagar, Vishaal Udandarao, Ameya Prabhu, and Matthias Bethge.
\newblock Curatedthoughts: Data curation for rl training datasets, 2025{\natexlab{b}}.
\newblock URL \url{https://huggingface.co/datasets/bethgelab/CuratedThoughts}.

\bibitem[Hu(2025)]{reinforce++}
Jian Hu.
\newblock Reinforce++: A simple and efficient approach for aligning large language models.
\newblock \emph{arXiv preprint arXiv:2501.03262}, 2025.

\bibitem[Hu et~al.(2025)Hu, Zhang, Han, Jiang, and Xiangyu~Zhang]{OpenReasonerZero2025}
Jingcheng Hu, Yinmin Zhang, Qi~Han, Daxin Jiang, and Heung-Yeung~Shum Xiangyu~Zhang.
\newblock Open-reasoner-zero: An open source approach to scaling reinforcement learning on the base model.
\newblock \url{https://github.com/Open-Reasoner-Zero/Open-Reasoner-Zero}, 2025.

\bibitem[Huang et~al.(2025{\natexlab{a}})Huang, Madala, Sidhu, Niu, Hockenmaier, and Zhang]{huang2025rag}
Jerry Huang, Siddarth Madala, Risham Sidhu, Cheng Niu, Julia Hockenmaier, and Tong Zhang.
\newblock Rag-rl: Advancing retrieval-augmented generation via rl and curriculum learning.
\newblock \emph{arXiv preprint arXiv:2503.12759}, 2025{\natexlab{a}}.

\bibitem[Huang et~al.(2025{\natexlab{b}})Huang, Hu, Ilhan, Tekin, Yahn, Xu, and Liu]{huang2025safety}
Tiansheng Huang, Sihao Hu, Fatih Ilhan, Selim~Furkan Tekin, Zachary Yahn, Yichang Xu, and Ling Liu.
\newblock Safety tax: Safety alignment makes your large reasoning models less reasonable.
\newblock \emph{arXiv preprint arXiv:2503.00555}, 2025{\natexlab{b}}.

\bibitem[HuggingFace(2025)]{openr1}
HuggingFace.
\newblock Open r1: A fully open reproduction of deepseek-r1, January 2025.
\newblock URL \url{https://github.com/huggingface/open-r1}.

\bibitem[Internet(2025)]{2025iithought}
Intelligent Internet.
\newblock Ii-thought : A large-scale, high-quality reasoning dataset.
\newblock \url{https://ii.inc/web/blog/post/ii-thought}, 2025.

\bibitem[Jaech et~al.(2024)Jaech, Kalai, Lerer, Richardson, El-Kishky, Low, Helyar, Madry, Beutel, Carney, et~al.]{openai-o1}
Aaron Jaech, Adam Kalai, Adam Lerer, Adam Richardson, Ahmed El-Kishky, Aiden Low, Alec Helyar, Aleksander Madry, Alex Beutel, Alex Carney, et~al.
\newblock Openai o1 system card.
\newblock \emph{arXiv preprint arXiv:2412.16720}, 2024.

\bibitem[Jha et~al.(2024)Jha, Jana, Suresh, Arora, and Ganesh]{jha2024rlsf}
Piyush Jha, Prithwish Jana, Pranavkrishna Suresh, Arnav Arora, and Vijay Ganesh.
\newblock Rlsf: Reinforcement learning via symbolic feedback.
\newblock \emph{arXiv preprint arXiv:2405.16661}, 2024.

\bibitem[Jiang et~al.(2025)Jiang, Xu, Li, Niu, Xiang, Li, Lin, and Poovendran]{jiang2025safechain}
Fengqing Jiang, Zhangchen Xu, Yuetai Li, Luyao Niu, Zhen Xiang, Bo~Li, Bill~Yuchen Lin, and Radha Poovendran.
\newblock Safechain: Safety of language models with long chain-of-thought reasoning capabilities.
\newblock \emph{arXiv preprint arXiv:2502.12025}, 2025.

\bibitem[Kaddour et~al.(2023)Kaddour, Harris, Mozes, Bradley, Raileanu, and McHardy]{kaddour2023challenges}
Jean Kaddour, Joshua Harris, Maximilian Mozes, Herbie Bradley, Roberta Raileanu, and Robert McHardy.
\newblock Challenges and applications of large language models.
\newblock \emph{arXiv preprint arXiv:2307.10169}, 2023.

\bibitem[Kimi-Team et~al.(2025)Kimi-Team, Du, Gao, Xing, Jiang, Chen, Li, Xiao, Du, Liao, Tang, Wang, Zhang, Yuan, Lu, Tang, Sung, Wei, Lai, Guo, Zhu, Ding, Hu, Yang, Zhang, Yao, Zhao, Lu, Li, Yu, Gao, Zheng, Yuan, Chen, Guo, Su, Wang, Zhao, Zhang, Liu, Yan, Wu, Shi, Ye, Yu, Dong, Zhang, Ma, Pan, Gong, Liu, Ma, Wei, Cao, Huang, Jiang, Gao, Xiong, He, Huang, Wu, He, Wei, Jia, Wu, Xu, Zu, Zhou, Pan, Charles, Li, Hu, Liu, Chen, Wang, Liu, Qin, Liu, Yang, Bao, Du, Wu, Wang, Zhou, Wang, Li, Zhu, Zhang, Wang, Yang, Huang, Huang, Xu, and Yang]{KIMI_scaling_RL}
Kimi-Team, Angang Du, Bofei Gao, Bowei Xing, Changjiu Jiang, Cheng Chen, Cheng Li, Chenjun Xiao, Chenzhuang Du, Chonghua Liao, Chuning Tang, Congcong Wang, Dehao Zhang, Enming Yuan, Enzhe Lu, Fengxiang Tang, Flood Sung, Guangda Wei, Guokun Lai, Haiqing Guo, Han Zhu, Hao Ding, Hao Hu, Hao Yang, Hao Zhang, Haotian Yao, Haotian Zhao, Haoyu Lu, Haoze Li, Haozhen Yu, Hongcheng Gao, Huabin Zheng, Huan Yuan, Jia Chen, Jianhang Guo, Jianlin Su, Jianzhou Wang, Jie Zhao, Jin Zhang, Jingyuan Liu, Junjie Yan, Junyan Wu, Lidong Shi, Ling Ye, Longhui Yu, Mengnan Dong, Neo Zhang, Ningchen Ma, Qiwei Pan, Qucheng Gong, Shaowei Liu, Shengling Ma, Shupeng Wei, Sihan Cao, Siying Huang, Tao Jiang, Weihao Gao, Weimin Xiong, Weiran He, Weixiao Huang, Wenhao Wu, Wenyang He, Xianghui Wei, Xianqing Jia, Xingzhe Wu, Xinran Xu, Xinxing Zu, Xinyu Zhou, Xuehai Pan, Y.~Charles, Yang Li, Yangyang Hu, Yangyang Liu, Yanru Chen, Yejie Wang, Yibo Liu, Yidao Qin, Yifeng Liu, Ying Yang, Yiping Bao, Yulun Du, Yuxin Wu, Yuzhi Wang, Zaida Zhou,
  Zhaoji Wang, Zhaowei Li, Zhen Zhu, Zheng Zhang, Zhexu Wang, Zhilin Yang, Zhiqi Huang, Zihao Huang, Ziyao Xu, and Zonghan Yang.
\newblock Kimi k1.5: Scaling reinforcement learning with llms, 2025.
\newblock URL \url{https://arxiv.org/abs/2501.12599}.

\bibitem[Kumar et~al.(2025)Kumar, Roh, Naseh, Karpinska, Iyyer, Houmansadr, and Bagdasarian]{kumar2025overthink}
Abhinav Kumar, Jaechul Roh, Ali Naseh, Marzena Karpinska, Mohit Iyyer, Amir Houmansadr, and Eugene Bagdasarian.
\newblock Overthink: Slowdown attacks on reasoning llms.
\newblock \emph{arXiv e-prints}, pp.\  arXiv--2502, 2025.

\bibitem[Kuo et~al.(2025)Kuo, Zhang, Ding, Wang, DiValentin, Bao, Wei, Juan, Li, and Chen]{kuo2025h}
Martin Kuo, Jianyi Zhang, Aolin Ding, Qinsi Wang, Louis DiValentin, Yujia Bao, Wei Wei, Da-Cheng Juan, Hai Li, and Yiran Chen.
\newblock H-cot: Hijacking the chain-of-thought safety reasoning mechanism to jailbreak large reasoning models, including openai o1/o3, deepseek-r1, and gemini 2.0 flash thinking.
\newblock \emph{arXiv preprint arXiv:2502.12893}, 2025.

\bibitem[Kydlíček(2024)]{math_verify}
Hynek Kydlíček.
\newblock Math-verify: Math verification library, 2024.
\newblock URL \url{https://github.com/huggingface/math-verify}.
\newblock If you use this software, please cite it using the metadata from this file.

\bibitem[Lai \& Nissim(2024)Lai and Nissim]{lai2024mcot}
Huiyuan Lai and Malvina Nissim.
\newblock mcot: Multilingual instruction tuning for reasoning consistency in language models.
\newblock \emph{arXiv preprint arXiv:2406.02301}, 2024.

\bibitem[Leike \& Sutskever(2023)Leike and Sutskever]{leike2023superalignment}
Jan Leike and Ilya Sutskever.
\newblock Introducing superalignment.
\newblock \url{https://openai.com/blog/introducing-superalignment}, 2023.
\newblock Accessed: 2024-04-01.

\bibitem[LG-Research et~al.(2025)LG-Research, Bae, Choi, Choi, Choi, Choi, Hong, Hwang, Jeon, Jeon, et~al.]{research2025exaone}
LG-Research, Kyunghoon Bae, Eunbi Choi, Kibong Choi, Stanley~Jungkyu Choi, Yemuk Choi, Seokhee Hong, Junwon Hwang, Hyojin Jeon, Kijeong Jeon, et~al.
\newblock Exaone deep: Reasoning enhanced language models.
\newblock \emph{arXiv preprint arXiv:2503.12524}, 2025.

\bibitem[Li et~al.(2025{\natexlab{a}})Li, Liu, Dinkel, Niu, Zhang, and Luan]{li2025reinforcement}
Gang Li, Jizhong Liu, Heinrich Dinkel, Yadong Niu, Junbo Zhang, and Jian Luan.
\newblock Reinforcement learning outperforms supervised fine-tuning: A case study on audio question answering.
\newblock \emph{arXiv preprint arXiv:2503.11197}, 2025{\natexlab{a}}.

\bibitem[LI et~al.(2024)LI, Beeching, Tunstall, Lipkin, Soletskyi, Huang, Rasul, Yu, Jiang, Shen, Qin, Dong, Zhou, Fleureau, Lample, and Polu]{numina_math_datasets}
Jia LI, Edward Beeching, Lewis Tunstall, Ben Lipkin, Roman Soletskyi, Shengyi~Costa Huang, Kashif Rasul, Longhui Yu, Albert Jiang, Ziju Shen, Zihan Qin, Bin Dong, Li~Zhou, Yann Fleureau, Guillaume Lample, and Stanislas Polu.
\newblock Numinamath.
\newblock \url{[https://huggingface.co/AI-MO/NuminaMath-1.5](https://github.com/project-numina/aimo-progress-prize/blob/main/report/numina_dataset.pdf)}, 2024.

\bibitem[Li et~al.(2023)Li, Fu, Zhang, Huang, Sun, Lyu, Liu, Jin, and Li]{li2023taco}
Rongao Li, Jie Fu, Bo-Wen Zhang, Tao Huang, Zhihong Sun, Chen Lyu, Guang Liu, Zhi Jin, and Ge~Li.
\newblock Taco: Topics in algorithmic code generation dataset.
\newblock \emph{arXiv preprint arXiv:2312.14852}, 2023.

\bibitem[Li et~al.(2024{\natexlab{a}})Li, Yang, Cheng, Liu, Yu, Yang, and Lam]{li2024large}
Siheng Li, Cheng Yang, Zesen Cheng, Lemao Liu, Mo~Yu, Yujiu Yang, and Wai Lam.
\newblock Large language models can self-improve in long-context reasoning.
\newblock \emph{arXiv preprint arXiv:2411.08147}, 2024{\natexlab{a}}.

\bibitem[Li et~al.(2024{\natexlab{b}})Li, Xu, Zhao, Jiao, Joty, and Bing]{crplanner}
Xingxuan Li, Weiwen Xu, Ruochen Zhao, Fangkai Jiao, Shafiq Joty, and Lidong Bing.
\newblock Can we further elicit reasoning in llms? critic-guided planning with retrieval-augmentation for solving challenging tasks.
\newblock \emph{arXiv preprint arXiv:2410.01428}, 2024{\natexlab{b}}.

\bibitem[Li et~al.(2025{\natexlab{b}})Li, Zou, and Liu]{li2025limr}
Xuefeng Li, Haoyang Zou, and Pengfei Liu.
\newblock Limr: Less is more for rl scaling.
\newblock \emph{arXiv preprint arXiv:2502.11886}, 2025{\natexlab{b}}.

\bibitem[Li et~al.(2025{\natexlab{c}})Li, Li, Kosuga, and Bian]{li2025output}
Xuying Li, Zhuo Li, Yuji Kosuga, and Victor Bian.
\newblock Output length effect on deepseek-r1's safety in forced thinking.
\newblock \emph{arXiv preprint arXiv:2503.01923}, 2025{\natexlab{c}}.

\bibitem[Li et~al.(2025{\natexlab{d}})Li, Yue, Xu, Jiang, Niu, Lin, Ramasubramanian, and Poovendran]{li2025small}
Yuetai Li, Xiang Yue, Zhangchen Xu, Fengqing Jiang, Luyao Niu, Bill~Yuchen Lin, Bhaskar Ramasubramanian, and Radha Poovendran.
\newblock Small models struggle to learn from strong reasoners.
\newblock \emph{arXiv preprint arXiv:2502.12143}, 2025{\natexlab{d}}.

\bibitem[Liang et~al.(2024)Liang, Xu, Hong, Shang, Wang, Fu, and Liu]{liang2024survey}
Zijing Liang, Yanjie Xu, Yifan Hong, Penghui Shang, Qi~Wang, Qiang Fu, and Ke~Liu.
\newblock A survey of multimodel large language models.
\newblock In \emph{Proceedings of the 3rd International Conference on Computer, Artificial Intelligence and Control Engineering}, pp.\  405--409, 2024.

\bibitem[Lin et~al.(2025{\natexlab{a}})Lin, Bras, Richardson, Sabharwal, Poovendran, Clark, and Choi]{lin2025zebralogicscalinglimitsllms}
Bill~Yuchen Lin, Ronan~Le Bras, Kyle Richardson, Ashish Sabharwal, Radha Poovendran, Peter Clark, and Yejin Choi.
\newblock Zebralogic: On the scaling limits of llms for logical reasoning, 2025{\natexlab{a}}.
\newblock URL \url{https://arxiv.org/abs/2502.01100}.

\bibitem[Lin et~al.(2025{\natexlab{b}})Lin, Lin, Xie, and Ji]{lin2025cppo}
Zhihang Lin, Mingbao Lin, Yuan Xie, and Rongrong Ji.
\newblock Cppo: Accelerating the training of group relative policy optimization-based reasoning models, 2025{\natexlab{b}}.

\bibitem[Liu \& Zhang(2025)Liu and Zhang]{code-r1}
Jiawei Liu and Lingming Zhang.
\newblock Code-r1: Reproducing r1 for code with reliable rewards.
\newblock \url{https://github.com/ganler/code-r1}, 2025.

\bibitem[Liu et~al.(2024)Liu, Li, Zhang, Zhou, Cheng, and He]{liu2024diving}
Wei Liu, Junlong Li, Xiwen Zhang, Fan Zhou, Yu~Cheng, and Junxian He.
\newblock Diving into self-evolving training for multimodal reasoning.
\newblock \emph{arXiv preprint arXiv:2412.17451}, 2024.

\bibitem[Liu et~al.(2025{\natexlab{a}})Liu, Gao, Zhai, Xia, Wu, Xue, Chen, Kawaguchi, Zhang, and Hooi]{liu2025guardreasoner}
Yue Liu, Hongcheng Gao, Shengfang Zhai, Jun Xia, Tianyi Wu, Zhiwei Xue, Yulin Chen, Kenji Kawaguchi, Jiaheng Zhang, and Bryan Hooi.
\newblock Guardreasoner: Towards reasoning-based llm safeguards.
\newblock \emph{arXiv preprint arXiv:2501.18492}, 2025{\natexlab{a}}.

\bibitem[Liu et~al.(2025{\natexlab{b}})Liu, Chen, Li, Pang, Du, and Lin]{liu2025there}
Zichen Liu, Changyu Chen, Wenjun Li, Tianyu Pang, Chao Du, and Min Lin.
\newblock There may not be aha moment in r1-zero-like training — a pilot study.
\newblock \url{https://oatllm.notion.site/oat-zero}, 2025{\natexlab{b}}.
\newblock Notion Blog.

\bibitem[Liu et~al.(2025{\natexlab{c}})Liu, Chen, Li, Qi, Pang, Du, Lee, and Lin]{DrGRPO}
Zichen Liu, Changyu Chen, Wenjun Li, Penghui Qi, Tianyu Pang, Chao Du, Wee~Sun Lee, and Min Lin.
\newblock Understanding r1-zero-like training: A critical perspective.
\newblock \emph{arXiv preprint arXiv:2503.20783}, 2025{\natexlab{c}}.

\bibitem[Luo et~al.(2025{\natexlab{a}})Luo, Tan, Huang, Shi, Xin, Cai, Patel, Ariyak, Wu, Zhang, Li, and Raluca Ada~Popa]{deepcoder2025}
Michael Luo, Sijun Tan, Roy Huang, Xiaoxiang Shi, Rachel Xin, Colin Cai, Ameen Patel, Alpay Ariyak, Qingyang Wu, Ce~Zhang, Li~Erran Li, and Ion~Stoica Raluca Ada~Popa.
\newblock Deepcoder: A fully open-source 14b coder at o3-mini level.
\newblock \url{https://pretty-radio-b75.notion.site/DeepCoder-A-Fully-Open-Source-14B-Coder-at-O3-mini-Level-1cf81902c14680b3bee5eb349a512a51}, 2025{\natexlab{a}}.
\newblock Notion Blog.

\bibitem[Luo et~al.(2025{\natexlab{b}})Luo, Tan, Wong, Shi, Tang, Roongta, Cai, Luo, Zhang, Li, Popa, and Stoica]{deepscaler2025}
Michael Luo, Sijun Tan, Justin Wong, Xiaoxiang Shi, William~Y. Tang, Manan Roongta, Colin Cai, Jeffrey Luo, Tianjun Zhang, Li~Erran Li, Raluca~Ada Popa, and Ion Stoica.
\newblock Deepscaler: Surpassing o1-preview with a 1.5b model by scaling rl.
\newblock \url{https://pretty-radio-b75.notion.site/DeepScaleR-Surpassing-O1-Preview-with-a-1-5B-Model-by-Scaling-RL-19681902c1468005bed8ca303013a4e2}, 2025{\natexlab{b}}.
\newblock Notion Blog.

\bibitem[Lyu et~al.(2025)Lyu, Gao, Gu, Zhang, Gao, Liu, Wang, Li, Zhao, Huang, Cao, Liu, Liu, Liu, Zhang, Lin, and Chen]{lyu2025exploringlimitoutcomereward}
Chengqi Lyu, Songyang Gao, Yuzhe Gu, Wenwei Zhang, Jianfei Gao, Kuikun Liu, Ziyi Wang, Shuaibin Li, Qian Zhao, Haian Huang, Weihan Cao, Jiangning Liu, Hongwei Liu, Junnan Liu, Songyang Zhang, Dahua Lin, and Kai Chen.
\newblock Exploring the limit of outcome reward for learning mathematical reasoning, 2025.
\newblock URL \url{https://arxiv.org/abs/2502.06781}.

\bibitem[Ma et~al.(2025{\natexlab{a}})Ma, Wang, Liu, Liu, Chen, Zhang, Zhou, Du, and Li]{ma2025s}
Ruotian Ma, Peisong Wang, Cheng Liu, Xingyan Liu, Jiaqi Chen, Bang Zhang, Xin Zhou, Nan Du, and Jia Li.
\newblock S $^{2}$ r: Teaching llms to self-verify and self-correct via reinforcement learning.
\newblock \emph{arXiv preprint arXiv:2502.12853}, 2025{\natexlab{a}}.

\bibitem[Ma et~al.(2025{\natexlab{b}})Ma, Chen, Wang, Chng, and Chen]{ma2025audio}
Ziyang Ma, Zhuo Chen, Yuping Wang, Eng~Siong Chng, and Xie Chen.
\newblock Audio-cot: Exploring chain-of-thought reasoning in large audio language model.
\newblock \emph{arXiv preprint arXiv:2501.07246}, 2025{\natexlab{b}}.

\bibitem[Mattern et~al.(2025)Mattern, Jaghouar, Basra, Straube, Ferrante, Gabriel, Ong, Weisser, and Hagemann]{2025synthetic1}
Justus Mattern, Sami Jaghouar, Manveer Basra, Jannik Straube, Matthew~Di Ferrante, Felix Gabriel, Jack~Min Ong, Vincent Weisser, and Johannes Hagemann.
\newblock Synthetic-1: Two million collaboratively generated reasoning traces from deepseek-r1, 2025.
\newblock URL \url{https://www.primeintellect.ai/blog/synthetic-1-release}.

\bibitem[Muennighoff et~al.(2025)Muennighoff, Yang, Shi, Li, Fei-Fei, Hajishirzi, Zettlemoyer, Liang, Candès, and Hashimoto]{muennighoff2025s1simpletesttimescaling}
Niklas Muennighoff, Zitong Yang, Weijia Shi, Xiang~Lisa Li, Li~Fei-Fei, Hannaneh Hajishirzi, Luke Zettlemoyer, Percy Liang, Emmanuel Candès, and Tatsunori Hashimoto.
\newblock s1: Simple test-time scaling, 2025.
\newblock URL \url{https://arxiv.org/abs/2501.19393}.

\bibitem[Nguyen et~al.(2023)Nguyen, Van~Nguyen, Lai, Man, Ngo, Dernoncourt, Rossi, and Nguyen]{nguyen2023culturax}
Thuat Nguyen, Chien Van~Nguyen, Viet~Dac Lai, Hieu Man, Nghia~Trung Ngo, Franck Dernoncourt, Ryan~A Rossi, and Thien~Huu Nguyen.
\newblock Culturax: A cleaned, enormous, and multilingual dataset for large language models in 167 languages.
\newblock \emph{arXiv preprint arXiv:2309.09400}, 2023.

\bibitem[Nguyen et~al.(2025)Nguyen, Muller, Yu, Costa-Jussa, Elbayad, Popuri, Ropers, Duquenne, Algayres, Mavlyutov, et~al.]{nguyen2025spirit}
Tu~Anh Nguyen, Benjamin Muller, Bokai Yu, Marta~R Costa-Jussa, Maha Elbayad, Sravya Popuri, Christophe Ropers, Paul-Ambroise Duquenne, Robin Algayres, Ruslan Mavlyutov, et~al.
\newblock Spirit-lm: Interleaved spoken and written language model.
\newblock \emph{Transactions of the Association for Computational Linguistics}, 13:\penalty0 30--52, 2025.

\bibitem[OpenAI(2024)]{openai2024gpt4o}
OpenAI.
\newblock Gpt-4o system card, August 2024.
\newblock URL \url{https://openai.com/index/gpt-4o-system-card/}.
\newblock Accessed: 2025-05-01.

\bibitem[OpenAI(2025)]{openai-o3}
OpenAI.
\newblock Introducing openai o3 and o4-mini, 2025.
\newblock URL \url{https://openai.com/index/introducing-o3-and-o4-mini/}.

\bibitem[OpenThoughts-Team(2025)]{openthoughts}
OpenThoughts-Team.
\newblock {Open Thoughts}.
\newblock \url{https://open-thoughts.ai}, January 2025.

\bibitem[Ouyang et~al.(2022)Ouyang, Wu, Jiang, Almeida, Wainwright, Mishkin, Zhang, Agarwal, Slama, Ray, Schulman, Hilton, Kelton, Miller, Simens, Askell, Welinder, Christiano, Leike, and Lowe]{RLHF}
Long Ouyang, Jeff Wu, Xu~Jiang, Diogo Almeida, Carroll~L. Wainwright, Pamela Mishkin, Chong Zhang, Sandhini Agarwal, Katarina Slama, Alex Ray, John Schulman, Jacob Hilton, Fraser Kelton, Luke~E. Miller, Maddie Simens, Amanda Askell, Peter Welinder, Paul~Francis Christiano, Jan Leike, and Ryan~J. Lowe.
\newblock Training language models to follow instructions with human feedback.
\newblock \emph{ArXiv}, abs/2203.02155, 2022.
\newblock URL \url{https://api.semanticscholar.org/CorpusID:246426909}.

\bibitem[Pan et~al.(2022)Pan, Bhatia, and Steinhardt]{pan2022effects}
Alexander Pan, Kush Bhatia, and Jacob Steinhardt.
\newblock The effects of reward misspecification: Mapping and mitigating misaligned models.
\newblock \emph{arXiv preprint arXiv:2201.03544}, 2022.

\bibitem[Pan et~al.(2025)Pan, Zhang, Wang, Yuan, Peng, and Suhr]{tinyzero}
Jiayi Pan, Junjie Zhang, Xingyao Wang, Lifan Yuan, Hao Peng, and Alane Suhr.
\newblock Tinyzero.
\newblock \url{https://github.com/Jiayi-Pan/TinyZero}, 2025.
\newblock Accessed: 2025-01-24.

\bibitem[Penedo et~al.(2025)Penedo, Lozhkov, Kydlíček, Allal, Beeching, Lajarín, Gallouédec, Habib, Tunstall, and von Werra]{penedo2025codeforces}
Guilherme Penedo, Anton Lozhkov, Hynek Kydlíček, Loubna~Ben Allal, Edward Beeching, Agustín~Piqueres Lajarín, Quentin Gallouédec, Nathan Habib, Lewis Tunstall, and Leandro von Werra.
\newblock Codeforces.
\newblock \url{https://huggingface.co/datasets/open-r1/codeforces}, 2025.

\bibitem[Qin et~al.(2024)Qin, Chen, Zhou, Chen, Li, Liao, Li, Che, and Yu]{qin2024multilingual}
Libo Qin, Qiguang Chen, Yuhang Zhou, Zhi Chen, Yinghui Li, Lizi Liao, Min Li, Wanxiang Che, and Philip~S Yu.
\newblock Multilingual large language model: A survey of resources, taxonomy and frontiers.
\newblock \emph{arXiv preprint arXiv:2404.04925}, 2024.

\bibitem[Qwen-Team(2024)]{qwq-32b-preview}
Qwen-Team.
\newblock Qwq: Reflect deeply on the boundaries of the unknown, November 2024.
\newblock URL \url{https://qwenlm.github.io/blog/qwq-32b-preview/}.

\bibitem[Qwen-Team(2025{\natexlab{a}})]{qwen3}
Qwen-Team.
\newblock Qwen3: Think deeper, act faster, 2025{\natexlab{a}}.
\newblock URL \url{https://qwenlm.github.io/blog/qwen3/}.

\bibitem[Qwen-Team(2025{\natexlab{b}})]{qwq-32b}
Qwen-Team.
\newblock Qwq-32b: Embracing the power of reinforcement learning, 2025{\natexlab{b}}.
\newblock URL \url{https://qwenlm.github.io/blog/qwq-32b/}.

\bibitem[Rafailov et~al.(2023)Rafailov, Sharma, Mitchell, Ermon, Manning, and Finn]{Rafailov2023DirectPO}
Rafael Rafailov, Archit Sharma, Eric Mitchell, Stefano Ermon, Christopher~D. Manning, and Chelsea Finn.
\newblock Direct preference optimization: Your language model is secretly a reward model.
\newblock In \emph{NeurIPS}, 2023.

\bibitem[Razin et~al.(2025)Razin, Wang, Strauss, Wei, Lee, and Arora]{razin2025makes}
Noam Razin, Zixuan Wang, Hubert Strauss, Stanley Wei, Jason~D Lee, and Sanjeev Arora.
\newblock What makes a reward model a good teacher? an optimization perspective.
\newblock \emph{arXiv preprint arXiv:2503.15477}, 2025.

\bibitem[RDKit(2025)]{RDKit}
RDKit.
\newblock Rdkit: Open-source cheminformatics software, 2025.
\newblock URL \url{https://www.rdkit.org}.

\bibitem[Rein et~al.(2024)Rein, Hou, Stickland, Petty, Pang, Dirani, Michael, and Bowman]{rein2024gpqa}
David Rein, Betty~Li Hou, Asa~Cooper Stickland, Jackson Petty, Richard~Yuanzhe Pang, Julien Dirani, Julian Michael, and Samuel~R. Bowman.
\newblock {GPQA}: A graduate-level google-proof q\&a benchmark.
\newblock In \emph{First Conference on Language Modeling}, 2024.
\newblock URL \url{https://openreview.net/forum?id=Ti67584b98}.

\bibitem[Sabbaghi et~al.(2025)Sabbaghi, Kassianik, Pappas, Singer, Karbasi, and Hassani]{sabbaghi2025adversarial}
Mahdi Sabbaghi, Paul Kassianik, George Pappas, Yaron Singer, Amin Karbasi, and Hamed Hassani.
\newblock Adversarial reasoning at jailbreaking time.
\newblock \emph{arXiv preprint arXiv:2502.01633}, 2025.

\bibitem[Schmid(2025)]{mini-r1}
Philipp Schmid.
\newblock Mini-r1: Reproduce deepseek r1 „aha moment“ a rl tutorial.
\newblock \url{https://huggingface.co/blog/open-r1/mini-r1-contdown-game}, 2025.

\bibitem[Schulman.(2020)]{KL_K3}
J.~Schulman.
\newblock Approximating kl divergence, 2020.
\newblock URL \url{http://joschu.net/blog/kl-approx.html}.

\bibitem[Schulman et~al.(2017)Schulman, Wolski, Dhariwal, Radford, and Klimov]{PPO}
John Schulman, Filip Wolski, Prafulla Dhariwal, Alec Radford, and Oleg Klimov.
\newblock Proximal policy optimization algorithms.
\newblock \emph{CoRR}, abs/1707.06347, 2017.
\newblock URL \url{http://arxiv.org/abs/1707.06347}.

\bibitem[Schulman et~al.(2018)Schulman, Moritz, Levine, Jordan, and Abbeel]{GAE}
John Schulman, Philipp Moritz, Sergey Levine, Michael Jordan, and Pieter Abbeel.
\newblock High-dimensional continuous control using generalized advantage estimation, 2018.
\newblock URL \url{https://arxiv.org/abs/1506.02438}.

\bibitem[Shao et~al.(2024)Shao, Wang, Zhu, Xu, Song, Bi, Zhang, Zhang, Li, Wu, et~al.]{GRPO}
Zhihong Shao, Peiyi Wang, Qihao Zhu, Runxin Xu, Junxiao Song, Xiao Bi, Haowei Zhang, Mingchuan Zhang, YK~Li, Y~Wu, et~al.
\newblock Deepseekmath: Pushing the limits of mathematical reasoning in open language models.
\newblock \emph{arXiv preprint arXiv:2402.03300}, 2024.

\bibitem[Song et~al.(2025)Song, Su, Qu, Zhou, and Cheng]{song2025prmbench}
Mingyang Song, Zhaochen Su, Xiaoye Qu, Jiawei Zhou, and Yu~Cheng.
\newblock Prmbench: A fine-grained and challenging benchmark for process-level reward models.
\newblock \emph{arXiv preprint arXiv:2501.03124}, 2025.

\bibitem[Su et~al.(2025)Su, Yu, Song, Li, Mi, Tu, Zhang, and Yu]{su2025crossingrewardbridgeexpanding}
Yi~Su, Dian Yu, Linfeng Song, Juntao Li, Haitao Mi, Zhaopeng Tu, Min Zhang, and Dong Yu.
\newblock Crossing the reward bridge: Expanding rl with verifiable rewards across diverse domains, 2025.
\newblock URL \url{https://arxiv.org/abs/2503.23829}.

\bibitem[Sui et~al.(2025)Sui, Chuang, Wang, Zhang, Zhang, Yuan, Liu, Wen, Shaochen, Zhong, Chen, and Hu]{sui2025stopoverthinkingsurveyefficient}
Yang Sui, Yu-Neng Chuang, Guanchu Wang, Jiamu Zhang, Tianyi Zhang, Jiayi Yuan, Hongyi Liu, Andrew Wen, Shaochen, Zhong, Hanjie Chen, and Xia Hu.
\newblock Stop overthinking: A survey on efficient reasoning for large language models, 2025.
\newblock URL \url{https://arxiv.org/abs/2503.16419}.

\bibitem[Sun et~al.(2025)Sun, Liang, Wei, Yu, Li, Yang, Zhou, and Zhang]{sun2025mm}
Linzhuang Sun, Hao Liang, Jingxuan Wei, Bihui Yu, Tianpeng Li, Fan Yang, Zenan Zhou, and Wentao Zhang.
\newblock Mm-verify: Enhancing multimodal reasoning with chain-of-thought verification.
\newblock \emph{arXiv preprint arXiv:2502.13383}, 2025.

\bibitem[Sutton \& Barto(2018)Sutton and Barto]{Sutton_RL}
Richard~S. Sutton and Andrew~G. Barto.
\newblock \emph{Reinforcement Learning: An Introduction}.
\newblock The MIT Press, second edition, 2018.
\newblock URL \url{http://incompleteideas.net/book/the-book-2nd.html}.

\bibitem[Suzgun et~al.(2022)Suzgun, Scales, Schärli, Gehrmann, Tay, Chung, Chowdhery, Le, Chi, Zhou, and Wei]{bbh-suzgun2022challengingbigbenchtaskschainofthought}
Mirac Suzgun, Nathan Scales, Nathanael Schärli, Sebastian Gehrmann, Yi~Tay, Hyung~Won Chung, Aakanksha Chowdhery, Quoc~V. Le, Ed~H. Chi, Denny Zhou, and Jason Wei.
\newblock Challenging big-bench tasks and whether chain-of-thought can solve them, 2022.
\newblock URL \url{https://arxiv.org/abs/2210.09261}.

\bibitem[Tang et~al.(2025)Tang, Cohen, Zhang, Valko, and Munos]{tang2025rl}
Yunhao Tang, Taco Cohen, David~W Zhang, Michal Valko, and R{\'e}mi Munos.
\newblock Rl-finetuning llms from on-and off-policy data with a single algorithm.
\newblock \emph{arXiv preprint arXiv:2503.19612}, 2025.

\bibitem[Tao et~al.(2024)Tao, Lin, Chen, Li, Wu, Li, Jin, Huang, Tao, and Zhou]{tao2024survey}
Zhengwei Tao, Ting-En Lin, Xiancai Chen, Hangyu Li, Yuchuan Wu, Yongbin Li, Zhi Jin, Fei Huang, Dacheng Tao, and Jingren Zhou.
\newblock A survey on self-evolution of large language models.
\newblock \emph{arXiv preprint arXiv:2404.14387}, 2024.

\bibitem[Taubenfeld et~al.(2024)Taubenfeld, Dover, Reichart, and Goldstein]{taubenfeld2024systematic}
Amir Taubenfeld, Yaniv Dover, Roi Reichart, and Ariel Goldstein.
\newblock Systematic biases in llm simulations of debates.
\newblock \emph{arXiv preprint arXiv:2402.04049}, 2024.

\bibitem[Team et~al.(2023)Team, Anil, Borgeaud, Alayrac, Yu, Soricut, Schalkwyk, Dai, Hauth, Millican, et~al.]{team2023gemini}
Gemini Team, Rohan Anil, Sebastian Borgeaud, Jean-Baptiste Alayrac, Jiahui Yu, Radu Soricut, Johan Schalkwyk, Andrew~M Dai, Anja Hauth, Katie Millican, et~al.
\newblock Gemini: a family of highly capable multimodal models.
\newblock \emph{arXiv preprint arXiv:2312.11805}, 2023.

\bibitem[Team et~al.(2024)Team, Georgiev, Lei, Burnell, Bai, Gulati, Tanzer, Vincent, Pan, Wang, et~al.]{team2024gemini}
Gemini Team, Petko Georgiev, Ving~Ian Lei, Ryan Burnell, Libin Bai, Anmol Gulati, Garrett Tanzer, Damien Vincent, Zhufeng Pan, Shibo Wang, et~al.
\newblock Gemini 1.5: Unlocking multimodal understanding across millions of tokens of context.
\newblock \emph{arXiv preprint arXiv:2403.05530}, 2024.

\bibitem[Trung et~al.(2024)Trung, Zhang, Jie, Sun, Jin, and Li]{trung2024reft}
Luong Trung, Xinbo Zhang, Zhanming Jie, Peng Sun, Xiaoran Jin, and Hang Li.
\newblock Reft: Reasoning with reinforced fine-tuning.
\newblock In \emph{Proceedings of the 62nd Annual Meeting of the Association for Computational Linguistics (Volume 1: Long Papers)}, pp.\  7601--7614, 2024.

\bibitem[Tu et~al.(2025)Tu, Lin, Tian, Zhang, Li, Fu, Xu, He, Lan, Jiang, et~al.]{tu2025enhancing}
Songjun Tu, Jiahao Lin, Xiangyu Tian, Qichao Zhang, Linjing Li, Yuqian Fu, Nan Xu, Wei He, Xiangyuan Lan, Dongmei Jiang, et~al.
\newblock Enhancing llm reasoning with iterative dpo: A comprehensive empirical investigation.
\newblock \emph{arXiv preprint arXiv:2503.12854}, 2025.

\bibitem[Uesato et~al.(2020)Uesato, Kumar, Krakovna, Everitt, Ngo, and Legg]{uesato2020avoiding}
Jonathan Uesato, Ramana Kumar, Victoria Krakovna, Tom Everitt, Richard Ngo, and Shane Legg.
\newblock Avoiding tampering incentives in deep rl via decoupled approval.
\newblock \emph{arXiv preprint arXiv:2011.08827}, 2020.

\bibitem[Wang et~al.(2024{\natexlab{a}})Wang, Wang, Wang, Li, Hovy, and Guo]{wang2024packinganalysispackingappropriate}
Shuhe Wang, Guoyin Wang, Yizhong Wang, Jiwei Li, Eduard Hovy, and Chen Guo.
\newblock Packing analysis: Packing is more appropriate for large models or datasets in supervised fine-tuning, 2024{\natexlab{a}}.
\newblock URL \url{https://arxiv.org/abs/2410.08081}.

\bibitem[Wang et~al.(2025{\natexlab{a}})Wang, Wu, Haddow, and Birch]{wang2025demystifying}
Weixuan Wang, Minghao Wu, Barry Haddow, and Alexandra Birch.
\newblock Demystifying multilingual chain-of-thought in process reward modeling.
\newblock \emph{arXiv preprint arXiv:2502.12663}, 2025{\natexlab{a}}.

\bibitem[Wang et~al.(2025{\natexlab{b}})Wang, Wu, Zhang, Wang, Liu, Luo, and Fei]{wang2025multimodal}
Yaoting Wang, Shengqiong Wu, Yuecheng Zhang, William Wang, Ziwei Liu, Jiebo Luo, and Hao Fei.
\newblock Multimodal chain-of-thought reasoning: A comprehensive survey.
\newblock \emph{arXiv preprint arXiv:2503.12605}, 2025{\natexlab{b}}.

\bibitem[Wang et~al.(2024{\natexlab{b}})Wang, Chen, Han, Lin, Zhao, Liu, Zhai, Yuan, You, and Yang]{wang2024exploring}
Yiqi Wang, Wentao Chen, Xiaotian Han, Xudong Lin, Haiteng Zhao, Yongfei Liu, Bohan Zhai, Jianbo Yuan, Quanzeng You, and Hongxia Yang.
\newblock Exploring the reasoning abilities of multimodal large language models (mllms): A comprehensive survey on emerging trends in multimodal reasoning.
\newblock \emph{arXiv preprint arXiv:2401.06805}, 2024{\natexlab{b}}.

\bibitem[Wang et~al.(2024{\natexlab{c}})Wang, Zhang, Li, Eisenschlos, Perot, Wang, Miculicich, Fujii, Shang, Lee, et~al.]{wang2024chain}
Zilong Wang, Hao Zhang, Chun-Liang Li, Julian~Martin Eisenschlos, Vincent Perot, Zifeng Wang, Lesly Miculicich, Yasuhisa Fujii, Jingbo Shang, Chen-Yu Lee, et~al.
\newblock Chain-of-table: Evolving tables in the reasoning chain for table understanding.
\newblock \emph{arXiv preprint arXiv:2401.04398}, 2024{\natexlab{c}}.

\bibitem[Wei et~al.(2025)Wei, Duchenne, Copet, Carbonneaux, Zhang, Fried, Synnaeve, Singh, and Wang]{wei2025swe}
Yuxiang Wei, Olivier Duchenne, Jade Copet, Quentin Carbonneaux, Lingming Zhang, Daniel Fried, Gabriel Synnaeve, Rishabh Singh, and Sida~I Wang.
\newblock Swe-rl: Advancing llm reasoning via reinforcement learning on open software evolution.
\newblock \emph{arXiv preprint arXiv:2502.18449}, 2025.

\bibitem[Wen et~al.(2025{\natexlab{a}})Wen, Cai, Xiao, He, An, Duan, Du, Liu, Tang, Lv, et~al.]{wen2025light}
Liang Wen, Yunke Cai, Fenrui Xiao, Xin He, Qi~An, Zhenyu Duan, Yimin Du, Junchen Liu, Lifu Tang, Xiaowei Lv, et~al.
\newblock Light-r1: Curriculum sft, dpo and rl for long cot from scratch and beyond.
\newblock \emph{arXiv preprint arXiv:2503.10460}, 2025{\natexlab{a}}.

\bibitem[Wen et~al.(2025{\natexlab{b}})Wen, Zhou, Mo, and Chen]{wen2025thinkguard}
Xiaofei Wen, Wenxuan Zhou, Wenjie~Jacky Mo, and Muhao Chen.
\newblock Thinkguard: Deliberative slow thinking leads to cautious guardrails.
\newblock \emph{arXiv preprint arXiv:2502.13458}, 2025{\natexlab{b}}.

\bibitem[Weng(2024)]{weng2024reward}
Lilian Weng.
\newblock Reward hacking in reinforcement learning.
\newblock \url{https://lilianweng.github.io/posts/2024-11-28-reward-hacking/}, Nov 2024.
\newblock Accessed: 2025-03-27.

\bibitem[Williams(1992)]{reinforce}
Ronald~J. Williams.
\newblock Simple statistical gradient-following algorithms for connectionist reinforcement learning.
\newblock \emph{Mach. Learn.}, 8\penalty0 (3–4):\penalty0 229–256, May 1992.
\newblock ISSN 0885-6125.
\newblock \doi{10.1007/BF00992696}.
\newblock URL \url{https://doi.org/10.1007/BF00992696}.

\bibitem[Wu et~al.(2023)Wu, Gan, Chen, Wan, and Yu]{wu2023multimodal}
Jiayang Wu, Wensheng Gan, Zefeng Chen, Shicheng Wan, and Philip~S Yu.
\newblock Multimodal large language models: A survey.
\newblock In \emph{2023 IEEE International Conference on Big Data (BigData)}, pp.\  2247--2256. IEEE, 2023.

\bibitem[Wu et~al.(2025)Wu, Feng, Zhang, Jin, Che, Wen, and Tao]{wu2025boosting}
Jinyang Wu, Mingkuan Feng, Shuai Zhang, Ruihan Jin, Feihu Che, Zengqi Wen, and Jianhua Tao.
\newblock Boosting multimodal reasoning with mcts-automated structured thinking.
\newblock \emph{arXiv preprint arXiv:2502.02339}, 2025.

\bibitem[Xia et~al.(2025)Xia, Shen, Wang, Liu, Sun, Wu, Hu, and Xu]{xia2025leetcodedatasettemporaldatasetrobust}
Yunhui Xia, Wei Shen, Yan Wang, Jason~Klein Liu, Huifeng Sun, Siyue Wu, Jian Hu, and Xiaolong Xu.
\newblock Leetcodedataset: A temporal dataset for robust evaluation and efficient training of code llms, 2025.
\newblock URL \url{https://arxiv.org/abs/2504.14655}.

\bibitem[Xiao et~al.(2025)Xiao, Yuan, Chen, Li, Liang, Ren, and Honavar]{xiao2025simper}
Teng Xiao, Yige Yuan, Zhengyu Chen, Mingxiao Li, Shangsong Liang, Zhaochun Ren, and Vasant~G Honavar.
\newblock Simper: A minimalist approach to preference alignment without hyperparameters.
\newblock \emph{arXiv preprint arXiv:2502.00883}, 2025.

\bibitem[{Xiaomi LLM-Core Team}(2025)]{xiaomi2025mimo}
{Xiaomi LLM-Core Team}.
\newblock Mimo: Unlocking the reasoning potential of language model – from pretraining to posttraining, 2025.
\newblock URL \url{https://github.com/XiaomiMiMo/MiMo}.

\bibitem[Xie et~al.(2025{\natexlab{a}})Xie, Gao, Ren, Luo, Hong, Dai, Zhou, Qiu, Wu, and Luo]{xie2025logicrlunleashingllmreasoning}
Tian Xie, Zitian Gao, Qingnan Ren, Haoming Luo, Yuqian Hong, Bryan Dai, Joey Zhou, Kai Qiu, Zhirong Wu, and Chong Luo.
\newblock Logic-rl: Unleashing llm reasoning with rule-based reinforcement learning, 2025{\natexlab{a}}.
\newblock URL \url{https://arxiv.org/abs/2502.14768}.

\bibitem[Xie et~al.(2025{\natexlab{b}})Xie, Lin, Liu, Wu, Yan, and Miao]{xie2025audio}
Zhifei Xie, Mingbao Lin, Zihang Liu, Pengcheng Wu, Shuicheng Yan, and Chunyan Miao.
\newblock Audio-reasoner: Improving reasoning capability in large audio language models.
\newblock \emph{arXiv preprint arXiv:2503.02318}, 2025{\natexlab{b}}.

\bibitem[Xiong et~al.(2024)Xiong, Song, Zhao, Wu, Wang, Wang, Li, Peng, and Li]{xiong2024watch}
Weimin Xiong, Yifan Song, Xiutian Zhao, Wenhao Wu, Xun Wang, Ke~Wang, Cheng Li, Wei Peng, and Sujian Li.
\newblock Watch every step! llm agent learning via iterative step-level process refinement.
\newblock In \emph{Proceedings of the 2024 Conference on Empirical Methods in Natural Language Processing}, pp.\  1556--1572, 2024.

\bibitem[Xu et~al.(2025{\natexlab{a}})Xu, Wu, Wang, Li, Zheng, Chen, Hu, Kang, Ji, Zhang, et~al.]{xu2025redstar}
Haotian Xu, Xing Wu, Weinong Wang, Zhongzhi Li, Da~Zheng, Boyuan Chen, Yi~Hu, Shijia Kang, Jiaming Ji, Yingying Zhang, et~al.
\newblock Redstar: Does scaling long-cot data unlock better slow-reasoning systems?
\newblock \emph{arXiv preprint arXiv:2501.11284}, 2025{\natexlab{a}}.

\bibitem[Xu et~al.(2025{\natexlab{b}})Xu, Liu, Yin, Zhou, and Poovendran]{xu2025kodcodediversechallengingverifiable}
Zhangchen Xu, Yang Liu, Yueqin Yin, Mingyuan Zhou, and Radha Poovendran.
\newblock Kodcode: A diverse, challenging, and verifiable synthetic dataset for coding, 2025{\natexlab{b}}.
\newblock URL \url{https://arxiv.org/abs/2503.02951}.

\bibitem[Xuan et~al.(2025)Xuan, Yang, Qi, Zeng, Xiao, Xing, Wang, Li, Li, Yu, et~al.]{xuan2025mmlu}
Weihao Xuan, Rui Yang, Heli Qi, Qingcheng Zeng, Yunze Xiao, Yun Xing, Junjue Wang, Huitao Li, Xin Li, Kunyu Yu, et~al.
\newblock Mmlu-prox: A multilingual benchmark for advanced large language model evaluation.
\newblock \emph{arXiv preprint arXiv:2503.10497}, 2025.

\bibitem[Yang et~al.(2024)Yang, Yang, Zhang, Hui, Zheng, Yu, Li, Liu, Huang, Wei, Lin, Yang, Tu, Zhang, Yang, Yang, Zhou, Lin, Dang, Lu, Bao, Yang, Yu, Li, Xue, Zhang, Zhu, Men, Lin, Li, Xia, Ren, Ren, Fan, Su, Zhang, Wan, Liu, Cui, Zhang, and Qiu]{qwen2.5}
An~Yang, Baosong Yang, Beichen Zhang, Binyuan Hui, Bo~Zheng, Bowen Yu, Chengyuan Li, Dayiheng Liu, Fei Huang, Haoran Wei, Huan Lin, Jian Yang, Jianhong Tu, Jianwei Zhang, Jianxin Yang, Jiaxi Yang, Jingren Zhou, Junyang Lin, Kai Dang, Keming Lu, Keqin Bao, Kexin Yang, Le~Yu, Mei Li, Mingfeng Xue, Pei Zhang, Qin Zhu, Rui Men, Runji Lin, Tianhao Li, Tingyu Xia, Xingzhang Ren, Xuancheng Ren, Yang Fan, Yang Su, Yichang Zhang, Yu~Wan, Yuqiong Liu, Zeyu Cui, Zhenru Zhang, and Zihan Qiu.
\newblock Qwen2.5 technical report.
\newblock \emph{arXiv preprint arXiv:2412.15115}, 2024.

\bibitem[Yao et~al.(2025)Yao, Tong, Wang, Wang, Li, Liu, Teng, and Wang]{yao2025mousetrap}
Yang Yao, Xuan Tong, Ruofan Wang, Yixu Wang, Lujundong Li, Liang Liu, Yan Teng, and Yingchun Wang.
\newblock A mousetrap: Fooling large reasoning models for jailbreak with chain of iterative chaos.
\newblock \emph{arXiv preprint arXiv:2502.15806}, 2025.

\bibitem[Ye et~al.(2025)Ye, Huang, Xiao, Chern, Xia, and Liu]{ye2025limoreasoning}
Yixin Ye, Zhen Huang, Yang Xiao, Ethan Chern, Shijie Xia, and Pengfei Liu.
\newblock Limo: Less is more for reasoning, 2025.
\newblock URL \url{https://arxiv.org/abs/2502.03387}.

\bibitem[Yeo et~al.(2025)Yeo, Tong, Niu, Neubig, and Yue]{yeo2025demystifying}
Edward Yeo, Yuxuan Tong, Morry Niu, Graham Neubig, and Xiang Yue.
\newblock Demystifying long chain-of-thought reasoning in llms.
\newblock \emph{arXiv preprint arXiv:2502.03373}, 2025.

\bibitem[Yi et~al.(2024)Yi, Liu, Sun, Cong, He, Song, Xu, and Li]{yi2024jailbreak}
Sibo Yi, Yule Liu, Zhen Sun, Tianshuo Cong, Xinlei He, Jiaxing Song, Ke~Xu, and Qi~Li.
\newblock Jailbreak attacks and defenses against large language models: A survey.
\newblock \emph{arXiv preprint arXiv:2407.04295}, 2024.

\bibitem[Yu et~al.(2021)Yu, Sun, Yu, and Cardie]{yu2021self}
Dian Yu, Kai Sun, Dong Yu, and Claire Cardie.
\newblock Self-teaching machines to read and comprehend with large-scale multi-subject question-answering data.
\newblock \emph{arXiv preprint arXiv:2102.01226}, 2021.

\bibitem[Yu et~al.(2025)Yu, Zhang, Zhu, Yuan, Zuo, Yue, Fan, Liu, Liu, Liu, et~al.]{DAPO}
Qiying Yu, Zheng Zhang, Ruofei Zhu, Yufeng Yuan, Xiaochen Zuo, Yu~Yue, Tiantian Fan, Gaohong Liu, Lingjun Liu, Xin Liu, et~al.
\newblock Dapo: An open-source llm reinforcement learning system at scale.
\newblock \emph{arXiv preprint arXiv:2503.14476}, 2025.

\bibitem[Yuan et~al.(2025)Yuan, Yue, Zhu, Fan, and Yan]{VC-PPO}
Yufeng Yuan, Yu~Yue, Ruofei Zhu, Tiantian Fan, and Lin Yan.
\newblock What's behind ppo's collapse in long-cot? value optimization holds the secret.
\newblock \emph{arXiv preprint arXiv:2503.01491}, 2025.

\bibitem[Yue et~al.(2025{\natexlab{a}})Yue, Chen, Lu, Zhao, Wang, Yue, Song, and Huang]{yue2025doesreinforcementlearningreally}
Yang Yue, Zhiqi Chen, Rui Lu, Andrew Zhao, Zhaokai Wang, Yang Yue, Shiji Song, and Gao Huang.
\newblock Does reinforcement learning really incentivize reasoning capacity in llms beyond the base model?, 2025{\natexlab{a}}.
\newblock URL \url{https://arxiv.org/abs/2504.13837}.

\bibitem[Yue et~al.(2025{\natexlab{b}})Yue, Yuan, Yu, Zuo, Zhu, Xu, Chen, Wang, Fan, Du, Wei, Liu, Liu, Liu, Lin, Lin, Ma, Zhang, Zhang, Zhang, Zhu, Zhang, Liu, Wang, Wu, and Yan]{yuyue2025vapoefficientreliablereinforcement-vapo}
Yu~Yue, Yufeng Yuan, Qiying Yu, Xiaochen Zuo, Ruofei Zhu, Wenyuan Xu, Jiaze Chen, Chengyi Wang, TianTian Fan, Zhengyin Du, Xiangpeng Wei, Gaohong Liu, Juncai Liu, Lingjun Liu, Haibin Lin, Zhiqi Lin, Bole Ma, Chi Zhang, Mofan Zhang, Wang Zhang, Hang Zhu, Ru~Zhang, Xin Liu, Mingxuan Wang, Yonghui Wu, and Lin Yan.
\newblock Vapo: Efficient and reliable reinforcement learning for advanced reasoning tasks, 2025{\natexlab{b}}.
\newblock URL \url{https://arxiv.org/abs/2504.05118}.

\bibitem[Zeng et~al.(2025{\natexlab{a}})Zeng, Huang, Liu, Liu, He, Ma, and He]{zeng2025simplerlzooinvestigatingtamingzero}
Weihao Zeng, Yuzhen Huang, Qian Liu, Wei Liu, Keqing He, Zejun Ma, and Junxian He.
\newblock Simplerl-zoo: Investigating and taming zero reinforcement learning for open base models in the wild, 2025{\natexlab{a}}.
\newblock URL \url{https://arxiv.org/abs/2503.18892}.

\bibitem[Zeng et~al.(2025{\natexlab{b}})Zeng, Huang, Liu, He, Liu, Ma, and He]{zeng2025simplerl}
Weihao Zeng, Yuzhen Huang, Wei Liu, Keqing He, Qian Liu, Zejun Ma, and Junxian He.
\newblock 7b model and 8k examples: Emerging reasoning with reinforcement learning is both effective and efficient.
\newblock \url{https://hkust-nlp.notion.site/simplerl-reason}, 2025{\natexlab{b}}.
\newblock Notion Blog.

\bibitem[Zhang et~al.(2023)Zhang, Li, and Bing]{zhang2023video}
Hang Zhang, Xin Li, and Lidong Bing.
\newblock Video-llama: An instruction-tuned audio-visual language model for video understanding.
\newblock \emph{arXiv preprint arXiv:2306.02858}, 2023.

\bibitem[Zhang et~al.(2025{\natexlab{a}})Zhang, Yao, Ye, Xiong, and Zhang]{zhang2025dpor1}
Hanning Zhang, Jiarui Yao, Chenlu Ye, Wei Xiong, and Tong Zhang.
\newblock Online-dpo-r1: Unlocking effective reasoning without the ppo overhead.
\newblock \url{https://efficient-unicorn-451.notion.site/Online-DPO-R1-Unlocking-Effective-Reasoning-Without-the-PPO-Overhead-1908b9a70e7b80c3bc83f4cf04b2f175?pvs=4}, 2025{\natexlab{a}}.
\newblock Notion Blog.

\bibitem[Zhang et~al.(2025{\natexlab{b}})Zhang, Wang, Cheng, Zhuang, Lin, Zhang, Wang, Cui, Wang, Peng, Jiang, Kuang, Yin, Wen, Zhang, Chen, and Yu]{zhang2025srpocrossdomainimplementationlargescale}
Xiaojiang Zhang, Jinghui Wang, Zifei Cheng, Wenhao Zhuang, Zheng Lin, Minglei Zhang, Shaojie Wang, Yinghan Cui, Chao Wang, Junyi Peng, Shimiao Jiang, Shiqi Kuang, Shouyu Yin, Chaohang Wen, Haotian Zhang, Bin Chen, and Bing Yu.
\newblock Srpo: A cross-domain implementation of large-scale reinforcement learning on llm, 2025{\natexlab{b}}.
\newblock URL \url{https://arxiv.org/abs/2504.14286}.

\bibitem[Zhao et~al.(2025{\natexlab{a}})Zhao, Wang, Peng, Zhao, Tian, Chen, Ji, and Li]{zhao202514millionopensourcedistilled}
Han Zhao, Haotian Wang, Yiping Peng, Sitong Zhao, Xiaoyu Tian, Shuaiting Chen, Yunjie Ji, and Xiangang Li.
\newblock 1.4 million open-source distilled reasoning dataset to empower large language model training, 2025{\natexlab{a}}.
\newblock URL \url{https://arxiv.org/abs/2503.19633}.

\bibitem[Zhao et~al.(2023)Zhao, Zhou, Li, Tang, Wang, Hou, Min, Zhang, Zhang, Dong, et~al.]{zhao2023survey}
Wayne~Xin Zhao, Kun Zhou, Junyi Li, Tianyi Tang, Xiaolei Wang, Yupeng Hou, Yingqian Min, Beichen Zhang, Junjie Zhang, Zican Dong, et~al.
\newblock A survey of large language models.
\newblock \emph{arXiv preprint arXiv:2303.18223}, 1\penalty0 (2), 2023.

\bibitem[Zhao et~al.(2025{\natexlab{b}})Zhao, Sui, Guo, Hu, Deng, Zhao, Qin, Che, Chua, and Liu]{zhao2025trade}
Weixiang Zhao, Xingyu Sui, Jiahe Guo, Yulin Hu, Yang Deng, Yanyan Zhao, Bing Qin, Wanxiang Che, Tat-Seng Chua, and Ting Liu.
\newblock Trade-offs in large reasoning models: An empirical analysis of deliberative and adaptive reasoning over foundational capabilities.
\newblock \emph{arXiv preprint arXiv:2503.17979}, 2025{\natexlab{b}}.

\bibitem[Zhou et~al.(2025)Zhou, Liu, Zhao, Jangam, Srinivasa, Liu, Song, and Wang]{zhou2025hidden}
Kaiwen Zhou, Chengzhi Liu, Xuandong Zhao, Shreedhar Jangam, Jayanth Srinivasa, Gaowen Liu, Dawn Song, and Xin~Eric Wang.
\newblock The hidden risks of large reasoning models: A safety assessment of r1.
\newblock \emph{arXiv preprint arXiv:2502.12659}, 2025.

\bibitem[Zhou et~al.(2024)Zhou, Wang, Xiong, Xia, Gu, Chai, Zhu, Huang, Dou, Xi, et~al.]{zhou2024easyjailbreak}
Weikang Zhou, Xiao Wang, Limao Xiong, Han Xia, Yingshuang Gu, Mingxu Chai, Fukang Zhu, Caishuang Huang, Shihan Dou, Zhiheng Xi, et~al.
\newblock Easyjailbreak: A unified framework for jailbreaking large language models.
\newblock \emph{arXiv preprint arXiv:2403.12171}, 2024.

\end{thebibliography}
\bibliographystyle{iclr2025_conference}
\clearpage
\appendix

\end{document}